\RequirePackage{fix-cm}
\documentclass[10pt,twocolumn,letterpaper]{article}
\usepackage[pagenumbers]{cvpr}

\usepackage{microtype}
\usepackage{tabularx}
\usepackage{multirow}
\usepackage{array}
\usepackage{pifont,tikz}
\usepackage{placeins}
\newcommand{\hwbscore}{\textsc{HWBScore}}

\newcommand{\hwbappref}[1]{\ref{#1}}

\newcommand{\hwbinlinecaption}[2]{\captionsetup{hypcap=false}\captionof{#1}{#2}}

\definecolor{cvprblue}{rgb}{0.21,0.49,0.74}
\usepackage[pagebackref,breaklinks,colorlinks,allcolors=cvprblue]{hyperref}
\hypersetup{pdftitle={HandWorldBench: A Unified Benchmark for Hand-Centric World Models},pdfauthor={Zhaolu Wang, Jiacheng Xu, Lihong Wang, Shuqing Wang, Jianjie Fang, Ziyou Wang, Xvyuan Liu, Rongze Tang, Yiding Ma, Xin Zhang, Jiawei Chen, Wei Wu, Chen Gao, Yong Li, Zhibo Chen}}
\usepackage{fontawesome}
\newcommand{\bench}{HandWorldBench}
\title{HandWorldBench: A Unified Benchmark\\for Hand-Centric World Models}
\author{%
Zhaolu Wang$^{1,*}$ \quad Jiacheng Xu$^{2,*}$ \quad Lihong Wang$^{1,*}$ \quad
Shuqing Wang$^{3,*}$ \quad Jianjie Fang$^{2,\ddagger}$\\[2pt]
Ziyou Wang$^{2}$ \quad Xvyuan Liu$^{2}$ \quad Rongze Tang$^{1}$ \quad
Yiding Ma$^{2}$ \quad Xin Zhang$^{4}$\\[2pt]
Jiawei Chen$^{5}$ \quad Wei Wu$^{4}$ \quad Chen Gao$^{2,\dagger}$ \quad
Yong Li$^{2,\dagger}$ \quad Zhibo Chen$^{6}$\\[5pt]
{\normalsize $^{1}$Zhongguancun Academy \quad $^{2}$Tsinghua University \quad
$^{3}$Tianjin University}\\[1pt]
{\normalsize $^{4}$Manifold AI \quad $^{5}$Zhejiang University \quad $^{6}$University of Science and Technology of China}%
}
\begin{document}
\raggedbottom
\twocolumn[{
\renewcommand\twocolumn[1][]{#1}
\maketitle
\vspace*{-14pt}
\noindent\makebox[\textwidth][c]{%
\href{https://embodiedcity.github.io/HandWorldBench/}{\textcolor{magenta}{\faGlobe}\,\textit{Project Page}}%
\hspace{1cm}
\href{https://github.com/EmbodiedCity/HandWorldBench.code}{\textcolor{black}{\faGithub}\,\textit{Code}}%
\hspace{1cm}
\href{https://huggingface.co/datasets/EmbodiedCity/HandWorldBench}{\textcolor{orange!85!yellow}{\faCheckSquareO}\,\textit{Evaluation Set}}%
\hspace{1cm}
\href{https://huggingface.co/datasets/EmbodiedCity/HandWorldBench-Corpus}{\textcolor{teal}{\faDatabase}\,\textit{Corpus}}%
}\par\vspace{0.55em}
\noindent\begin{minipage}{\textwidth}\centering
\includegraphics[width=\textwidth]{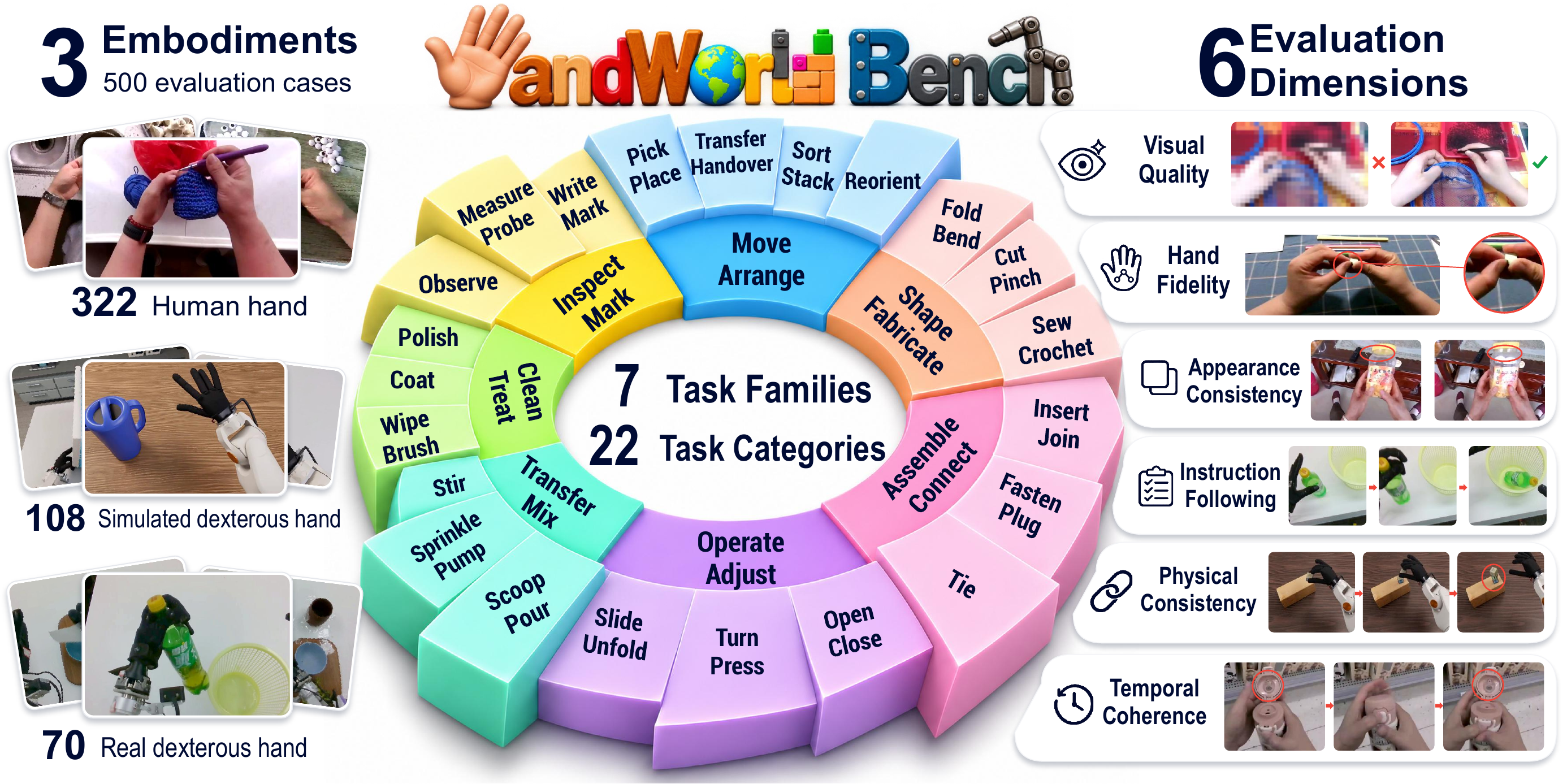}
\captionof{figure}{\textbf{Overview of \bench{}.} Left: 500 evaluation cases span human hands, simulated dexterous hands, and real dexterous hands. Center: seven task families encompass 22 task categories. Right: six evaluation dimensions capture complementary aspects of hand-centric generation and interaction prediction.}
\label{fig:teaser}
\end{minipage}
}]
\begingroup
\renewcommand{\thefootnote}{}
\makeatletter\renewcommand{\@makefntext}[1]{\noindent#1}\makeatother
\footnotetext{\noindent
$^{*}$Equal contribution.\\
$^{\dagger}$Corresponding authors: \href{mailto:chgao96@gmail.com}{chgao96@gmail.com}, \href{mailto:liyong07@tsinghua.edu.cn}{liyong07@tsinghua.edu.cn}.\\
$^{\ddagger}$Project Leader: \href{mailto:fjj25@mails.tsinghua.edu.cn}{fjj25@mails.tsinghua.edu.cn}.}
\endgroup
\begin{abstract}
Recent advances in world models are opening new possibilities for fine-grained embodied tasks, with hand--object interaction offering a particularly demanding setting.
Modeling these interactions involves capturing how hand actions change objects and how the resulting states persist over time.
Yet systematic evaluation tailored to these hand-centric capabilities remains underdeveloped.
To address this gap, we introduce \textbf{\bench}, a unified benchmark that systematically evaluates hand-centric world models across human hands, simulated dexterous hands, and real dexterous hands along six dimensions: visual quality, hand fidelity, appearance consistency, instruction following, physical consistency, and temporal coherence.
To construct the benchmark, we develop a data curation pipeline that extracts hand--object interaction clips and screens them for quality, yielding 20,500 high-quality egocentric clips from over 600 hours of footage.
We then manually select 500 clips covering seven task families for evaluation.
Benchmarking eleven models under a unified protocol reveals that stronger hand-pose control does not consistently translate into coherent object-state evolution and persistence or robust cross-embodiment performance, with several specialized hand-control models underperforming general-purpose text/image-to-video generators on dexterous-hand tasks.
\end{abstract}

\begin{table*}[t]
  \centering
  \caption{\textbf{Coverage of representative benchmark frameworks and evaluation protocols.}}
  \label{tab:comparison}
  \begingroup
\definecolor{refblue}{HTML}{154F8B}
\definecolor{Tcol}{HTML}{287BB4}
\definecolor{Vcol}{HTML}{D35C3F}
\definecolor{Ccol}{HTML}{33867D}
\definecolor{Acol}{HTML}{8060A5}
\newcommand{\hwbCmpYes}{\textcolor{black}{\ding{51}}}
\newcommand{\hwbCmpNo}{\textcolor{black!13}{\ding{55}}}
\newcommand{\hwbInputTile}[2]{\tikz[baseline=(letter.base)]{
  \ifnum#2=1
    \node[rounded corners=1.5pt,draw=#1col!85!black,line width=.35pt,top color=#1col!68,bottom color=#1col,minimum width=9.5pt,minimum height=9.5pt,inner sep=0pt,text=white,font=\sffamily\bfseries\fontsize{7.5}{8}\selectfont] (letter) {#1};
  \else
    \node[rounded corners=1.5pt,draw=black!10,line width=.3pt,fill=white,minimum width=9.5pt,minimum height=9.5pt,inner sep=0pt,text=black!20,font=\sffamily\bfseries\fontsize{7.5}{8}\selectfont] (letter) {#1};
  \fi}}
\setlength{\tabcolsep}{1.7pt}
\begin{minipage}{\textwidth}
\footnotesize
\renewcommand{\arraystretch}{1.13}
\begin{tabular*}{\linewidth}{@{\extracolsep{\fill}}lc*{3}{c}c*{6}{c}cr@{}}
\toprule[.8pt]
\multirow[c]{3}{*}[-1.5pt]{\shortstack[l]{Benchmark /\\Evaluation}} &
\multirow[c]{3}{*}[-1.5pt]{\shortstack{Unified\\hand\\evaluation}} &
\multicolumn{3}{c}{Video data} &
\multirow[c]{3}{*}[-1.5pt]{\shortstack{Native\\inputs}} &
\multicolumn{6}{c}{Evaluation} &
\multirow[c]{3}{*}[-1.5pt]{\shortstack{Human\\eval.}} &
\multirow[c]{3}{*}[-1.5pt]{Scale} \\
\cmidrule(lr){3-5}\cmidrule(lr){7-12}
& & Human & Sim. & Real & & Visual & Hand & Appear. & Instr. & Phys. & Temp. & & \\[-1pt]
& & hands & robots & robots & & Quality & Fidelity & Consist. & Follow. & Consist. & Coher. & & \\
\midrule[.45pt]
\multicolumn{14}{@{}l}{\textit{Benchmarks}} \\[-1pt]
WBench~\cite{ying2026wbench} & \hwbCmpNo & \hwbCmpNo & \hwbCmpNo & \hwbCmpNo & \hwbInputTile{T}{1}\hspace{1.2pt}\hwbInputTile{V}{1}\hspace{1.2pt}\hwbInputTile{C}{1}\hspace{1.2pt}\hwbInputTile{A}{0} & \hwbCmpYes & \hwbCmpNo & \hwbCmpYes & \hwbCmpYes & \hwbCmpYes & \hwbCmpYes & \hwbCmpYes & 289 cases \\
WorldArena 2.0~\cite{li2026worldarena2,shang2026worldarena} & \hwbCmpNo & \hwbCmpNo & \hwbCmpYes & \hwbCmpYes & \hwbInputTile{T}{1}\hspace{1.2pt}\hwbInputTile{V}{1}\hspace{1.2pt}\hwbInputTile{C}{0}\hspace{1.2pt}\hwbInputTile{A}{1} & \hwbCmpYes & \hwbCmpNo & \hwbCmpYes & \hwbCmpYes & \hwbCmpYes & \hwbCmpYes & \hwbCmpYes & 3 platforms \\
iWorld-Bench~\cite{fang2026iworldbench} & \hwbCmpNo & \hwbCmpNo & \hwbCmpYes & \hwbCmpYes & \hwbInputTile{T}{1}\hspace{1.2pt}\hwbInputTile{V}{1}\hspace{1.2pt}\hwbInputTile{C}{1}\hspace{1.2pt}\hwbInputTile{A}{0} & \hwbCmpYes & \hwbCmpNo & \hwbCmpYes & \hwbCmpYes & \hwbCmpNo & \hwbCmpYes & \hwbCmpYes & 4,900 tasks \\
RoboWM-Bench~\cite{jiang2026robowm} & \hwbCmpNo & \hwbCmpYes & \hwbCmpYes & \hwbCmpYes & \hwbInputTile{T}{1}\hspace{1.2pt}\hwbInputTile{V}{1}\hspace{1.2pt}\hwbInputTile{C}{0}\hspace{1.2pt}\hwbInputTile{A}{0} & \hwbCmpYes & \hwbCmpYes & \hwbCmpYes & \hwbCmpYes & \hwbCmpYes & \hwbCmpYes & \hwbCmpNo & 24 tasks \\
H2R-Bench~\cite{rong2026h2r} & \hwbCmpNo & \hwbCmpYes & \hwbCmpNo & \hwbCmpNo & \hwbInputTile{T}{1}\hspace{1.2pt}\hwbInputTile{V}{1}\hspace{1.2pt}\hwbInputTile{C}{0}\hspace{1.2pt}\hwbInputTile{A}{0} & \hwbCmpYes & \hwbCmpYes & \hwbCmpNo & \hwbCmpYes & \hwbCmpYes & \hwbCmpYes & \hwbCmpYes & 240 cases \\
WorldOracle~\cite{duan2026worldoracle} & \hwbCmpNo & \hwbCmpYes & \hwbCmpNo & \hwbCmpYes & \hwbInputTile{T}{1}\hspace{1.2pt}\hwbInputTile{V}{1}\hspace{1.2pt}\hwbInputTile{C}{1}\hspace{1.2pt}\hwbInputTile{A}{1} & \hwbCmpYes & \hwbCmpNo & \hwbCmpYes & \hwbCmpNo & \hwbCmpYes & \hwbCmpYes & \hwbCmpYes & 335 episodes \\
\midrule[.45pt]
\multicolumn{14}{@{}l}{\textit{Evaluation protocols introduced with a model}} \\[-1pt]
Hand2World~\cite{wang2026hand2world} & \hwbCmpNo & \hwbCmpYes & \hwbCmpNo & \hwbCmpNo & \hwbInputTile{T}{1}\hspace{1.2pt}\hwbInputTile{V}{1}\hspace{1.2pt}\hwbInputTile{C}{1}\hspace{1.2pt}\hwbInputTile{A}{1} & \hwbCmpYes & \hwbCmpNo & \hwbCmpNo & \hwbCmpNo & \hwbCmpNo & \hwbCmpYes & \hwbCmpNo & 3 datasets \\
EgoHOI~\cite{li2026egohoi} & \hwbCmpNo & \hwbCmpYes & \hwbCmpNo & \hwbCmpNo & \hwbInputTile{T}{1}\hspace{1.2pt}\hwbInputTile{V}{1}\hspace{1.2pt}\hwbInputTile{C}{1}\hspace{1.2pt}\hwbInputTile{A}{1} & \hwbCmpYes & \hwbCmpYes & \hwbCmpYes & \hwbCmpNo & \hwbCmpYes & \hwbCmpYes & \hwbCmpNo & 100 clips \\
JointControlVideo~\cite{zhang2026jointcontrol} & \hwbCmpNo & \hwbCmpYes & \hwbCmpNo & \hwbCmpYes & \hwbInputTile{T}{1}\hspace{1.2pt}\hwbInputTile{V}{1}\hspace{1.2pt}\hwbInputTile{C}{0}\hspace{1.2pt}\hwbInputTile{A}{1} & \hwbCmpYes & \hwbCmpYes & \hwbCmpNo & \hwbCmpNo & \hwbCmpNo & \hwbCmpNo & \hwbCmpYes & 70 sequences \\
\midrule[.45pt]
\textbf{HandWorldBench} & \hwbCmpYes & \hwbCmpYes & \hwbCmpYes & \hwbCmpYes & \hwbInputTile{T}{1}\hspace{1.2pt}\hwbInputTile{V}{1}\hspace{1.2pt}\hwbInputTile{C}{1}\hspace{1.2pt}\hwbInputTile{A}{1} & \hwbCmpYes & \hwbCmpYes & \hwbCmpYes & \hwbCmpYes & \hwbCmpYes & \hwbCmpYes & \hwbCmpYes & \textbf{500 cases} \\
\bottomrule[.8pt]
\end{tabular*}
\par\vspace{5pt}
\fontsize{7}{8.7}\selectfont
\hwbInputTile{T}{1}\,Text\quad\hwbInputTile{V}{1}\,Visual context\quad\hwbInputTile{C}{1}\,Camera/navigation\quad\hwbInputTile{A}{1}\,Hand/arm control.\quad Unified hand evaluation jointly assesses hand structure and control, object-state changes, and interaction memory within shared tasks. Human eval. denotes reported human ratings or preferences; the WorldArena 2.0 entry includes the original WorldArena human evaluation.
\end{minipage}

  \endgroup
\end{table*}

\section{Introduction}
\label{sec:introduction}

Interactive world models are evolving from open-world exploration toward fine-grained embodied interaction.
Advances in generating action-controllable environments have established a foundation for predicting visual experience under user intervention~\cite{bruce2024genie,parkerholder2024genie2}.
In embodied tasks, this prediction must capture how an agent's actions change the objects it engages with.
Hand--object interaction brings this requirement into focus: hands support varied contacts and transformations, connecting egocentric experience to dexterous manipulation.
This connection has motivated growing interest in hand-centric world models.

A growing body of work extends world modeling from locomotion and robotic-arm manipulation to hand-centric interaction.
Recent methods use meshes, joints, and poses to control hand motion~\cite{wang2026hand2world,zhang2026jointcontrol,gao2026pam,fang2026worldscapemoe}.
Contact and scene modeling support object interaction and continued manipulation~\cite{li2026egohoi,kim2026dwm,hao2026egosim}, with applications to data synthesis and digital teleoperation~\cite{gao2026pam,zhao2026rynnworldteleop}.
Yet these advances are largely assessed through model-specific tasks and validation targets, making progress across methods difficult to compare.
This growing field still lacks a unified benchmark for systematic evaluation across tasks and embodiments.

Existing video and world-model benchmarks evaluate visual quality, interaction and navigation control, and embodied utility~\cite{huang2024vbench,ying2026wbench,fang2026iworldbench,li2026worldarena2}.
Manipulation-focused benchmarks further assess physical executability, human-to-robot transfer, and agreement with real executions~\cite{jiang2026robowm,rong2026h2r,duan2026worldoracle} (Table~\ref{tab:comparison}).
These benchmarks provide evidence about interaction quality and task outcomes; comparing hand-centric models further requires resolving how the prescribed motion produces those outcomes.
Small finger contacts can determine an operation's outcome, while occlusion requires object states and previously observed details to remain consistent when they reappear.
Coordinated finger motions also make control fidelity consequential: similar overall hand movements can differ in joint configuration, contact location, and object response.
Evaluation must distinguish errors in following the supplied control from errors in the resulting interaction, using correspondences between native control inputs and generated motion.
These judgments must extend across human hands and simulated and real dexterous hands with different morphologies and kinematics, as well as operations with different completion and persistence requirements.
Existing protocols have yet to organize these requirements into a common evaluation of hand-centric world models.

We introduce \textbf{\bench}, a unified benchmark for hand-centric world models across human hands, simulated dexterous hands, and real dexterous hands.
Our data curation pipeline yields 20,500 high-quality egocentric clips from over 600 hours of footage.
From this corpus, we manually select 500 clips spanning seven task families to form the evaluation set.
To compare methods across embodiments, we establish shared structural semantics and spatiotemporal correspondences, then adapt the resulting controls to each model's native interface.
Task annotations connect action requirements to evidence of object-state changes, subsequent persistence, and the preservation of previously observed details.
Our six-dimensional protocol combines visual and geometric measurements with these task-specific judgments to evaluate generated interactions under shared scoring criteria.

Our evaluation of eleven models shows that stronger hand-pose control can coexist with weak object-state evolution and persistence.
Several specialized hand-control models also underperform general-purpose text/image-to-video generators on dexterous-hand tasks.
These findings identify a gap between motion control and interaction prediction: realizing a prescribed movement does not ensure that its consequences are modeled coherently, and specialization in hand control does not ensure robust performance across embodiments.

Our contributions are:
\begin{itemize}
    \item \textbf{A unified benchmark.}
    \bench{} establishes common interaction tasks and six evaluation dimensions for comparing world models across human hands and simulated and real dexterous hands.
    \item \textbf{A data curation pipeline and corpus.}
    We construct high-quality interaction data through quality screening and signal-alignment checks, producing 20,500 clips from which we manually select 500 for evaluation.
    \item \textbf{A systematic evaluation.}
    We systematically evaluate 11 representative models under a unified protocol, characterizing their performance across hand-centric tasks and embodiments.
\end{itemize}

\begin{figure*}[t]
\centering
\includegraphics[width=\textwidth]{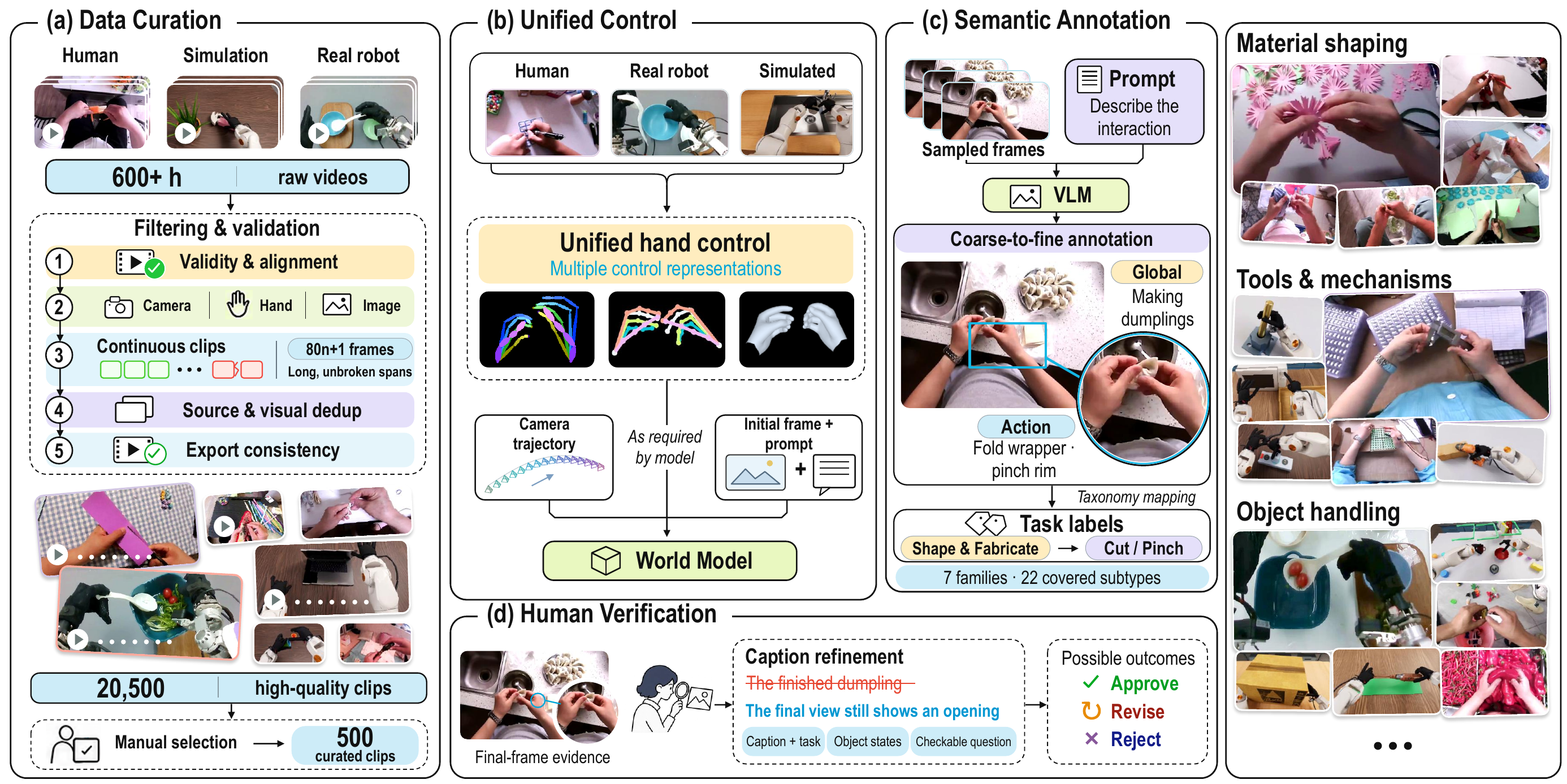}
\caption{\textbf{Data construction and annotation pipeline.} (a) Filtering and validation yield 20,500 high-quality clips from over 600 hours of footage; manual selection produces 500 evaluation clips. (b) Multiple hand-control representations support different model interfaces. (c) VLM-assisted annotation provides interaction descriptions and task labels. (d) Human verification refines descriptions, object states, and evaluation questions. Right: representative manipulation scenarios.}
\label{fig:data_pipeline}
\end{figure*}

\setcounter{dbltopnumber}{1}
\begin{table*}[t]
\centering
\begingroup
\renewenvironment{table*}[1][]{}{}
\captionsetup{type=table}
\begin{table*}[t]
\centering
\caption{Overall results across all 17 metrics on common valid cases per metric and protocol. Higher is better. Metrics are reported on a 0--1 scale and HWBScore on a 0--100 scale. \textbf{Bold} and \underline{underlined} values denote the best and second-best results, determined before rounding. Both panels use the same model order, sorted by HWBScore. Cases reports reference-defined applicability (JEPA: collection size).}
\label{tab:main_results}
\begingroup
\fontsize{8}{9.2}\selectfont
\setlength{\tabcolsep}{1.5pt}
\renewcommand{\arraystretch}{1.14}
\begin{tabular*}{\textwidth}{@{\extracolsep{\fill}}l*{9}{c}@{}}
\toprule
\multirow[c]{3}{*}[-1.5pt]{Model} & \multicolumn{3}{c}{Visual quality} & \multicolumn{3}{c}{Hand fidelity} & \multicolumn{3}{c}{Appearance consistency} \\
\cmidrule(lr){2-4}\cmidrule(lr){5-7}\cmidrule(lr){8-10}
 & \multirow[c]{2}{*}{MUSIQ} & Visual & JEPA & Pose & Keypoint & Hand & Entity & Background & Photometric \\
 &  & Plausibility & Similarity & Control & Coverage & Morphology & Consistency & Consistency & Consistency \\
\midrule
JointControlVideo~\cite{zhang2026jointcontrol} & 0.640 & 0.585 & \underline{0.817} & 0.416 & 0.574 & 0.844 & \textbf{0.847} & \underline{0.869} & 0.612 \\
HunyuanVideo 1.5~\cite{wu2025hunyuanvideo15} & \underline{0.685} & \textbf{0.612} & \textbf{0.829} & 0.303 & 0.622 & \underline{0.878} & \underline{0.842} & \textbf{0.892} & \textbf{0.740} \\
Wan2.2~\cite{wan2025wan,wan2025wan22} & 0.630 & \underline{0.601} & 0.757 & 0.227 & 0.623 & 0.841 & 0.842 & 0.856 & \underline{0.680} \\
Wan2.1~\cite{wan2025wan} & 0.678 & 0.588 & 0.780 & 0.228 & 0.554 & 0.829 & 0.821 & 0.846 & 0.616 \\
EgoHOI~\cite{li2026egohoi} & 0.655 & 0.579 & 0.715 & 0.350 & 0.637 & 0.870 & 0.803 & 0.825 & 0.552 \\
DWM~\cite{kim2026dwm} & \textbf{0.695} & 0.564 & 0.578 & 0.423 & 0.577 & \textbf{0.887} & 0.826 & 0.825 & 0.675 \\
EgoSim~\cite{hao2026egosim} & 0.656 & 0.562 & 0.760 & \underline{0.428} & \underline{0.637} & 0.851 & 0.827 & 0.793 & 0.613 \\
PAM~\cite{gao2026pam} & 0.571 & 0.571 & 0.660 & 0.299 & 0.474 & 0.842 & 0.810 & 0.835 & 0.593 \\
CogVideoX1.5~\cite{yang2025cogvideox} & 0.578 & 0.584 & 0.712 & 0.218 & 0.510 & 0.789 & 0.826 & 0.820 & 0.611 \\
RynnWorld-Teleop~\cite{zhao2026rynnworldteleop} & 0.509 & 0.548 & 0.757 & 0.279 & 0.418 & 0.789 & 0.807 & 0.805 & 0.635 \\
Hand2World~\cite{wang2026hand2world} & 0.645 & 0.558 & 0.390 & \textbf{0.465} & \textbf{0.718} & 0.771 & 0.721 & 0.476 & 0.328 \\
\midrule
Cases & 500 & 500 & 500 & 500 & 500 & 322 & 500 & 500 & 500 \\
\bottomrule
\end{tabular*}
\par\vspace{4pt}
\begin{tabular*}{\textwidth}{@{\extracolsep{\fill}}l*{9}{c}@{}}
\toprule
\multirow[c]{3}{*}[-1.5pt]{Model} & \multicolumn{2}{c}{Instruction following} & \multicolumn{4}{c}{Physical consistency} & \multicolumn{2}{c}{Temporal coherence} & \multirow[c]{3}{*}[-1.5pt]{HWBScore} \\
\cmidrule(lr){2-3}\cmidrule(lr){4-7}\cmidrule(lr){8-9}
 & Action & Event & Trajectory & Causal & Evolution & State & State & Memory &  \\
 & Adherence & Adherence & Accuracy & Plausibility & Correctness & Regulation & Progression & Consistency &  \\
\midrule
JointControlVideo & \underline{0.480} & \textbf{0.520} & 0.214 & \underline{0.586} & \textbf{0.345} & \textbf{0.425} & \textbf{0.385} & \underline{0.356} & \textbf{55.51} \\
HunyuanVideo 1.5 & 0.445 & 0.399 & 0.194 & \textbf{0.590} & 0.284 & 0.240 & 0.339 & 0.159 & \underline{52.22} \\
Wan2.2 & \textbf{0.483} & \underline{0.433} & 0.170 & 0.542 & \underline{0.291} & \underline{0.333} & \underline{0.366} & 0.209 & 51.65 \\
Wan2.1 & 0.478 & 0.393 & 0.143 & 0.541 & 0.250 & 0.228 & 0.335 & 0.293 & 50.34 \\
EgoHOI & 0.453 & 0.372 & 0.174 & 0.465 & 0.143 & 0.208 & 0.181 & \textbf{0.357} & 48.74 \\
DWM & 0.421 & 0.346 & \underline{0.232} & 0.444 & 0.161 & 0.112 & 0.234 & 0.168 & 47.31 \\
EgoSim & 0.406 & 0.317 & \textbf{0.273} & 0.429 & 0.071 & 0.100 & 0.134 & 0.182 & 46.33 \\
PAM & 0.454 & 0.402 & 0.217 & 0.503 & 0.172 & 0.117 & 0.214 & 0.183 & 46.06 \\
CogVideoX1.5 & 0.427 & 0.312 & 0.208 & 0.498 & 0.184 & 0.150 & 0.233 & 0.248 & 45.88 \\
RynnWorld-Teleop & 0.427 & 0.329 & 0.216 & 0.493 & 0.112 & 0.085 & 0.122 & 0.084 & 42.60 \\
Hand2World & 0.334 & 0.195 & 0.080 & 0.238 & 0.008 & 0.000 & 0.019 & 0.038 & 34.42 \\
\midrule
Cases & 500 & 175 & 500 & 500 & 209 & 120 & 209 & 22 &  \\
\bottomrule
\end{tabular*}
\endgroup
\end{table*}

\endgroup
\end{table*}

\begin{figure*}[t]
\centering
\includegraphics[width=\textwidth]{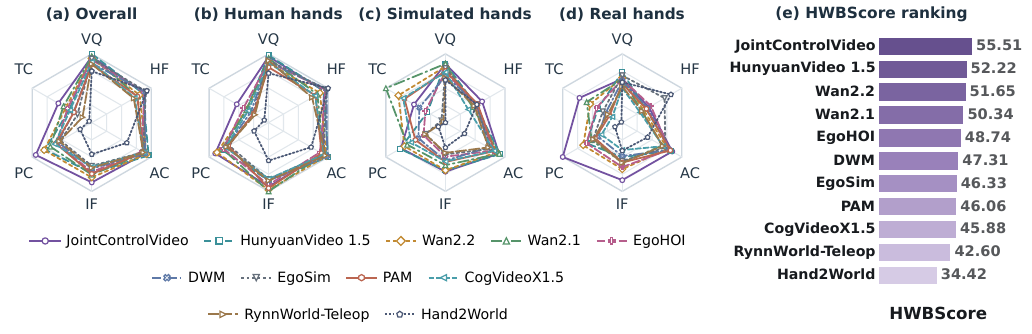}
\caption{\textbf{Overall ranking and cross-embodiment profiles.} Left: visual quality (VQ), hand fidelity (HF), appearance consistency (AC), instruction following (IF), physical consistency (PC), and temporal coherence (TC). Radar colors and markers identify models. For each category, the four radars share the maximum over eleven models and four evaluation sets as their outer radius. Right: raw \hwbscore{}, colored by rank.}
\label{fig:radar}
\end{figure*}

\begin{figure}[t]
\centering
\includegraphics[width=\columnwidth]{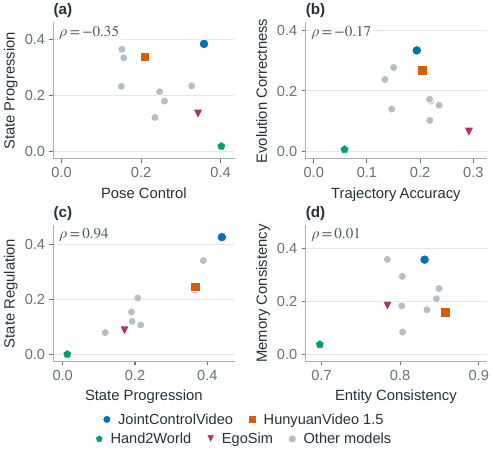}
\caption{\textbf{Relationships between interaction capabilities.}
Each point represents one of eleven models, averaged over matched tasks for both metrics within each panel ($n=209,200,117,22$ for a--d).
$\rho$ denotes Spearman rank correlation.}
\label{fig:capability_diagnostics_scatter}
\end{figure}

\begin{figure*}[t]
\centering
\includegraphics[width=\textwidth]{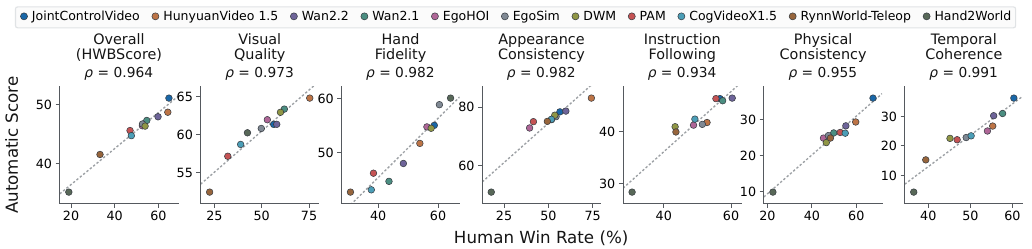}
\caption{\textbf{Alignment with human judgments.} Each point represents one model on matched human-evaluation tasks (75 tasks for the overall panel). Human win rates and automatic scores are displayed on a 0--100 scale. Each panel reports Spearman\textquotesingle s $\rho$ across eleven models; dashed lines indicate linear trends.}
\label{fig:human_alignment}
\par\vspace{8pt}
\centering
\includegraphics[width=\textwidth]{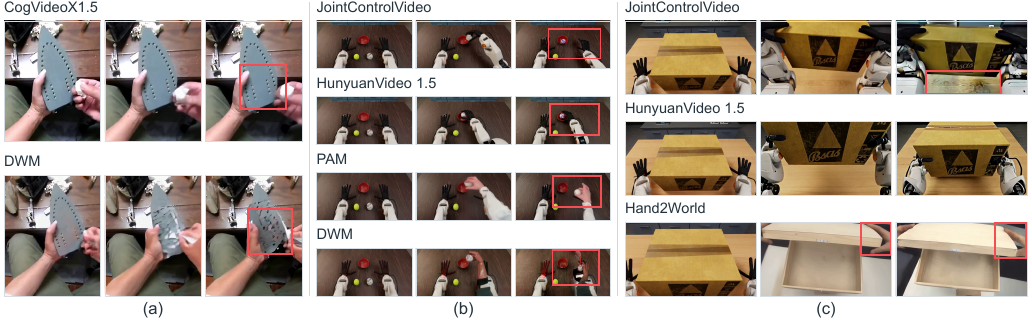}
\caption{\textbf{Qualitative comparison of hand-centric interactions.}
(a) Soleplate wiping. (b) Ball placement. (c) Carton lifting.
Frames progress chronologically from left to right; red boxes highlight local details, embodiment changes, and background distortions.}
\label{fig:qualitative}
\end{figure*}

\section{Related Work}
\label{sec:related}

\paragraph{Hand-centric world models.}
Controllable hand--object generation combines motion constraints with scene modeling.
Projected meshes, sparse joints, and pose-based rendering provide explicit motion constraints within visually conditioned generation~\cite{wang2026hand2world,zhang2026jointcontrol,gao2026pam}.
The interaction model determines how this prescribed motion affects objects: contact-aware embeddings encode geometric and kinematic priors, while scene-conditioned generation anchors the interaction to surrounding geometry~\cite{li2026egohoi,kim2026dwm}.
Explicitly updating the underlying scene state carries these effects into later interactions~\cite{hao2026egosim}.
Applications to pose-estimation data synthesis and digital teleoperation connect generated interactions to downstream learning~\cite{gao2026pam,zhao2026rynnworldteleop}.

\paragraph{Video and world model evaluation.}
Video-quality benchmarks organize perceptual, temporal, and semantic diagnostics, extending to anatomical, physical, and commonsense assessments of plausible world behavior~\cite{huang2024vbench,zheng2025vbench2}.
Interactive benchmarks test instruction-driven transitions and cross-turn consistency~\cite{ying2026wbench}, with navigation protocols additionally linking action-following accuracy to spatial memory on return paths~\cite{fang2026iworldbench}.
Embodied benchmarks assess the utility of predictions for data synthesis, planning, and policy learning across modalities and platforms~\cite{shang2026worldarena,li2026worldarena2}.
Manipulation evaluation adds physical execution and task-specific references: simulation replay tests whether generated behavior can be executed~\cite{jiang2026robowm}, transfer annotations test preservation of goals and functional contacts across embodiments~\cite{rong2026h2r}, and paired real executions test the predicted effects of specified actions~\cite{duan2026worldoracle}.
\bench{} unifies hand-control, object-state, and memory assessment across human and dexterous hands.

\section{HandWorldBench}
\label{sec:benchmark}

We construct HandWorldBench by curating egocentric interaction data across embodiments (Section~\ref{sec:dataset}), organizing model inputs through unified control interfaces, and establishing task annotations (Section~\ref{sec:annotations}).
We evaluate the generated interactions with seventeen metrics across six dimensions (Section~\ref{sec:metrics}), then aggregate these measurements into category scores and \hwbscore{} for model comparison (Section~\ref{sec:aggregation}).

\subsection{Dataset Construction}
\label{sec:dataset}
Figure~\ref{fig:data_pipeline}(a) outlines data curation from our in-house human-hand recordings, HandsOn~\cite{luo2026handson} (simulated dexterous hands), and our own real dexterous-hand recordings.
Starting from over 600 hours of footage across the three embodiments, we validate video readability, timestamps, and alignment with available hand and camera signals.
Screening checks camera stability, hand visibility, and image quality.
Visual checks assess blur, exposure, and hand size.
We extract continuous human-hand clips of $80n+1$ frames ($n\geq1$) and retain complete robot episodes, screen for overlapping sources and visual duplicates, and check exported frame counts and signal correspondence.
Source intervals and available validity masks accompany each clip throughout processing.
From 20,500 curated clips, we manually select 500 for evaluation using the criteria in Appendix~\hwbappref{app:data_curation:selection}; 20,000 remain for non-test use.
Evaluation cases cover seven task families and 22 task categories (composition, durations, and coverage in Appendices~\hwbappref{app:statistics:composition}--\hwbappref{app:statistics:subsets}), including material shaping, tools and mechanisms, and object handling (Figure~\ref{fig:data_pipeline}, right).

\subsection{Control Interfaces and Annotations}
\label{sec:annotations}
We unify control through native-signal parsing, structural normalization, spatiotemporal alignment, and model-specific encoding (Figure~\ref{fig:data_pipeline}(b)).
Human-hand inputs provide native 3D keypoints, left/right validity, camera calibration, and visual-inertial odometry poses.
For dexterous hands, forward kinematics uses joint states, palm poses, and robot assets to obtain landmarks and surfaces, which are transformed into camera coordinates.
Each hand follows a fixed 21-point order: one wrist or palm anchor and four landmarks per finger, with hand identity and validity retained alongside native kinematics and surface geometry.
Human meshes are fitted to keypoints or visually reconstructed as required.
Coordinates use meters, explicit world-to-camera transforms, and image-matched projection and viewport mappings.
Source frame indices, timestamps, and shared inference windows align images, geometry, and cameras.
Adapters encode these assets as skeletal videos, 3D joints, surface renderings, or scene conditions for each model.
The common interface establishes structural and spatiotemporal correspondence while preserving each model's conditioning format.

Shared operation categories, global task descriptions, and segmented action descriptions establish task semantics across embodiments (Figure~\ref{fig:data_pipeline}(c)).
Evaluation annotations anchor action requirements and state checkpoints to the same manipulated object and its before, during, and after windows, distinguishing minimum change from full completion.
Persistence checks follow state establishment to identify unsupported reversals.
Memory annotations specify local attributes and history, occlusion, and reappearance windows; evaluation first verifies a hiding-and-return opportunity, then compares separately observed historical and returning attributes.
Human reviewers verify task descriptions, object states, and evaluation evidence against reference frames (Figure~\ref{fig:data_pipeline}(d); schemas and examples in Appendices~\hwbappref{app:annotations:schema} and~\hwbappref{app:annotations:examples}).
Appendices~\hwbappref{app:unified_control:adapters} and~\hwbappref{app:baselines:conditioning} detail control adaptation and model information access.

\subsection{Evaluation Metrics}
\label{sec:metrics}

We organize 17 metrics into six dimensions that characterize visual quality, hand fidelity, appearance consistency, instruction following, physical consistency, and temporal coherence.
Task-specific evaluation uses reference-defined cases and their annotated temporal windows.
Semantic metrics use local Qwen3.5-9B~\cite{qwen2026qwen35} with fixed task-specific rubrics; Appendix~\hwbappref{app:metrics} details computations, rubrics, and scales for all seventeen metrics.
Individual-metric checks are described in Appendix~\hwbappref{app:human_evaluation:agreement}.

\paragraph{Visual Quality.}
MUSIQ~\cite{ke2021musiq} measures perceptual image quality; Visual Plausibility assesses visible physical artifacts with WBench's dedicated reward model~\cite{ying2026wbench,wbench2026vpweights}.
JEPA Similarity measures collection-level similarity between generated and reference video distributions using JEDi~\cite{luo2025jedi}, V-JEPA features~\cite{bardes2024vjepa}, and WorldArena's temporal sampling protocol~\cite{shang2026worldarena}.

\paragraph{Hand Fidelity.}
Pose Control measures agreement with reference keypoint positions using a capped image-space distance, with zero credit for missing predictions.
Keypoint Coverage measures the fraction of reference-valid, in-frame keypoints recovered by the detector, including geometrically valid occluded points.
Hand Morphology uses the VBench-2.0 hand classifier~\cite{zheng2025vbench2} to estimate anatomical plausibility in the human-hand subset.

\paragraph{Appearance Consistency.}
Entity Consistency compares CLIP features~\cite{radford2021clip} of reference-defined hands and manipulated objects, with penalties for missing predicted regions.
Background Consistency measures the temporal stability of DINOv3 features~\cite{simeoni2025dinov3} over background regions.
Photometric Consistency uses Depth Anything 3~\cite{lin2025da3} to establish background correspondences and compare colors across frames.

\paragraph{Instruction Following.}
Action Adherence evaluates requested hand behavior, while Event Adherence evaluates explicitly specified interaction events.
Both use fixed questions adapted from WBench~\cite{ying2026wbench}, grounded in task descriptions and annotated windows.

\paragraph{Physical Consistency.}
Trajectory Accuracy evaluates the motion of manipulated objects against their reference trajectories.
Causal Plausibility assesses whether visible effects follow plausibly from the interaction.
Evolution Correctness evaluates the physical plausibility of a transformation, with a progress gate accounting for whether the required change occurs.
State Regulation checks whether a specified state is established and retained in subsequent windows, accounting for permitted reversals.

\paragraph{Temporal Coherence.}
State Progression measures attainment of visible state milestones, including minimum-change and completion conditions.
Memory Consistency checks whether previously observed attributes remain consistent when they reappear after interaction or occlusion.
Historical and reappearance windows supply the evidence; frozen checklists specify the attributes to compare and their allowed changes.

\subsection{HWBScore and Aggregation}
\label{sec:aggregation}

All reported metrics are oriented so that higher values indicate better performance and are normalized to $[0,1]$.
For model $m$, metric $k$, and subset $b$, let $\bar{s}_{m,k}^{(b)}$ denote the score on the common valid cases for that metric and protocol.
Per-case metrics are averaged with equal case weights; JEPA Similarity is computed jointly over the corresponding video collection.
The overall collection pools applicable cases across embodiments (Appendix~\hwbappref{app:statistics:subsets}) before computing category scores.

Let $\mathcal{K}_{g}^{(b)}$ be the fixed set of applicable metrics in category $g$ for subset $b$.
We define the category score and the \textbf{HandWorldBench Score (\hwbscore{})} as
\begin{align}
 c_{m,g}^{(b)} &= \frac{1}{|\mathcal{K}_{g}^{(b)}|}
                  \sum_{k\in\mathcal{K}_{g}^{(b)}}\bar{s}_{m,k}^{(b)}, \\
 \mathrm{HWBScore}_{m}^{(b)} &= \frac{100}{6}\sum_{g=1}^{6}c_{m,g}^{(b)}.
\end{align}
Equal category weights give each dimension the same contribution, independent of its number of metrics.
The overall \hwbscore{} uses the pooled metric scores rather than averaging the three embodiment-level composite scores.
The overall leader remains unchanged across the aggregation schemes tested in Appendix~\hwbappref{app:diagnostics:sensitivity}.

\section{Experiments}
\label{sec:experiments}

\subsection{Experimental Setup}
\label{sec:experimental_setup}

We evaluate eleven models on 500 cases comprising 322 human-hand, 108 simulated dexterous-hand, and 70 real dexterous-hand interactions across seven task families and 22 task categories.
The comparison covers general-purpose video generators, including HunyuanVideo 1.5 and Wan2.2~\cite{yang2025cogvideox,wu2025hunyuanvideo15,wan2025wan,wan2025wan22}, and hand- or scene-conditioned models, such as JointControlVideo, EgoHOI, and EgoSim~\cite{gao2026pam,zhang2026jointcontrol,zhao2026rynnworldteleop,kim2026dwm,wang2026hand2world,li2026egohoi,hao2026egosim}.
We compare released configurations on shared reference cases using native visual, control, and scene inputs; details are shown in Appendix~\hwbappref{app:baselines:conditioning}.
The seventeen metrics and six-category \hwbscore{} (Sections~\ref{sec:metrics}--\ref{sec:aggregation}) yield overall results in Table~\ref{tab:main_results} and complete embodiment-level results in Appendices~\hwbappref{app:results:human}--\hwbappref{app:results:real}.
Generation protocols, stability, and scoring costs are detailed in Appendices~\hwbappref{app:baselines:versions}--\hwbappref{app:baselines:generation}, \hwbappref{app:diagnostics:stability}, and~\hwbappref{app:protocol:runtime}.

\subsection{Overall Performance}
\label{sec:overall}

\paragraph{Appearance and interaction outcomes.}
JointControlVideo leads with a \hwbscore{} of 55.51, followed by HunyuanVideo 1.5 at 52.22 and Wan2.2 at 51.65 (Table~\ref{tab:main_results}).
Appendix~\hwbappref{app:diagnostics:stability} specifies score and paired-difference confidence-interval estimation.
HunyuanVideo 1.5 leads visual quality and appearance consistency, while JointControlVideo gains its advantage in instruction following, physical consistency, and temporal coherence (Figure~\ref{fig:radar}).
Photometric Consistency favors HunyuanVideo 1.5 (0.740 versus 0.612), while State Regulation favors JointControlVideo (0.425 versus 0.240).
These measures test retaining visible scene properties and producing changed states that remain available to subsequent actions.
Selecting for appearance preservation therefore favors a different model from selecting for persistent manipulation outcomes.

\paragraph{Motion control and state prediction.}
The clearest separation occurs between variables constrained by motion inputs and the object changes that must be inferred from them.
Hand2World leads Pose Control at 0.465 but scores 0.019 on State Progression and 0.000 on State Regulation.
JointControlVideo has lower Pose Control of 0.416, yet reaches 0.385 and 0.425 on the two state metrics.
The separation extends to object motion: EgoSim leads Trajectory Accuracy at 0.273, while its Evolution Correctness is 0.071.
Geometric conditions specify how the hand should move; they leave the model to infer how contact changes the object and how that change persists.
The ranking reversal separates accurate tracking of explicit motion inputs from prediction of their object-state consequences.

\subsection{Cross-Embodiment and Task Analysis}
\label{sec:cross_embodiment}

\paragraph{Cross-embodiment performance.}
JointControlVideo's pose-tracking advantage over HunyuanVideo 1.5 persists across dexterous-hand settings, while their ordering in state prediction reverses.
On simulated dexterous hands, JointControlVideo scores 0.335 in Pose Control, compared with 0.164 for HunyuanVideo 1.5.
HunyuanVideo 1.5 nevertheless leads Evolution Correctness and State Progression at 0.323 and 0.399, versus 0.304 and 0.364.
On real dexterous hands, JointControlVideo leads both state metrics at 0.443 and 0.475, against 0.228 and 0.270.
Appendix~\hwbappref{app:diagnostics:relationships} describes controlled checks of visual context, evaluation horizon, and reference access for these comparisons.
The advantage in tracking a geometric input thus persists more consistently than the advantage in predicting its effects on objects.

\paragraph{Performance across task families.}
Task-level results further locate the gap between reproducing an operation and realizing its defining object relation.
In Assemble \& Connect, PAM leads Action Adherence at 0.552, whereas Wan2.2 leads Evolution Correctness at 0.652.
For assembly, matching a manipulation pattern can leave the required connection between parts unrealized; the two rankings capture this distinction between recognizable behavior and its physical result.
Move \& Arrange also exposes dependence on embodiment within one task family.
On real dexterous hands, JointControlVideo leads action adherence and evolution correctness at 0.563 and 0.682.
On simulated hands, it ties HunyuanVideo 1.5 in action adherence, while HunyuanVideo 1.5 leads evolution correctness.
The relative advantage of specialized control therefore depends on the embodiment--task combination and the object relation that defines successful execution.

\subsection{Interaction Diagnostics}
\label{sec:diagnostics}

Motion and outcome rankings diverge even on matched tasks: Spearman correlations across the eleven models are $\rho=-0.35$ for Pose Control--State Progression and $\rho=-0.17$ for Trajectory Accuracy--Evolution Correctness (Figure~\ref{fig:capability_diagnostics_scatter}(a,b)).
Task-bootstrap and leave-one-model-out checks are specified in Appendix~\hwbappref{app:diagnostics:relationships}.

State Progression and State Regulation share a task-achievement requirement, reflected in their strong correlation of $\rho=0.94$ (Figure~\ref{fig:capability_diagnostics_scatter}(c)).
State Regulation gates retention credit on state establishment, so both scores depend on reaching the required state.
Yet EgoSim and CogVideoX1.5 have similar progression scores of 0.171 and 0.190, while their regulation scores are 0.085 and 0.154.
The shared prerequisite brings the overall rankings closer; subsequent retention still separates models with comparable progress.

Entity similarity and historical consistency operate at different levels of detail ($\rho=0.01$; Figure~\ref{fig:capability_diagnostics_scatter}(d)).
On this matched subset, JointControlVideo and HunyuanVideo 1.5 have similar entity scores but different memory retention (Appendix~\hwbappref{app:diagnostics:relationships}).
Similar overall features can coexist with changes to local attributes established earlier in the interaction.
Memory evaluation therefore asks whether the current object remains faithful to its observed history.

\subsection{Alignment with Human Judgments}
\label{sec:reliability}

We collect 20,244 preference judgments from 153 participants, covering 2,911 video pairs generated by eleven models across three egocentric embodiment settings.
Each judgment records a video-pair preference for overall performance or one evaluation dimension.

Figure~\ref{fig:human_alignment} shows strong alignment between automatic scores and human preferences at the model level: Spearman\textquotesingle s $\rho$ is 0.964 for \hwbscore{} and ranges from 0.934 to 0.991 across the six dimensions.
At the video-pair level, preference agreement is 81.7\% for overall performance and ranges from 76.9\% to 84.1\% across the six dimensions.
Physical Consistency achieves the highest pairwise agreement of 84.1\%.
Response handling and task-bootstrap confidence intervals are detailed in Appendices~\hwbappref{app:human_evaluation:instructions}--\hwbappref{app:human_evaluation:agreement}.

\subsection{Qualitative Analysis}
\label{sec:case-studies}

During soleplate wiping, CogVideoX1.5 retains the perforation pattern, whereas DWM replaces the holes with irregular markings (Figure~\ref{fig:qualitative}(a)).
Wiping changes contact and visibility while the underlying pattern should persist.
The contrast exposes a failure to keep the current appearance consistent with local properties established by the object's history, even as the manipulation continues.

Ball placement separates embodiment preservation from the required object relation (Figure~\ref{fig:qualitative}(b)).
JointControlVideo places the ball in the bowl while retaining the dexterous hands.
PAM and DWM introduce human hands, and the ball remains outside the bowl in the last displayed frame.
Continued hand motion accompanies both a change of acting embodiment and an unmet placement goal, showing how these two requirements can fail within one generated interaction.

Carton lifting tests continuity beyond the manipulated object (Figure~\ref{fig:qualitative}(c)).
JointControlVideo lifts the carton but distorts the background beneath it; Hand2World replaces the dexterous hands with human hands in later frames.
As the action reveals more of the scene, the generated background must remain compatible with the surrounding structure, and the manipulator must retain its supplied embodiment throughout the operation.
Further examples appear in Appendices~\hwbappref{app:qualitative:embodiments}--\hwbappref{app:qualitative:failures}.

\section{Conclusion}
\label{sec:conclusion}

We presented \bench{}, a unified benchmark that brings human hands, simulated dexterous hands, and real dexterous hands into a common framework of interaction tasks, control interfaces, and evaluation criteria.
A curation pipeline yields 20,500 high-quality egocentric clips, from which 500 interactions covering seven task families are manually selected for evaluation.
Across eleven models, accurate tracking of control signals does not consistently translate into the object changes and persistent states that those signals should induce.
The benefits of specialized control vary with embodiment and task, while general-purpose video generators achieve stronger interaction outcomes in several comparisons.
Strong agreement with human judgments supports the evaluation at both the model-ranking and video-pair levels.
\bench{} provides a common basis for assessing whether progress in motion control extends to interaction prediction that remains coherent across embodiments and tasks.

{\small
\bibliographystyle{ieeenat_fullname}
\bibliography{references}
}
\appendix
\twocolumn[{\centering{\Large\bfseries Appendix\par}\vspace{1.2em}}]

\makeatletter
\setlength{\@fptop}{0pt}
\setlength{\@fpsep}{12pt}
\setlength{\@fpbot}{0pt plus 1fil}
\setlength{\@dblfptop}{0pt}
\setlength{\@dblfpsep}{12pt}
\setlength{\@dblfpbot}{0pt plus 1fil}
\makeatother
\setlength{\dblfloatsep}{12pt plus 2pt minus 2pt}
\setlength{\dbltextfloatsep}{13pt plus 2pt minus 2pt}
\renewcommand{\dbltopfraction}{.98}
\renewcommand{\dblfloatpagefraction}{.80}
\renewcommand{\topfraction}{.93}
\renewcommand{\textfraction}{.06}
\setcounter{dbltopnumber}{4}

\section{Data Curation Pipeline}
\label{app:data_curation}
\bench{} organizes continuous hand--object interactions into reproducible evaluation cases. Each case preserves its source identity, frame interval, video hash, native measurements, and evaluation annotations. Source-specific screening establishes the visual reference; explicit time and image-coordinate mappings connect that reference to its motion signals and scoring windows.

\subsection{Data Sources}
\label{app:data_curation:sources}
\paragraph{Human egocentric recordings.}
Our in-house collection provides first-person manipulation recordings in MCAP containers, including image streams, available left/right hand skeletons, visual--inertial poses, camera calibration, timestamps, and task messages. The human-source inventory contains 58,629 MCAP portions, corresponding to about 612 hours of cumulative source-frame ranges. Selected clips retain 21 landmarks per hand, camera-space coordinates, image projections, validity masks, VIO poses, calibration, source frame indices, and available subtask text. This combination supports spatial conditioning and temporal traceability at the level of the original recording.

\paragraph{Simulated dexterous-hand episodes.}
HandsOn~\cite{luo2026handson} provides front and wrist renderings, native state/action tables, task instructions, and rendering provenance. Its source manifest contains 18,000 episodes across 54 task identifiers. The evaluation selection retains two episodes per identifier, producing 108 references. Selection favors variation in object instances, layout, hand configuration, and interaction. Episode Parquet files, shared schemas, camera/render metadata, and source task identifiers remain associated with the corresponding video.

\paragraph{Real dexterous-hand episodes.}
Our real dexterous-hand episodes are recorded on TI5 dexterous hands performing tabletop manipulation, with head and wrist camera videos, HDF5 state streams, and episode JSON metadata. Embodiment checks identify dexterous hands and remove parallel-jaw-gripper recordings, empty scenes, debugging footage, and executions with unclear completion evidence. The evaluation set contains 70 episodes across seven source task identifiers. Native instructions, recorded states, control commands, timestamps, and unprocessed videos retain their original field definitions and source associations.

\paragraph{Corpus and evaluation inventories.}
The selected pool contains 20,500 clips: 10,322 human, 6,108 simulated, and 4,070 real dexterous-hand clips. We hold out 500 references for evaluation (322, 108, and 70, respectively), leaving a non-test corpus of 20,000 clips (10,000, 6,000, and 4,000). Both partitions retain their source identities and selection records. In human corpus production, exclusion is applied at the original-recording level to separate non-test material from the benchmark references. A split audit confirms that the non-test corpus and the evaluation set share no recording or episode identity and no identical encoded video (SHA-256). Appendix~\ref{app:statistics} reports non-test corpus composition and the detailed statistics of the 500 evaluation references. The source footage totals over 600 hours: about 612 hours of human source-frame ranges, 56.1 hours of HandsOn episodes, and 37.7 hours of TI5 dexterous-hand episodes.

\subsection{Filtering and Validation}
\label{app:data_curation:filtering}
\paragraph{Time and coordinate alignment.}
Video, hand, and camera streams are joined through timestamps and original frame indices. Missing measurements retain their positions on the timeline through explicit masks. Alignment checks verify frame count, frame rate, image size, camera model, units, and coordinate conventions before projection. Image-space skeletal renderings share the RGB viewport and spatial transform. Robot measurements and estimated reference geometry retain the conventions and provenance of their respective sources.

\begin{figure*}[!t]
\centering\includegraphics[width=0.94\textwidth]{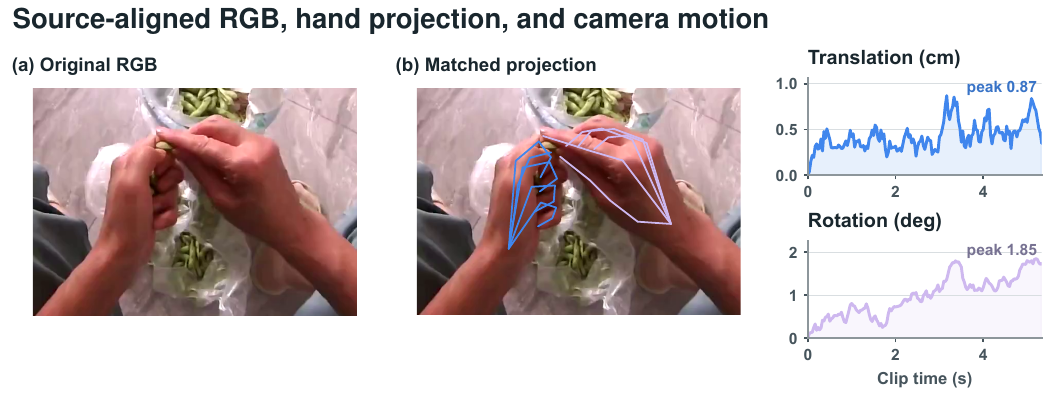}
\caption{\textbf{Source-aligned RGB, hand projection, and camera motion.} Both RGB panels show clip time 0\,s (MCAP-local frame 584). The 161 samples span MCAP-local frames $[584,745)$ and recording-global frames $[8234,8395)$. The traces show translation from the first camera position in centimeters and the relative quaternion angle in degrees.}
\label{fig:supp_alignment}
\end{figure*}

\paragraph{Human camera stability.}
Stable intervals are searched before export using camera trajectories and background registration. VIO checks constrain displacement from the first-frame anchor, local velocity, and sudden disturbances. Background registration measures accumulated image drift and adjacent-frame displacement using spatially distributed features outside expanded hand regions. Table~\ref{tab:supp_camera} gives the long-window settings used for the human reference pipeline. Registration also requires sufficient feature support, inliers, and image coverage. Insufficient registration evidence routes a window to rejection or review.

\begin{table}[t]
\centering\small\setlength{\tabcolsep}{4pt}
\begin{tabularx}{\columnwidth}{@{}Xr@{}}
\toprule
Human camera check & Limit\\
\midrule
Anchor translation / rotation & 3\,cm / $2^\circ$\\
Translation speed, P95 / peak & 0.10 / 0.20\,m\,s$^{-1}$\\
Angular speed, P95 / peak & 8 / 15$^\circ$\,s$^{-1}$\\
RGB anchor displacement & 2\% of diagonal\\
RGB adjacent-frame displacement & 0.4\% of diagonal\\
\bottomrule
\end{tabularx}
\caption{Human long-window camera settings. Anchor checks measure displacement from the first frame; speed and RGB checks detect local disturbance. Robot episodes use the checks appropriate to their camera configuration.}
\label{tab:supp_camera}
\end{table}

\paragraph{Hands and image quality.}
Geometry and RGB jointly establish that participating hands are sufficiently large, in frame, and legible. Participation includes a stationary hand that holds or supports the target. Image checks assess first-frame clarity, sustained blur, exposure, clipping, and local hand detail. Motion salience and aesthetic ranking prioritize useful manipulation among eligible windows. These decisions retain the underlying numeric evidence and configuration version for later inspection.

\paragraph{Robot visibility and exposure.}
HandsOn screening checks end-effector and target visibility, task coverage, and agreement between the instruction, caption, and rendered action. TI5 screening first examines the complete episode and its visible result. The mild branch for dim recordings applies the fixed 8-bit lookup table
\begin{equation}
 I_{\rm out}=\operatorname{round}\!\left[255\left(I_{\rm in}/255\right)^{0.78}\right].
\end{equation}
The encoded result is checked for clipped detail, excessive exposure, flicker, and noise. Frame order and timing are preserved, and the unprocessed head video remains linked to the processed reference.

\paragraph{Window extraction and duplicate control.}
Human clips follow $L=80n+1$ frames, favoring longer eligible intervals. This convention supports rollouts that reuse the previous segment's last frame. Robot evaluation references retain complete native episode lengths; model-facing sampling is recorded through a time mapping. Intervals use the half-open convention $[s,e)$ with $L=e-s$. Container-local and recording-global indices are retained together, so overlap checks span MCAP boundaries on a shared recording clock. Deduplication checks frame intervals, recording and episode IDs, and visual duplicates. Export verification independently decodes the product and checks frame count, order, first-frame correspondence, dimensions, timing, file integrity, and annotation alignment.

\subsection{Evaluation Set Selection}
\label{app:data_curation:selection}
After the checks in Appendix~\ref{app:data_curation:filtering}, manual selection examines reference footage for persistent blur, abnormal exposure, recording or encoding flicker, and occlusion or framing that obscures the critical operation.
Task-relevant contact, hiding, and reappearance remain part of the interaction.
Selection uses the reference videos and their annotations.

Selected clips have a clear primary operation, identifiable hand--object relations, and visible behavior corresponding to the task description.
An observable stage of a longer activity can form an evaluation case when its action and outcome are sufficiently clear.

Evidence requirements follow metric applicability.
Physical and state-change cases expose the initial condition, interaction, and result; retention cases provide a check after state establishment.
Memory cases show the same local region before hiding and after reappearance, while action and event cases provide a judgeable window.
Each case supplies the evidence needed by its applicable metrics.

Among eligible clips, selection considers coverage of embodiments, operations, objects, and scenes. Counts vary across the 22 task categories.
Human cases cover flexible materials, tools, mechanisms, and household objects.
Simulation retains all 54 native task identifiers with two references each; real dexterous-hand cases cover seven recorded identifiers and variations in hand use, object position, and action sequence.
The final manifest links case identifiers, sources, time intervals, task labels, and evaluation windows for review.
Task-category coverage is reported in Appendix~\ref{app:statistics:taxonomy}; duplicate control follows Appendix~\ref{app:data_curation:filtering}.

\subsection{Annotation Production and Release Records}
\label{app:data_curation:records}

\paragraph{Global and interval descriptions.}
The local-VLM interface produces an English global task description and successive interval descriptions. The global description uses five sampled frames. The implemented interval schedule uses $J=\max(3,\lceil L/240\rceil)$ intervals; each interval contributes five frame indices spanning its boundaries. Records retain the sampled clip indices, corresponding source indices, text, model identifier, prompt/configuration signature, and review status. These interval captions provide temporal language coverage. Event and state annotations add their own task-specific windows and questions.

\paragraph{Operation assignment.}
Operation annotation selects a dominant family/task category from the registered vocabulary and records the visual evidence used for the decision. Secondary assignments capture distinct additional operations. Task instructions provide the intended behavior, while descriptions and labels follow the visible execution. Validation checks the family/task category hierarchy and primary-label consistency for all 500 unique identifiers. Caption revisions synchronize the JSONL text with the linked caption and clip records.

\begin{figure*}[!t]
\centering\includegraphics[width=.91\textwidth]{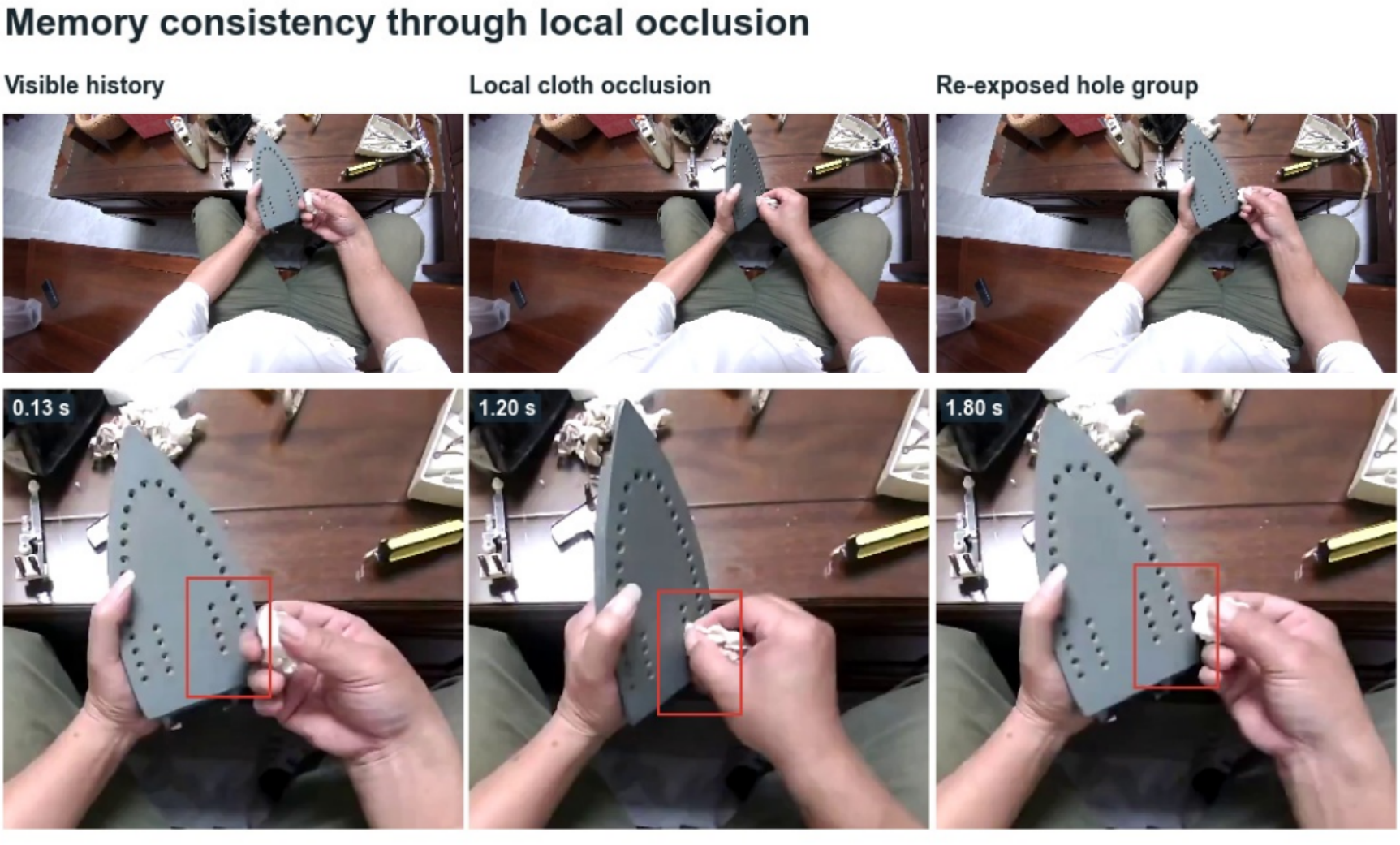}
\caption{\textbf{From a reference sequence to an auditable memory question.} Case \texttt{f2d1997c} shows the soleplate at 0.13, 1.20, and 1.80\,s. The upper row shows scene context and the lower row enlarges the marked hole group before, during, and after cloth occlusion. The annotated history, hidden, and return windows are 0.0--0.4, 0.8--1.6, and 1.8--1.9667\,s. The reference answer is yes: the lower-right outer-hole layout remains consistent relative to the inner row. Viewpoint, object position, and wiping-cloth motion are permitted changes.}
\label{fig:supp_memory_reference}
\end{figure*}

\paragraph{Evidence-linked questions.}
A question record binds an entity, one or more temporal windows, the property to inspect, an expected answer, and allowed changes. Figure~\ref{fig:supp_memory_reference} illustrates this structure using a human wiping reference. The same object region is visible in the history and check windows, while the wiping cloth covers it during the specified hidden interval. The criterion concerns the hole layout relative to the inner row; viewpoint, object position, and cloth motion are permitted changes.

\paragraph{A complete release record.}
The manifest joins video and first-frame paths, frame count and rate, source intervals, annotation paths, captions, operation labels, and checksums. Human NPZ files retain the per-frame skeleton, projection, VIO, calibration, timestamps, and validity arrays. Robot records link native state/action files, episode metadata, camera views, schemas, and source-file inventories. All paths are relative to the release root. These links connect every derived sample to its original measurements and source-specific access terms.

\paragraph{Recoverable processing.}
Screening evidence, export validation, semantic review, and publication status are separate record fields. Completed technical results are retained across interrupted jobs, and outputs are registered after validation. This structure supports retries and subsequent annotation layers while preserving the source-to-clip mapping and the provenance of each decision.

\section{Unified Control}
\label{app:unified_control}
This section details the interface normalization used in Sec.~\ref{sec:annotations}. The common representation specifies hand identity, landmark order, image correspondence, and time correspondence. Source-specific geometry and additional model conditions remain explicit. Dataset acquisition and clip selection are described in Appendix~\ref{app:data_curation}; numerical inference settings belong to Appendix~\ref{app:baselines}.

\subsection{Embodiment Representations}
\label{app:unified_control:embodiments}
The three embodiments provide different upstream signals. Human references use available vision-derived hand geometry and associated camera information. The evaluated human-hand package describes the geometry as an estimate derived from stereo vision. Separate reconstruction tools serve the individual source pipelines.

For simulated dexterous hands, episode-specific observed joint states, the robot model, and camera metadata provide the inputs for forward kinematics and projection. Forward kinematics uses the recorded realized configuration. Real-hand references use recorded hand joint angles, end-effector poses, and synchronization information with a nominal kinematic model and estimated mounting/camera transforms. The projection accuracy depends on the nominal kinematics and estimated transforms.

The common evaluation asset contains image coordinates $\mathbf U\in\mathbb R^{T\times2\times21\times2}$, a time axis, and hand-validity information. The two hand slots are ordered left then right. Each slot retains its anatomical side and its validity flag. $T$ denotes the number of geometry samples and need not equal the decoded video frame count. Continuous geometry, meshes, and camera assets needed for generation are retained in the corresponding preparation pipeline; the compact two-dimensional evaluation assets alone do not replace those inputs.

\subsection{Geometric Correspondence}
\label{app:unified_control:geometry}
Table~\ref{tab:control_landmarks} specifies the order in the reference catalog. Index~0 is the human wrist or the robot palm anchor. Each remaining group contains four landmarks along one finger. For robots, these are named base, middle, distal, and tip in the released convention. The shared indexing expresses a structural correspondence; the human wrist and robot palm anchor are not anatomically identical points, and a robot landmark need not coincide with a native motor joint.

\begin{table}[t]
\centering
\small
\setlength{\tabcolsep}{4pt}
\begin{tabularx}{\linewidth}{@{}lXX@{}}
\toprule
Indices & Human-hand names & Robot-hand names \\
\midrule
0 & Wrist & Palm anchor \\
1--4 & Thumb 1, 2, 3, tip & Thumb base, middle, distal, tip \\
5--8 & Index 1, 2, 3, tip & Index base, middle, distal, tip \\
9--12 & Middle 1, 2, 3, tip & Middle base, middle, distal, tip \\
13--16 & Ring 1, 2, 3, tip & Ring base, middle, distal, tip \\
17--20 & Little 1, 2, 3, tip & Little base, middle, distal, tip \\
\bottomrule
\end{tabularx}
\caption{\textbf{Verified landmark ordering.} The reference catalog uses \texttt{pinky} for the little finger. Native joint-state indices and model-specific skeletal drawing conventions are separate mappings.}
\label{tab:control_landmarks}
\end{table}

Conceptually, robot landmarks are obtained by evaluating the source robot's kinematics and transforming them into the camera frame. Projection then gives
\begin{equation}
\mathbf u_{t,h,j}=\phi\!\left(\pi_{\theta}(\mathbf X^{\mathrm{cam}}_{t,h,j})\right),
\end{equation}
where $\pi_{\theta}$ is the declared camera projection and $\phi$ is the recorded image transform. Human three-dimensional references use their own source camera convention before the same image-correspondence step. For an adapter consuming camera-space three-dimensional coordinates, its unit and axis convention must match those coordinates; image-space coordinates are expressed in pixels of the associated image version. A two-dimensional projection is not inverted to invent missing three-dimensional structure.

Validity is carried separately from coordinates. The compact evaluation assets contain a hand-validity array for all 500 references and a positive-depth array for 499; the remaining reference provides two-dimensional geometry only and is interpreted through its declared fields. Non-finite coordinates remain invalid. Hand validity, landmark depth, and image bounds are checked separately. Occlusion is a visibility condition.

\subsection{Temporal and Camera Alignment}
\label{app:unified_control:alignment}
Geometry samples, rendered conditions, and images are associated with the same source interval. A preparation step selects corresponding source times or frame indices, then applies the image and camera transforms required by the target interface. A resized or rectified first frame therefore requires the corresponding transformed controls and camera parameters. Image dimensions alone do not specify a crop, padding operation, distortion correction, or camera change. In particular, a pinhole-only consumer of fisheye data requires an explicitly recorded rectification rather than reuse of the original image parameters.

The input format supports timestamp-based correspondence and, where available, reference-frame-index binding. The compact catalog preserves explicit source-frame indices for the 70 real-hand references. These indices distinguish the video presentation clock from acquisition timestamps. Subsampling or changing playback speed must update geometry and annotation windows together; a nominal output frame rate does not establish this correspondence. Multi-window generation likewise uses the controls and source interval assigned to each window, with the recorded initial image or preceding generated image required by that adapter.

At evaluation time, the declared time map converts reference times into generated-video times, and the spatial map converts reference coordinates into the generated viewport. Explicit mappings take precedence. The ordinary-input fallback assumes a full-frame resize and a complete-clip uniform speed change, using the first and last decoded timestamps. This fallback does not recover an omitted action, a trimmed interval, or a nonlinear timing change. Appendix~\ref{app:protocol} specifies the evaluation consequences of these assumptions.

\begin{figure*}[!t]
\centering
\includegraphics[width=0.94\textwidth]{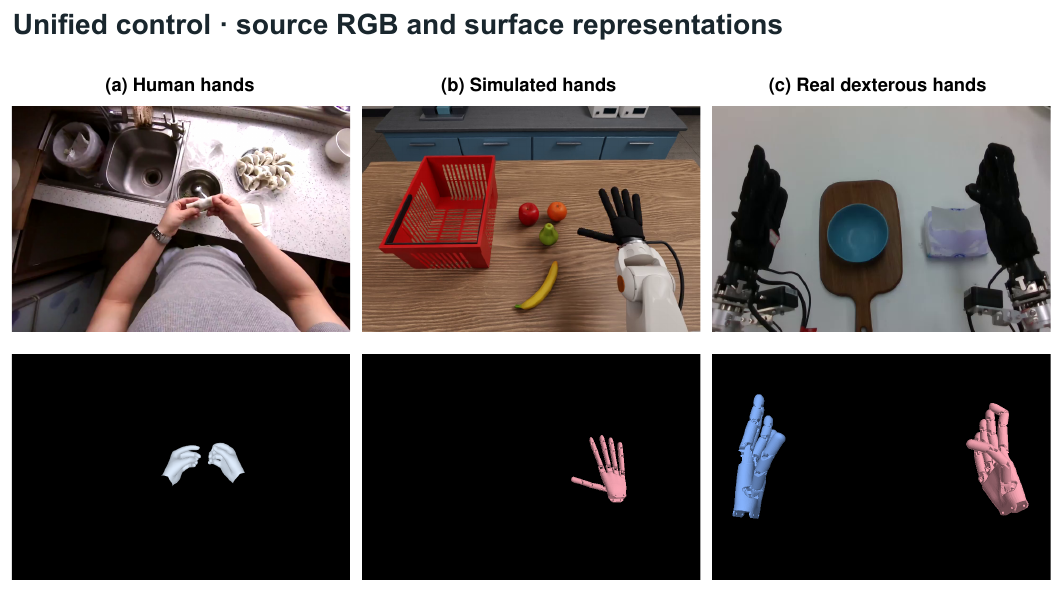}
\caption{\textbf{Source RGB and corresponding surface controls.} The columns show (a) human hands, (b) simulated hands, and (c) real dexterous hands. Each example pairs the source RGB above with its corresponding surface control below.}
\label{fig:control_alignment}
\end{figure*}

\subsection{Model-Specific Adapters}
\label{app:unified_control:adapters}
The recorded generation entry points accept a starting RGB image, task text, and their inference configuration. The four general I2V baselines (CogVideoX1.5, HunyuanVideo 1.5, Wan2.1, and Wan2.2) use image and text conditioning. The controlled-baseline interfaces consume different representations, summarized in Table~\ref{tab:control_adapters}. The prepared control tensors follow each model interface. Appendix~\ref{app:baselines} reports the recorded run settings.

\begin{table*}[t]
\centering
\small
\setlength{\tabcolsep}{5pt}
\renewcommand{\arraystretch}{1.08}
\begin{tabularx}{\textwidth}{@{}lXX@{}}
\toprule
Interface & Preparation inputs beyond starting RGB/text & Additional representation consumed by the model \\
\midrule
PAM~\cite{gao2026pam} & Continuous hand geometry and matching image projection & Skeletal RGB control video in the PAM drawing convention. \\
JointControlVideo~\cite{zhang2026jointcontrol} & Camera-space hand landmarks and matching pinhole intrinsics & Joint array with 42 left/right landmarks per frame and an intrinsic matrix. \\
RynnWorld-Teleop~\cite{zhao2026rynnworldteleop} & Continuous landmarks, hand identity, and matching projection & Skeletal RGB video with its required background and left/right coloring. \\
DWM~\cite{kim2026dwm} & Time-varying surface meshes and projection information & Dynamic surface rendering plus a static video formed from the starting image. \\
Hand2World~\cite{wang2026hand2world} & Meshes, camera information, and the scene starting image & X-ray mesh control video, camera sequence, and starting scene image. \\
EgoHOI~\cite{li2026egohoi} & Meshes, cameras, and the manipulated object's initial mask & Rendered mesh frames, camera information, and the object RGB image obtained by masking the starting frame. \\
EgoSim~\cite{hao2026egosim} & Landmarks and the declared scene/background preparation & Skeletal video, scene RGB video, scene-mask video, and starting image. \\
\bottomrule
\end{tabularx}
\caption{\textbf{Model input formats in the recorded preparation interfaces.} Meshes are additional assets, not a renaming of the 21 landmarks. Robot mesh renderings use robot assets/kinematics. Image-derived masks or reconstructed human meshes retain their estimated status. Sampling and run configurations are given in Appendix~\ref{app:baselines}.}
\label{tab:control_adapters}
\end{table*}

For example, a human reference can supply skeletal controls from its estimated hand geometry, whereas a simulated reference obtains the corresponding controls from its episode state and camera. A real-hand reference uses its synchronized hand/end-effector records and nominal geometric model. The resulting sequence is then encoded for the target interface. A mesh-conditioned method additionally requires a time-varying surface representation; landmarks alone are insufficient. Likewise, the manipulated-object mask used by EgoHOI is a preparation input, and the model consumes its masked RGB image rather than a semantic evaluation answer.

Scene preparation is also method- and embodiment-dependent. The recorded EgoSim human pipeline includes depth/camera-based scene preparation, while the robot pipeline uses a declared static-background/mask variant with dynamic skeletal controls. Structured event, state, and memory questions belong to the scoring protocol.

\paragraph{Adapter checks.}
We audit 24 prepared model inputs: JointControlVideo on four cases per embodiment, and Hand2World and EgoSim on two cases per embodiment. At the first, middle, and last control frames, projecting the reference landmarks reproduces the prepared control coordinates with a maximum residual of 0.15\,px, and the video--control index offset is zero in every sample. Swapping the left and right hand slots raises the mean residual to at least 89.9\,px, which confirms the hand assignment.

For real dexterous hands, the camera-space landmarks consumed by JointControlVideo depend on the nominal kinematic model and the estimated camera transform. We therefore prepared a refined real-robot geometry and compared it with the controls used in the main evaluation. Across the 69 real-hand references with three-dimensional geometry, the per-case maximum projected difference has a median of 4.7\,px (range 3.3--95.9\,px). A paired ablation regenerates JointControlVideo on eight real-hand cases with both geometries, sharing the first frame, text, windows, seed, and scoring protocol. Under seed~0, refined geometry changes Pose Control by $+0.038$ (95\% CI $[-0.021, 0.078]$; 7 of 8 cases improve) and Event Adherence by $-0.125$ ($[-0.275, 0.000]$). Evolution Correctness changes by $+0.115$, and Action Adherence, State Progression, State Regulation, and Trajectory Accuracy change by at most 0.062 in magnitude. The Pose Control and Event Adherence intervals include zero, so the reported results retain the original controls.

\section{Task Annotations and Verification}
\label{app:annotations}
The evaluation annotations specify what can be inspected in a reference interaction, the target entity or local region, and the temporal evidence supporting that inspection. They extend the corpus captions described in Appendix~\ref{app:data_curation} with task-specific actions, outcomes, state criteria, and memory checks. They are distinct from the human preference judgments in Appendix~\ref{app:human_evaluation}.

\subsection{Annotation Schema}
\label{app:annotations:schema}
Each case associates its reference video and geometry with a task description, per-metric applicability, and structured annotations. The compact catalog has seven top-level fields: case identifier, embodiment category, reference video information, geometry, applicability, annotations, and task description. Within \texttt{annotations}, the semantic branch stores entities, actions, events, causal dimensions, and named time windows; evaluation views bind state specifications and protocols; supplementary branches bind the event and memory additions. Detailed records may be inline or referenced by external JSON files. These are alternative storage forms of the same evaluation inputs.

\begin{table*}[t]
\centering
\small
\setlength{\tabcolsep}{5pt}
\renewcommand{\arraystretch}{1.08}
\begin{tabularx}{\textwidth}{@{}lXX@{}}
\toprule
Annotation & Required meaning & Evaluation use \\
\midrule
Entity and region & Object/tool identity, neutral locator, relevant physical surface or part, and scope constraints & Defines which entity or local feature is observed; does not assert success. \\
Action & Requested manipulation with one or more named time windows & Action Adherence; executing a motion does not establish its outcome. \\
Event & A finite visible occurrence and its permitted evidence window & Event Adherence; distinguishes a result from an attempted action. \\
State change & Initial/process/result windows, checkpoint identity, minimum and full completion conditions, and process checks & State Progression and Evolution Correctness. \\
State regulation & Newly established state, establishment evidence, later check windows, and allowed changes or supported reversals & Tests retention of that newly acquired state. \\
Memory & Same object region, history/hidden/return windows, stable properties, return opportunities, and property checklist/codebook & Memory Consistency after a genuine local loss of inspectability. \\
Applicability & Applicable, not applicable, or pending for each metric and relevant unit & Defines the reference population before generated outputs are scored. \\
\bottomrule
\end{tabularx}
\caption{\textbf{Task annotation schema.} A case may contain multiple units and contribute to several metrics. Window endpoints are expressed in reference-video seconds and transformed by the declared prediction-time mapping. Metric formulas are given in Appendix~\ref{app:metrics}.}
\label{tab:annotation_schema}
\end{table*}

State criteria separate partial progress from a fully attained result. For instance, lifting an object can satisfy a minimum condition while deposit and release define full completion. A state-regulation unit additionally identifies when a \emph{new} state is established and when it can subsequently be checked. An object being present throughout a clip does not establish a new state. A later change is permitted only under the unit's declared rules; temporary occlusion or camera motion alone does not demonstrate a deliberate reversal.

Memory annotations identify a physical region and stable properties that can be independently inspected before and after an interval of local unobservability. A turn to a different face is not automatically a return to the original face. Viewpoint changes, manipulation of other parts, and allowed deformations are listed explicitly. The execution checklist asks the same single-window questions with the same categorical codebook in history and return evidence, without supplying the expected historical feature value to either observation. Unknown, clearly absent, and a valid count of zero have different meanings. The annotation's reference criterion defines the intended comparison; it is not a substitute for observing the generated video.

Applicability is defined from reference evidence. In the 500-case catalog, actions cover all cases; events cover 175. State progression and evolution correctness each cover 209 cases; state regulation covers 120; memory consistency covers 22. These are case counts, not counts of annotation files or guaranteed successful model observations. The release contains 503 external detailed JSON files because one case may reference several specifications, protocols, or checklists.

\subsection{Annotation and Review Process}
\label{app:annotations:review}
The evaluation annotations were prepared in several model-assisted batches. The general sequence is reference-evidence preparation, an initial structured annotation, a separate visual review, resolution of disagreements, and validation of the exported records. Review evidence includes clip overviews, time-labeled frames, and denser frames or local crops when an outcome, release, or reappearance is ambiguous. Initial captions provide context. The reviewer checks state transitions against the frames.

For the human references, a visual annotation pass produced candidate entities, questions, windows, and applicability. A separate Qwen3.5-9B pass reviewed the visual evidence independently of the initial answers and captions. Visual adjudication then revisited disagreements; later batches also permitted separately instantiated visual agents as the review route. The state and memory revisions used time-labeled reference evidence and separate cross-review to refine minimum/full criteria, establishment windows, target regions, and return opportunities. Robot references and the later event and memory annotations followed the same review sequence.

Verification checks whether the specified entity is identifiable, the window contains the required evidence, the claimed outcome is visible, and the allowed-change rules match the interaction. An ambiguous final result is not promoted to successful completion. A continuously visible feature is not introduced as a hidden-return memory test. Unsupported opportunities are excluded or left pending. Export checks cover entity/window references, time bounds, asset binding, and the correspondence between the catalog and detailed records.

Human reviewers then check every released case against its reference frames, covering the task description, the annotated object states, and the evidence windows used by each applicable metric (Figure~\ref{fig:data_pipeline}(d)). Cases with an unclear outcome or an unsupported window are revised or removed before release.

\subsection{Annotated Examples}
\label{app:annotations:examples}
The following examples are extracted from the reference catalog and its bound specifications. Their times describe reference evidence, not manually inferred boundaries for generated videos. Short identifiers are shown for readability; full identifiers and source bindings are included in the released annotation files.

\paragraph{Action, state change, and a later containment check.}
Real-hand case \texttt{real\_04762576} transfers red food pieces from a blue bowl into a dark-green container using a white spoon, followed by a leaf transfer. Its action window spans the 48.3667\,s reference. The red-food event is assigned 0--31.5\,s; the leaf event has its own 36--48.3567\,s window. This separates two outcomes within one interaction.

For the red-food state unit, the before, process, and after windows are 0--1, 14--30, and 30.5--31.5\,s. The minimum criterion requires at least one initially visible red piece to leave the bowl; moving an empty spoon is insufficient. Full completion requires the pieces to be visibly deposited in the destination container. The linked regulation unit checks establishment at 32--32.6\,s and containment at 33.5--37\,s, after the spoon has moved away. It tracks the deposited pieces rather than the spoon's presence. Settling inside the container is allowed; a clearly supported deliberate removal terminates the corresponding containment obligation.

\begin{figure*}[!t]
\centering
\includegraphics[width=0.91\textwidth]{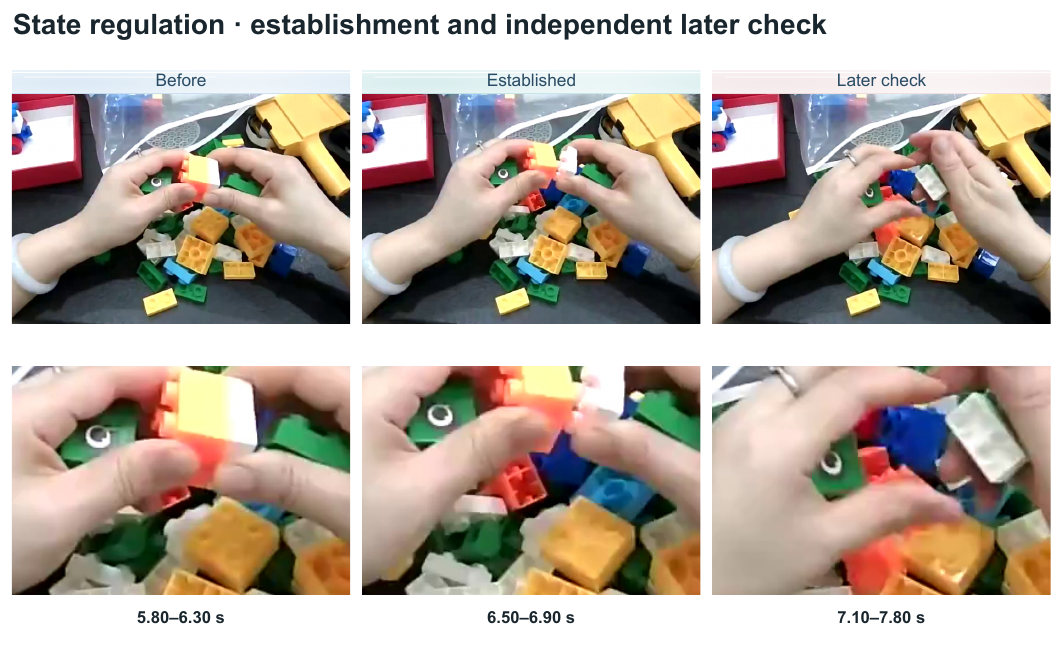}
\caption{\textbf{Establishment and independent later check.} Case \texttt{ef471799} shows the assembly before separation, during establishment, and at a later check. The displayed frames are 175, 196, and 218, sampled at 5.83, 6.53, and 7.27\,s. The detached subassembly remains separate at the later check. Printed intervals give the annotation windows and gaps are unassigned.}
\label{fig:annotation_timeline}
\end{figure*}

\paragraph{A local separation state with an independent later check.}
Human-hand case \texttt{ef471799} provides a compact state-regulation example (Fig.~\ref{fig:annotation_timeline}). The target is the small green-and-white component and its relation to the red-and-green main assembly. The before interval is 5.8--6.3\,s; establishment is 6.5--6.9\,s; the later check is 7.1--7.8\,s. Establishment requires visible detachment and separate holding, rather than a small opening between still-connected parts. The later question asks whether this newly separated component stays apart without unsupported reattachment. Independent rigid rotation and further disassembly are allowed. A visibly deliberate reattachment ends the separation requirement. This case is selected for state regulation, while State Progression and Evolution Correctness are marked not applicable in the catalog; its establishment checkpoint supports the regulation gate rather than adding an unselected metric.

\paragraph{Memory concerns the same local feature after occlusion.}
Human-hand case \texttt{f2d1997c}, shown in Fig.~\ref{fig:supp_memory_reference}, wipes a detached triangular iron soleplate. Its action window is 0--5.3667\,s. The memory unit selects a local hole group near the physical outer edge and its relation to the neighboring inner row. History, hidden, and return windows are 0--0.4, 0.8--1.6, and 1.8--1.9667\,s. The checklist inspects hole shape, coarse arrangement, relation to the inner group, and coarse spacing. The same physical edge and tip/tail references define the target despite changes in screen position. Cloth motion and viewpoint changes are allowed; cleaning success is not evaluated. The short return window and coarse categorical properties limit what this example can resolve: changes within a spacing category or an unidentifiable target region cannot be treated as verified fine-detail preservation.

\begin{figure*}[!t]
\centering
\includegraphics[width=0.91\textwidth]{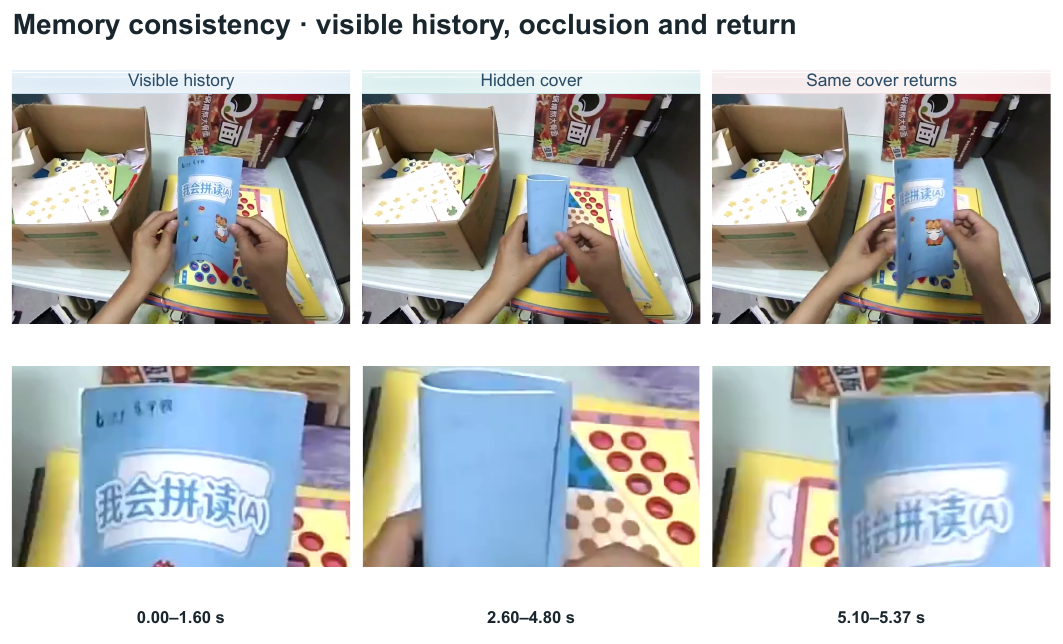}
\caption{\textbf{A memory question after local occlusion.} The printed cover is visible, rolled inward, and visible again in case \texttt{fcd4e468}. Compare the cover layout after it becomes uninspectable. The annotated history, hidden, and return windows are 0--1.6, 2.6--4.8, and 5.1--5.3667\,s. Panel spacing does not encode time.}
\label{fig:supp_memory_booklet}
\end{figure*}

The booklet example in Figure~\ref{fig:supp_memory_booklet} tests the same history--occlusion--return sequence on a printed cover. Rolling the cover inward removes the print from view. The return window checks the layout after it becomes visible again.

\section{Dataset Statistics}
\label{app:statistics}
\subsection{Corpus and Evaluation Set}
\label{app:statistics:composition}
\begin{figure*}[!tp]
\centering\includegraphics[width=0.72\textwidth,trim=0 18 0 0,clip]{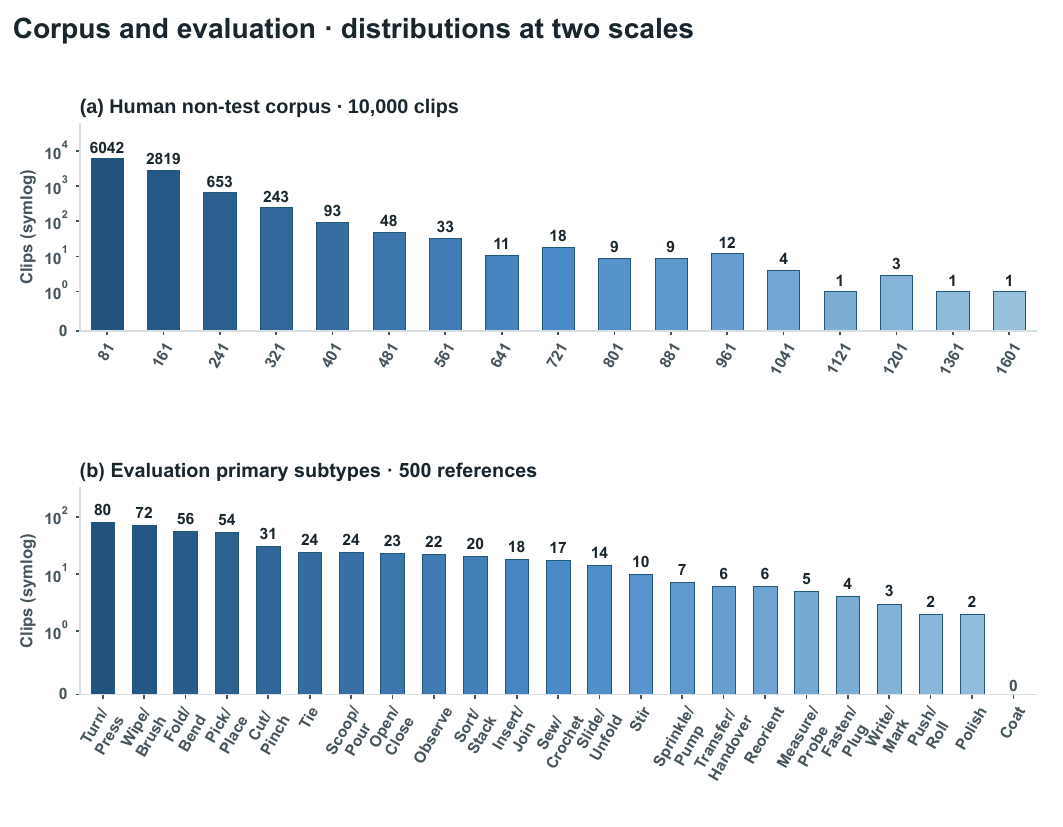}
\caption{\textbf{Non-test corpus and evaluation references.} The upper panel gives frame-tier counts for the 10,000 human non-test clips. The lower panel gives primary operation-subtype counts for the separate inventory of 500 evaluation references.}
\label{fig:supp_composition}
\par\vspace{5pt}
{\raggedright\noindent Human non-test clips use discrete frame-count tiers. Figure~\ref{fig:supp_ego10k_lengths} shows the frequency of each tier within the 10,000 human clips.\par}
\vspace{6pt}
\centering
\includegraphics[width=0.72\textwidth]{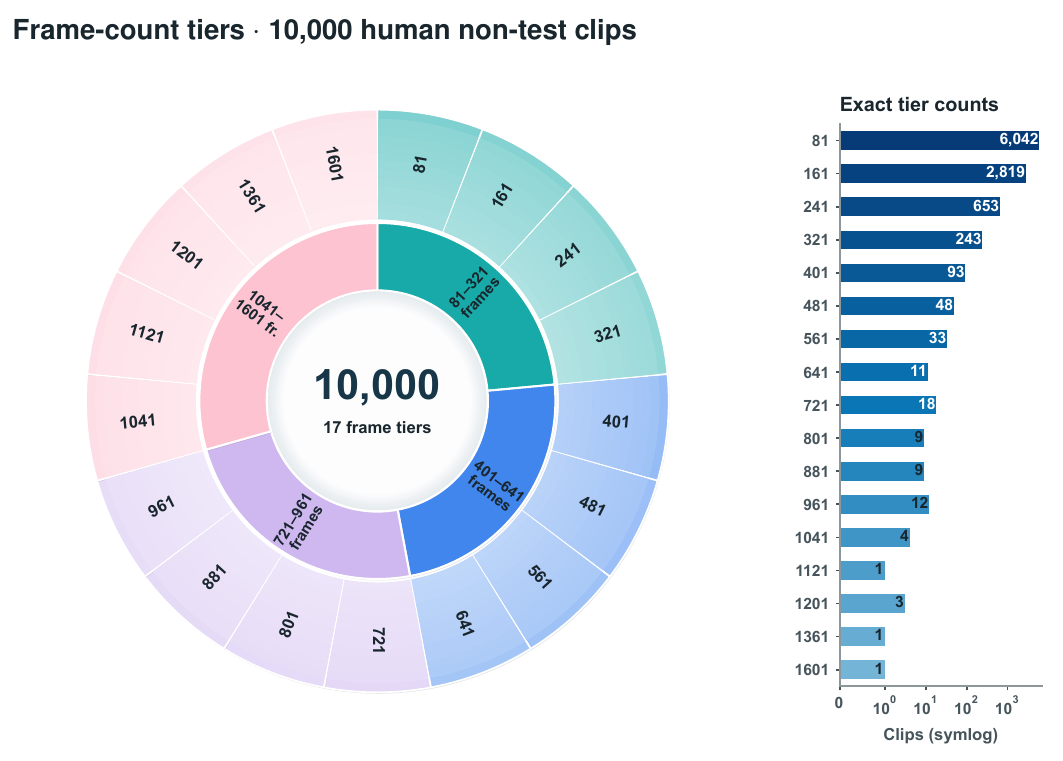}
\caption{\textbf{Frame-count tiers in the 10,000 human non-test clips.} The tier map groups the 17 observed frame counts. Equal-angle sectors identify tiers and the adjacent bars give exact clip counts. The inner bands group frame lengths and the outer tiers retain each observed frame count.}
\label{fig:supp_ego10k_lengths}
\end{figure*}

The selected pool contains 20,500 clips, partitioned into 20,000 non-test clips and 500 evaluation references. The non-test corpus contains 10,000 human, 6,000 simulated, and 4,000 real dexterous-hand clips, with proportions of 50\%, 30\%, and 20\%. The evaluation set contains 322 human, 108 simulated, and 70 real dexterous-hand references. Figure~\ref{fig:supp_composition} shows the frame tiers of the human non-test clips and the primary operation counts of the evaluation references. Each evaluation case is counted once; auxiliary views and multiple annotation units remain associated with that case.

\noindent\begin{minipage}{\columnwidth}
\centering\small\setlength{\tabcolsep}{2pt}
\begin{tabular}{lrrrr}
\toprule
Reference source & Clips & Time (min) & Median (s) & Range (s)\\
\midrule
Human & 322 & 55.47 & 8.03 & 5.37--37.37\\
Simulated & 108 & 28.31 & 13.93 & 2.80--38.83\\
Real robot & 70 & 32.72 & 26.75 & 8.00--50.20\\
\midrule
Total & 500 & 116.50 & & \\
\bottomrule
\end{tabular}
\hwbinlinecaption{table}{Frozen evaluation-reference statistics. Duration equals frame count divided by frame rate, including the final frame's display interval. All 500 references are packaged at 30\,fps.}
\label{tab:supp_duration}
\end{minipage}\par\vspace{6pt}

The evaluation references total 116.50 minutes (1.94 hours). The human duration interquartile range is 8.03--10.70\,s; the corresponding ranges are 8.30--20.81\,s for simulation and 15.40--38.76\,s for real robots. The different profiles reflect human continuous-window selection and complete robot episode references.

\begin{figure*}[!tp]
\centering
\includegraphics[width=0.76\textwidth]{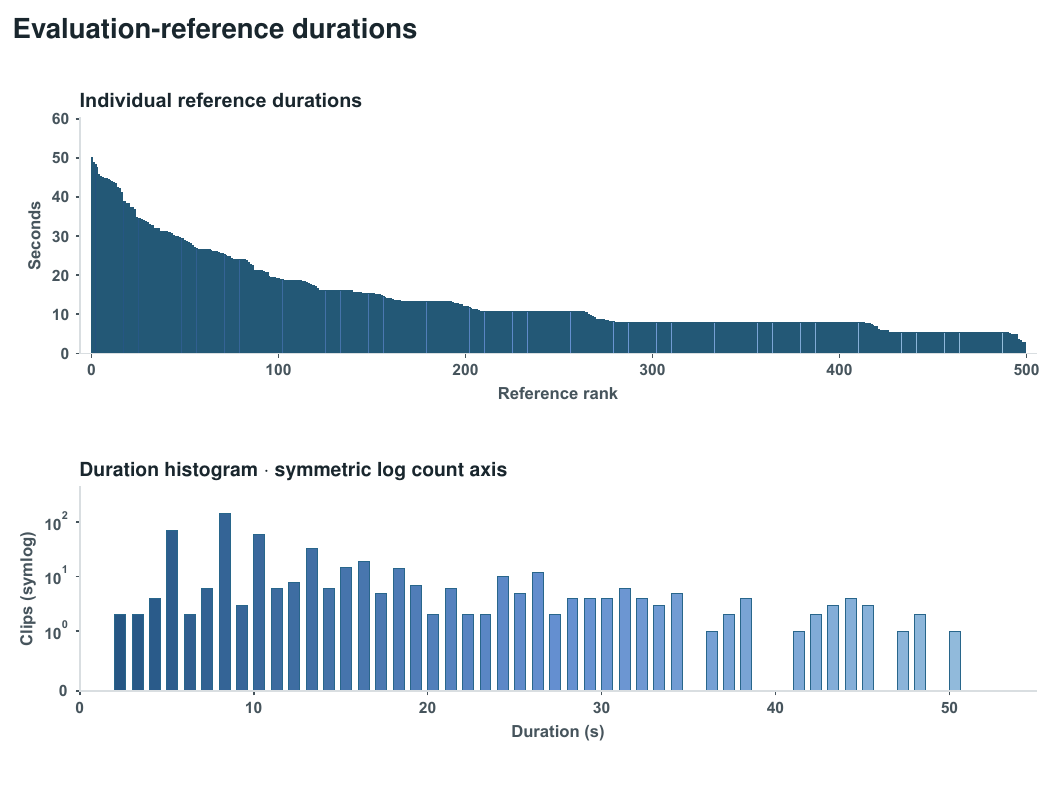}
\caption{\textbf{Evaluation-reference duration.} Each duration is frame count divided by frame rate. The upper panel ranks the 500 references by duration, and the lower histogram shows the distribution on a symmetric logarithmic count axis.}
\label{fig:supp_eval500_duration}
\par\vspace{5pt}
{\raggedright\noindent Scene descriptions are available for 2,281 of the human non-test clips. Figure~\ref{fig:supp_ego10k_diversity} shows the known descriptions and the unlabeled portion.\par}
\vspace{6pt}
\centering
\includegraphics[width=0.72\textwidth]{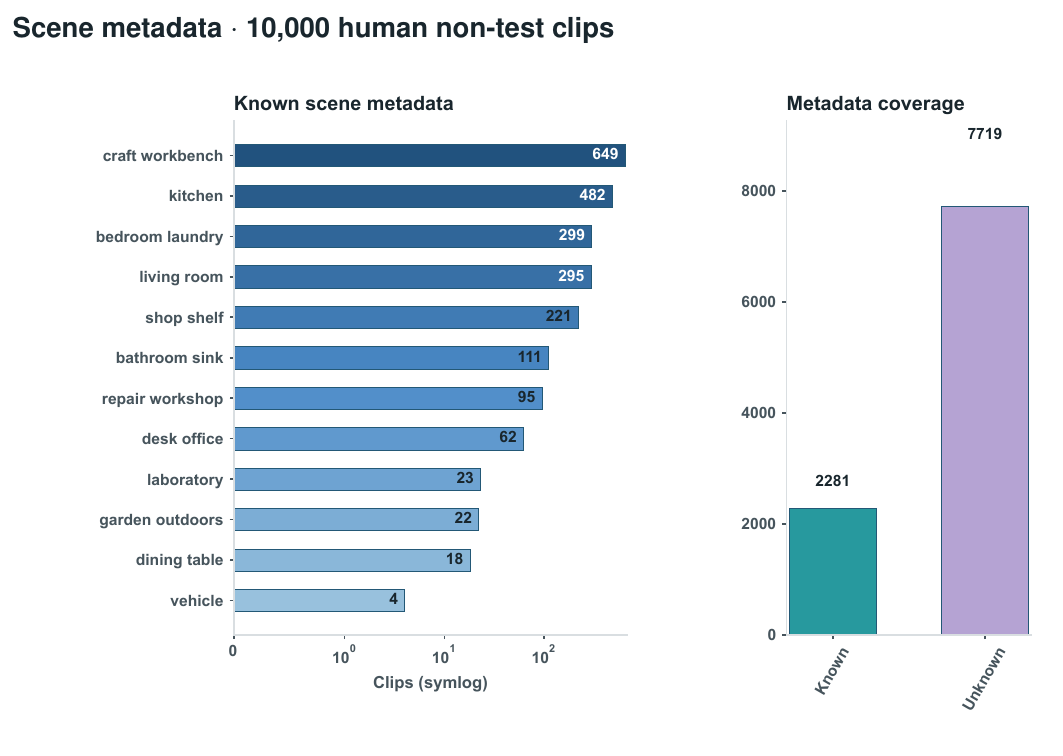}
\caption{\textbf{Scene metadata coverage in the human non-test clips.} The left panel breaks down known scene strings. The right panel shows coverage across the 10,000 clips: 2,281 known and 7,719 unknown.}
\label{fig:supp_ego10k_diversity}
\end{figure*}

Figure~\ref{fig:supp_eval500_duration} shows the ranked lengths and duration histogram of the 500 evaluation references. The table gives per-embodiment duration summaries.

All 322 human references follow $80n+1$: 161 frames (61 cases), 241 (133), 321 (55), 401 (27), 481 (17), 561 (8), 641 (5), 721 (6), 801 (4), 881 (1), 961 (2), 1,041 (1), and 1,121 (2). There are 128 human cases with at least 321 frames. Robot reference durations follow their native episode boundaries. Model generation lengths and time mappings are defined separately by the corresponding inference configuration.

\subsection{Task Taxonomy}
\label{app:statistics:taxonomy}
\begin{figure*}[!tp]
\centering\includegraphics[width=0.67\textwidth,trim=0 36 0 0,clip]{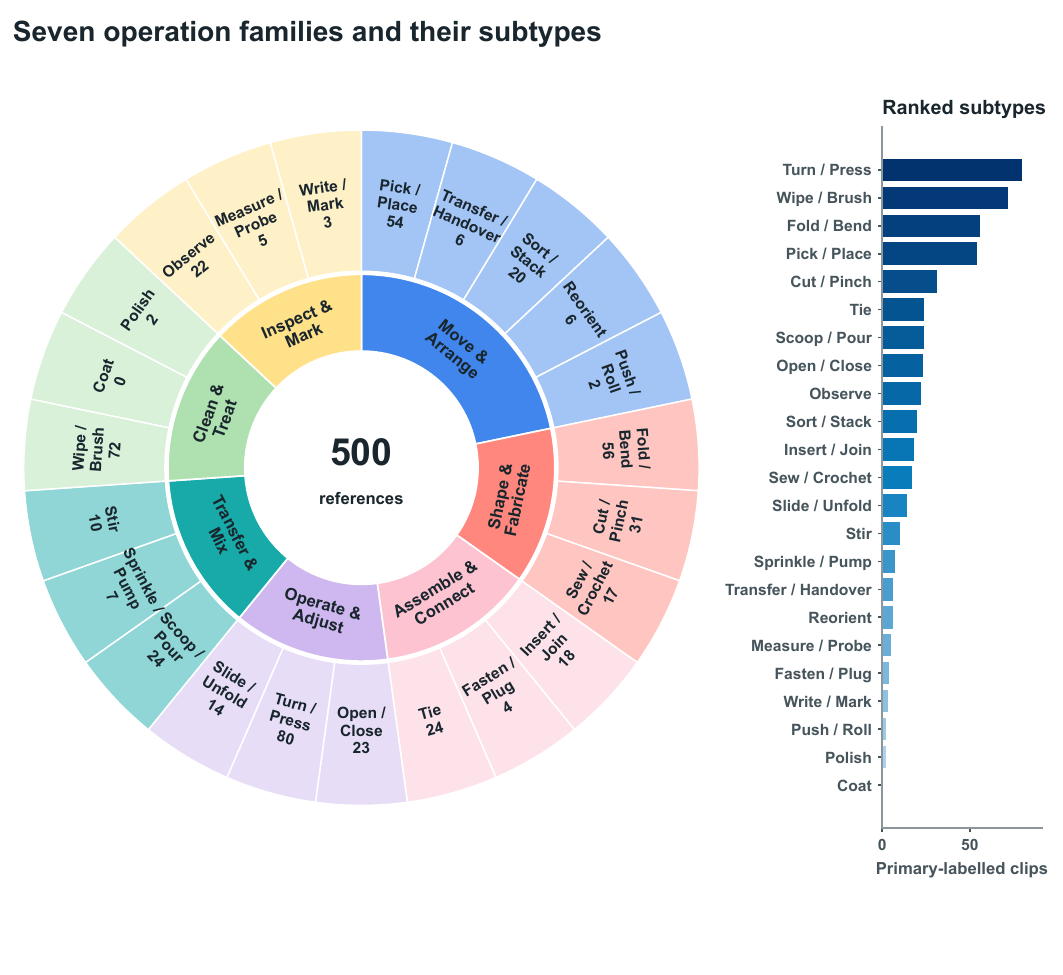}
\caption{\textbf{Seven operation families and their subtypes.} The wheel organizes the taxonomy, with equal angular width per subtype. Leaf counts and the right-hand ranking use the 500 primary assignments. Coat is defined and has zero assignments.}
\label{fig:supp_task_tree}
\par\vspace{5pt}
{\raggedright\noindent The source mix varies by family (Figure~\ref{fig:supp_taxonomy}). Human references contribute the selected shaping and textile examples, while simulation broadens mechanism and object-handling coverage. Real-robot references emphasize object handling, material transfer, and surface interaction. HandsOn contributes 54 native task identifiers, each with two references. The seven TI5 identifiers contribute ingredient transfer (10), bottle lifting (10), tissue pulling (10), salt dispensing (10), bowl wiping (18), spoon grasping/transfer (7), and stirring (5). Native task identifiers and semantic operation labels tabulate coverage at different levels.\par}
\vspace{6pt}
\centering\includegraphics[width=0.62\textwidth,trim=0 14 0 0,clip]{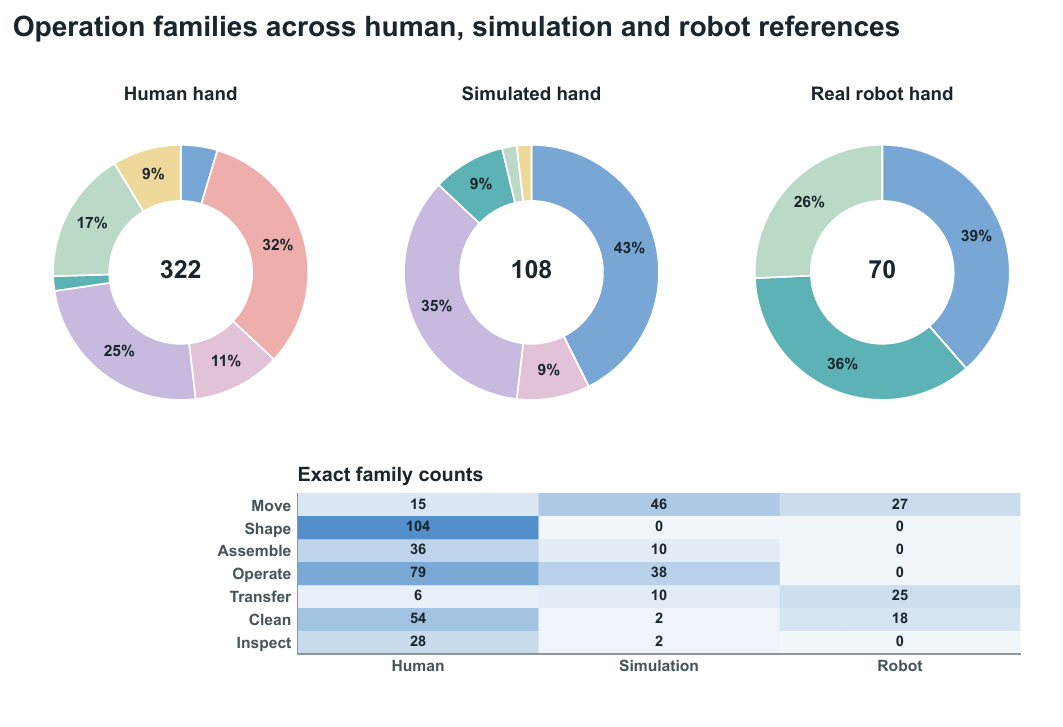}
\caption{\textbf{Operation families by embodiment.} Each ring shows family shares within its source. The matrix gives exact counts for 322 human, 108 simulated, and 70 real-robot references.}
\label{fig:supp_taxonomy}
\end{figure*}

One primary family and task category identify the dominant visible operation in each reference. Primary counts sum to 500. The hierarchy in Figure~\ref{fig:supp_task_tree} contains 22 base task category definitions and one evidence-backed extension, Push / Roll. Coat has zero selected primary examples, so 22 task categories are represented in the evaluation set. Secondary labels capture 83 additional operation assignments in 79 cases.

\paragraph{Family definitions.}
Move \& Arrange covers whole-object transport, exchange, grouping, and reorientation. Shape \& Fabricate covers changes to material shape and textile structure. Assemble \& Connect covers engagement of interfaces, fasteners, plugs, and retaining cords. Operate \& Adjust covers mechanisms and controls. Transfer \& Mix covers loose contents, dispensing, and mixing. Clean \& Treat covers surface-treatment strokes. Inspect \& Mark covers object presentation, probing, measurement, and physical marks.

\paragraph{Assignment rules.}
Operation labels describe the observed manipulation, with completion assessed by event and state questions. A pinch is assigned to Cut / Pinch when it changes material shape; ordinary object grasping falls under object handling. Transporting a container and transferring its contents receive the labels appropriate to the visible operation. Slash-separated names denote task category groups. Additional labels record distinct, task-salient operations alongside the primary assignment.

\begin{figure*}[!tp]
\centering
\includegraphics[width=0.87\textwidth]{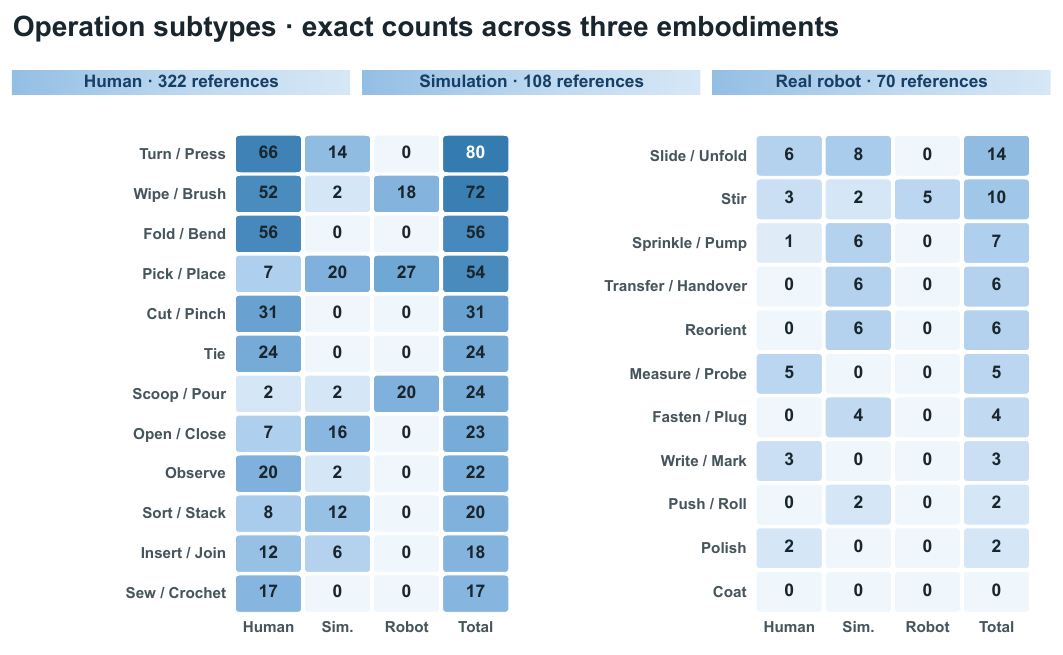}
\caption{\textbf{Primary operation subtypes by embodiment.} Counts are from the frozen 500-reference evaluation inventory. All 23 subtype definitions are shown, including the zero count for Coat. Equal-size cells use a $\log(1+\mathrm{count})$ color scale.}
\label{fig:supp_subtype_sources}
\par\vspace{5pt}
{\raggedright\noindent Figure~\ref{fig:supp_objects} pairs object cutouts with examples of the scene view used to judge the interaction.\par}
\vspace{6pt}
\centering
\includegraphics[width=0.93\textwidth]{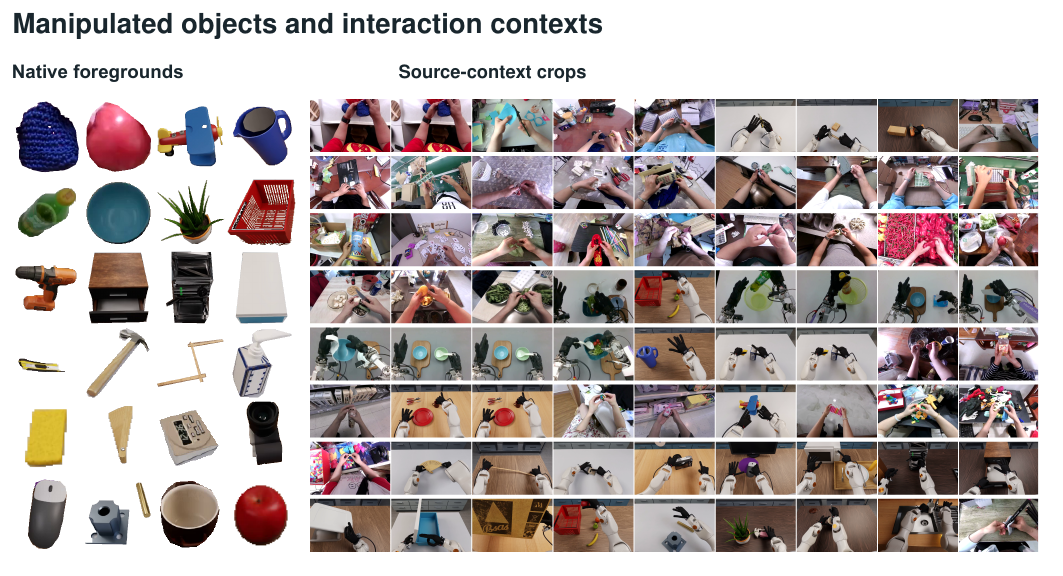}
\caption{\textbf{Manipulated objects and interaction contexts.} The left panel shows native foreground assets. The right panel aligns source-context crops in a common viewport, with uniform row and column spacing.}
\label{fig:supp_objects}
\end{figure*}

Figure~\ref{fig:supp_subtype_sources} provides the full subtype-by-embodiment counts behind the family-level shares.

\subsection{Interaction Attributes and Evaluation Subsets}
\label{app:statistics:subsets}
The native annotation layers retain object descriptions, scene and interaction information, hand configuration, camera views, and source metadata. Participation includes both active manipulation and static support. Object descriptions preserve the visible role of a container, wrapper, tool, or loose contents. Validity masks identify the usable geometric evidence, while annotated windows specify visibility and reappearance opportunities. Quantitative task and subset counts use the standardized primary labels and reference applicability flags.

\begin{figure*}[!tp]
\centering\includegraphics[width=0.72\textwidth]{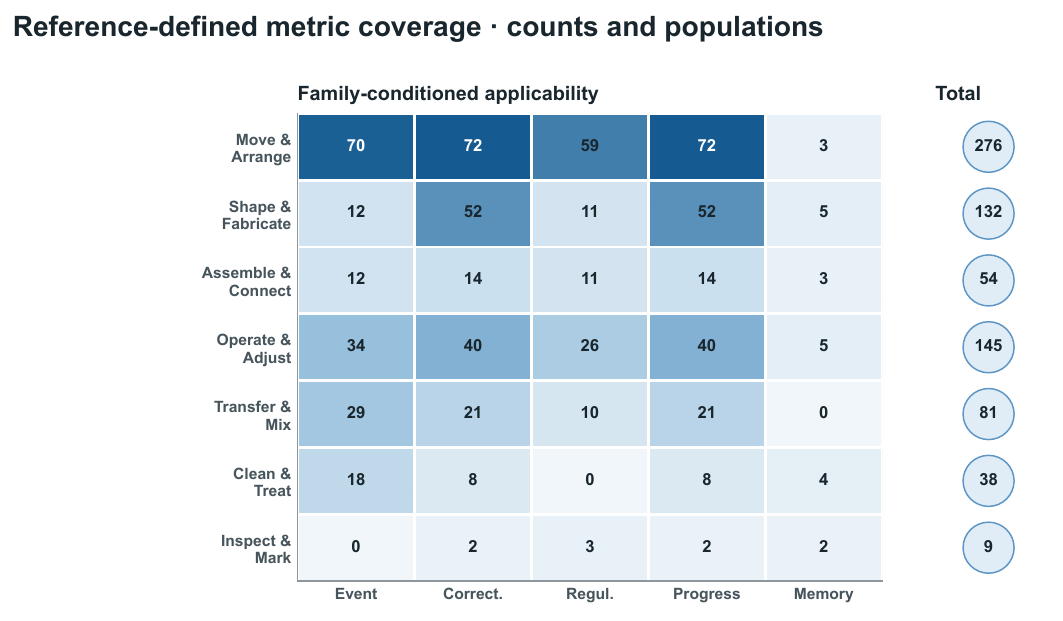}
\caption{\textbf{Reference-defined metric coverage.} Cells count applicable references within each operation family. The five metric totals are 175, 209, 120, 209, and 22. The circles sum metric memberships by family, so a reference may contribute more than once. The counts describe reference applicability before scoring.}
\label{fig:supp_applicability}
\par\vspace{5pt}
{\raggedright\noindent Figure~\ref{fig:supp_metric_subtypes} shows how the same applicability masks intersect the primary operation subtypes. A single reference can support several diagnostic questions when it contains the required evidence for each.\par}
\vspace{6pt}
\centering
\includegraphics[width=0.78\textwidth]{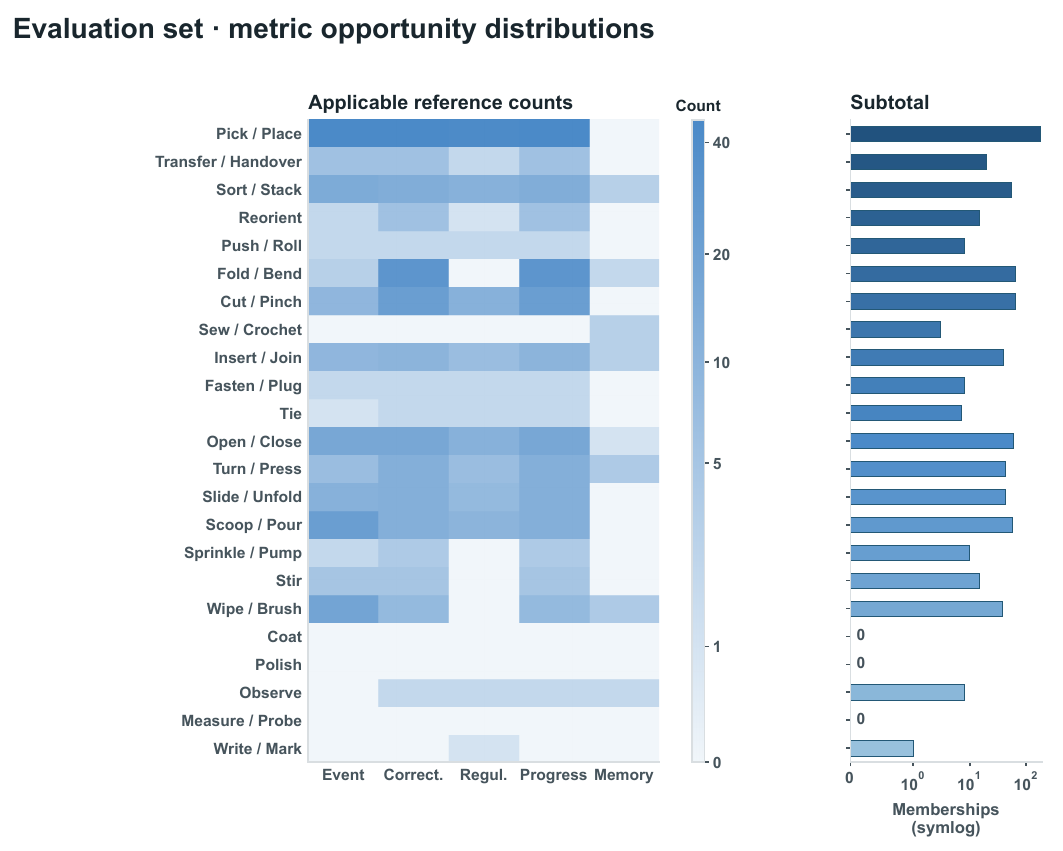}
\caption{\textbf{Evaluation opportunities by operation subtype.} The matrix counts applicable references for five metrics in the frozen 500-reference inventory. The right panel totals metric memberships by subtype. A reference may contribute to more than one metric. The columns are Event Adherence, Evolution Correctness, State Regulation, State Progression, and Memory Consistency.}
\label{fig:supp_metric_subtypes}
\end{figure*}

The evaluation catalog defines 175 Event Adherence cases (25 human, 80 simulated, 70 real robot); 209 Evolution Correctness and State Progression cases each (67, 92, 50); 120 State Regulation cases (20, 63, 37); and 22 Memory Consistency cases (19, 3, 0). These memberships overlap. Reported model comparisons additionally apply the common-valid execution and protocol rules in Section~\ref{sec:aggregation}.

The subsets address distinct temporal evidence. State Progression assesses attainment of visible milestones. Evolution Correctness assesses the transformation process with a progress gate. State Regulation checks a later state after establishment, using explicitly permitted reversals. Memory Consistency compares a historical property of the same object region across a gap or occlusion. Figure~\ref{fig:supp_memory_reference} illustrates the entity, windows, question, expected answer, and allowed changes used to define such a memory opportunity. Together, these subsets preserve intermediate progress, established outcomes, and historical detail in the evaluation. The annotated windows leave measurable gaps for these checks. Across the 22 memory cases, the median interval from the end of the history window to the start of the first return window is 3.50\,s (range 0.80--10.10\,s), the hidden window lasts a median of 1.95\,s (0.45--9.20\,s), and the return window lasts a median of 0.55\,s (0.17--2.80\,s). The 120 state-regulation cases contain 124 retention units, with a median gap of 0.50\,s (0.03--7.00\,s) between the end of establishment and the start of the later check.

\section{Evaluation Protocol and Efficiency}
\label{app:protocol}

This section specifies input binding, evidence reuse, and reporting for the seventeen metrics in Appendix~\ref{app:metrics}. The evaluation separates a reference-defined requirement from observations extracted from a generated video. The reference task determines which criteria apply.

\subsection{Input Preparation and Alignment}
\label{app:protocol:preparation}

\paragraph{Task binding.}
Each submitted video is associated with a unique task identifier and model identifier. The task record supplies the reference video, embodiment, task description, geometric controls, metric applicability, and semantic evaluation windows. Preparation freezes the selected inputs and configuration, records video fingerprints, and creates metric-specific observation and scoring tasks. Duplicate model--task--metric--protocol records are rejected during report collection. Reference and prediction observations use separate roles, including when a reference video is evaluated as a prediction for a sanity check.

\paragraph{Time correspondence.}
A declared mapping $\tau:t_{\mathrm{ref}}\mapsto t_{\mathrm{pred}}$ binds reference timestamps and annotation windows to the generated sequence. Identity, affine, and piecewise mappings are supported. For an ordinary input without an explicit mapping, the input protocol assumes that the complete reference interval is represented with a uniform change in playback speed. After normalizing each video's first decoded timestamp to zero, it uses
\begin{equation}
 \tau(t)=t\,\frac{T_{\mathrm{pred}}}{T_{\mathrm{ref}}},
\end{equation}
where $T_{\mathrm{ref}}$ and $T_{\mathrm{pred}}$ are the spans between the first and last decoded frame timestamps. Both sequences must supply at least two finite, strictly increasing timestamps. Equal spans give identity mapping, independently of nominal frame rate or frame count. Trimmed, paused, or nonlinearly retimed outputs require an explicit time map.

\paragraph{Image coordinates.}
An explicit spatial mapping takes precedence over automatic preparation. Without one, the input protocol assumes full-frame resizing. If the reference and prediction widths are $W_r$ and $W_p$, pixel-center coordinates follow
\begin{equation}
 x_p=(x_r+\tfrac{1}{2})W_p/W_r-\tfrac{1}{2},
\end{equation}
with the analogous transformation for height. Equal dimensions give identity. Image dimensions do not identify cropping, padding, undistortion, or a changed viewpoint. Those operations require a recorded transform and, where necessary, camera-space geometry and calibration. The declared transform is applied to the reference geometry used by the metric, while generated-video observations are extracted from the generated images themselves.

\paragraph{Sampling.}
The shared image pipeline defaults to sixteen frames, selected uniformly over decoded frame indices, including the endpoints; shorter videos use their available frames. Reference frames and geometry are bound to these samples through the same declared time mapping. Frame-indexed geometry can instead bind directly to decoded reference frames when its index sequence and reference-video fingerprint are provided. Metric-specific samplers, including the photometric and JEPA branches, retain their own sampling rules as detailed in Appendix~\ref{app:metrics}. Semantic windows are mapped before evidence extraction, and short windows use their actual available frames.

\subsection{Scoring and Reporting}
\label{app:protocol:reporting}

\begin{figure*}[!t]
\centering
\includegraphics[width=0.94\textwidth]{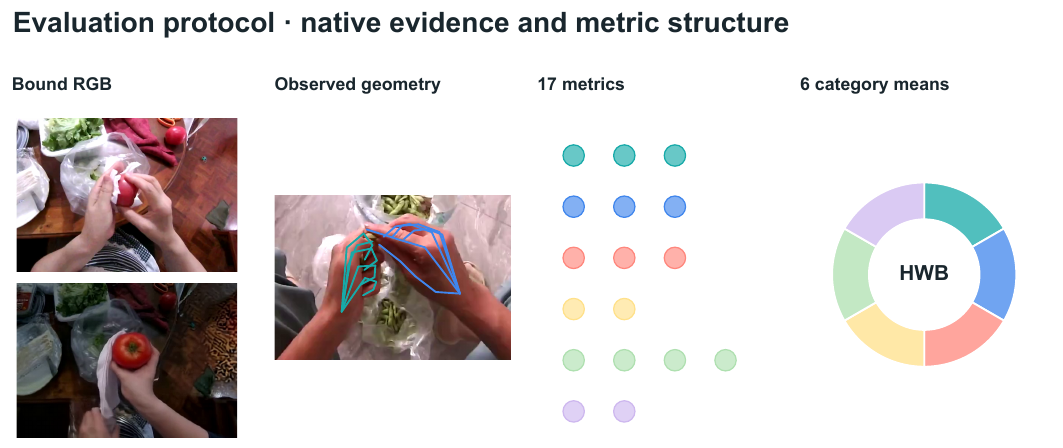}
\caption{\textbf{Evidence and aggregation in the scoring protocol.} Bound RGB shows a reference and its PAM prediction. Observed geometry shows a separate native hand projection. Reference-defined applicability precedes scoring. The six rows of dots show the respective metric counts (3, 3, 3, 2, 4, and 2). The six category means receive equal weight and their average is multiplied by 100 to obtain HWBScore.}
\label{fig:scoring_protocol}
\end{figure*}

\paragraph{Shared evidence.}
Grounding DINO detections support hand-keypoint estimation, hand masks, and relevant hand-localization consumers. RTMPose supplies predicted keypoints, while SAM3 produces box-conditioned masks. Manipulated objects are localized using the configured Qwen3-VL observer. Feature, geometry, and semantic branches then consume their required evidence. Generated and reference regions are observed from their respective images; a missing generated hand or object is not replaced by its reference observation. State and memory branches use reference annotations to specify what to check and extract the corresponding generated evidence at the bound windows.

Evidence reuse is scoped by video content, reference/prediction role, transforms, sampling, relevant annotation content, configured model resources, and observation protocol. For example, changing a memory checklist invalidates the affected memory evidence without requiring unrelated image-quality observations to be recomputed. Model residency, batched detection and feature extraction, and reference-cache reuse reduce repeated work. Batch size may be reduced after a device-allocation failure; such execution changes do not alter the metric's prescribed evaluation population.

\paragraph{Scores and non-scores.}
A numeric zero can be a legitimate poor score or an explicit penalty for missing generated evidence. In Pose Control and Keypoint Coverage, for example, missing hands, low-confidence points, non-finite coordinates, and out-of-frame predictions receive the prescribed zero contribution against a fixed reference-valid population. Structural non-applicability is excluded from that metric's population. Pending annotation decisions, unavailable required references, invalid input bindings, and execution errors must retain their reasons instead of being silently treated as successful observations. JEPA feature records are collection inputs, not per-video numerical scores.

The native reporter preserves these states and constructs common-valid summaries by intersecting eligible task identifiers across the selected runs, separately for each metric, embodiment subset, and metric protocol. The resulting population is identified by the selected runs and case membership, rather than by the nominal benchmark size. A reference sanity-check run can contribute a reference-validity constraint, but is not a competing model.

\paragraph{Common-case reporting.}
A common-case report is defined by a frozen snapshot specifying the compared runs, metric protocol, embodiment subset, export policy, and contributing case IDs. For metric $k$, protocol $p$, and embodiment subset $b$, let $\mathcal{U}_{k,p}^{(b)}$ denote the reference-defined applicable tasks selected by that report. Each compared model uses the same declared case membership for that metric, protocol, and subset. Benchmark size, source-run coverage, and effective metric sample count are distinct quantities because applicability and reference eligibility are metric-specific. Missing-observation penalties are retained under the chosen reporting policy; any exclusions and their reasons are recorded with the effective case IDs and sample count. Multiple applicable event or state units are averaged within a case before equally weighting cases. JEPA is recomputed on the exact corresponding video collection, separately for overall and embodiment reports. The overall result first pools the applicable cases for each metric, then aggregates the category scores.

The comparison contains 500 generated videos for each of the 11 models. The Cases row in Table~\ref{tab:main_results} reports reference-defined metric applicability, with JEPA counts indicating collection membership.

\subsection{Hardware and Runtime}
\label{app:protocol:runtime}
We measure five complete scoring runs, each evaluating one model's 500 videos with the seventeen-metric suite on eight NVIDIA H20 GPUs. The median wall-clock time is 7.96 hours, corresponding to a median of 63.67 allocated GPU-hours (Table~\ref{tab:scoring_cost}). The per-backend measurements in Table~\ref{tab:backend_memory} identify the largest memory requirements: Visual Plausibility peaks at 61.57 GiB allocated and 62.52 GiB reserved. State Progression, Evolution Correctness, and State Regulation share the state-evidence backend.

\begin{table}[t]\centering
\caption{\textbf{Cost of scoring 500 videos.} Ranges cover five model-specific runs; device memory is measured over the whole GPU.}
\label{tab:scoring_cost}
\small\begin{tabularx}{\columnwidth}{@{}Xr@{}}\toprule
Quantity & Measurement \\\midrule
Videos / metrics per run & 500 / 17 \\
Measured runs & 5 \\
GPU configuration & $8\times$ NVIDIA H20 \\
Wall time, median [range] & 7.96 [7.41--7.99] \\
GPU-hours, median [range] & 63.67 [59.25--63.93] \\
Device peak memory (GiB) & 89.25--94.45 \\
Mean sampled utilization & 20.24--25.91\% \\
\bottomrule\end{tabularx}\end{table}

\begin{table*}[t]\centering
\caption{\textbf{Peak GPU memory by evaluation backend.} Allocated and reserved memory are PyTorch peaks in GiB. Shared backends are listed once.}
\label{tab:backend_memory}
\small\begin{tabularx}{\textwidth}{@{}lXrr@{}}\toprule
Backend & Metrics / role & Allocated & Reserved \\\midrule
Visual plausibility & Visual Plausibility & 61.57 & 62.52 \\
State evidence & Progression / Correctness / Regulation & 32.43 & 42.87 \\
Memory evidence & Memory Consistency & 31.31 & 35.57 \\
Background geometry & Photometric Consistency & 22.22 & 28.35 \\
Causal evidence & Causal Plausibility & 21.13 & 22.60 \\
Action evidence & Action Adherence & 21.10 & 22.10 \\
Event evidence & Event Adherence & 20.74 & 22.10 \\
Object localization & Trajectory / shared object observations & 15.22 & 15.76 \\
Entity features & Entity Consistency & 13.43 & 15.76 \\
Hand detection & Shared hand localization & 10.71 & 13.18 \\
Background features & Background Consistency & 4.69 & 4.98 \\
Hand masks & Shared hand segmentation & 4.55 & 4.85 \\
Video features & JEPA Similarity & 2.65 & 2.75 \\
Hand morphology & Hand Morphology & 0.42 & 0.48 \\
\bottomrule\end{tabularx}\end{table*}

\section{Metric Definitions and Aggregation}
\label{app:metrics}
We report seventeen measurements across six dimensions. Each score summarizes the observable evidence defined by its metric. All scores are oriented so that larger is better. Unless specified below, an applicable metric is computed per case and then averaged with equal case weights. Missing predictions remain in the reference-defined evaluation population. Reference availability, task applicability, and execution failures are recorded separately.
The formulas below describe the evaluator used for all reported results.

\subsection{Visual Quality}
\label{app:metrics:vq}
\paragraph{MUSIQ.}
We use the SPAQ-trained MUSIQ image-quality model~\cite{ke2021musiq} with its native preprocessing. The shared image schedule samples up to sixteen uniformly spaced decoded frames, including the endpoints. For image predictions $r_t$, the video score is
\begin{equation}
 s_{\mathrm{MUSIQ}}=\operatorname{clip}_{[0,1]}\!\left(\frac{1}{100|\mathcal T|}\sum_{t\in\mathcal T}r_t\right).
\end{equation}
MUSIQ measures image quality without a reference image.

\paragraph{Visual Plausibility.}
We use WBench's dedicated Qwen3-VL-A3B visual-plausibility reward checkpoint~\cite{ying2026wbench,wbench2026vpweights}, distinct from the general semantic judge. The video is sampled at 2~fps. The fixed prompt asks about overall appearance, motion, common sense, and physical rationality. At the first output position, a softmax over the five rating-token logits gives probabilities $p_j$ for Perfect, Good, Fair, Poor, and Bad. The normalized reward is $\sum_{j=1}^{5}(6-j)p_j/5$. The reward is the expected value of this five-level rating. Its native range is $[0.2,1]$ before the common clipping operation.

\paragraph{JEPA Similarity.}
We adapt JEDi~\cite{luo2025jedi} using V-JEPA~\cite{bardes2024vjepa} and WorldArena's video sampling~\cite{shang2026worldarena}. Each complete video yields one sixteen-frame, $224\times224$ clip; indices follow the official uniform-selection rule, repeating the final frame if fewer than sixteen frames are available. The encoder is V-JEPA ViT-H/16 with the associated 174-class attentive probe. Let $X=\{x_i\}_{i=1}^{n}$ and $Y=\{y_j\}_{j=1}^{r}$ denote generated and reference feature collections of dimension $d$. With $k(x,y)=(x^\top y/d)^2$, the implemented biased polynomial-MMD distance and positive score are
\begin{align}
 M(A,B)&=\frac{1}{|A||B|}\sum_{a\in A}\sum_{b\in B}k(a,b),\\
 D&=100\big[M(X,X)+M(Y,Y)\nonumber\\
 &\hspace{4em}-2M(X,Y)\big],\\
 s_{\mathrm{JEPA}}&=\exp[-0.4\max(D,0)].
\end{align}
JEPA is a \emph{collection-level} statistic. We recompute it from the exact comparison collection, including each embodiment-specific collection. Per-video records contain features, not individual JEPA scores; neither copying a collection score to every video nor filling these feature records with zero is valid.

\subsection{Hand Fidelity}
\label{app:metrics:hf}
\paragraph{Reference population and detection.}
Grounding DINO~\cite{liu2023groundingdino} supplies embodiment-specific hand detections and RTMPose~\cite{jiang2023rtmpose} estimates twenty-one points per detected hand. Let $\mathcal J_t$ contain the reference points that are valid, finite, and inside the aligned image at sample $t$, preserving anatomical hand side and joint index. Geometrically valid occluded points are included: there is no additional reference visibility filter. A predicted point is recovered only if the corresponding hand side exists, both coordinates and confidence are finite, the point is inside the image, and confidence is at least $0.3$. Denote this indicator by $a_{tj}$.

\paragraph{Pose Control.}
For aligned points $p_{tj}$ and $\hat p_{tj}$ on an image of width $W_t$ and height $H_t$, define
\begin{align}
 u_{tj}&=\begin{cases}
 \max\!\left(1-\frac{\|\hat p_{tj}-p_{tj}\|_2}{0.1\max(W_t,H_t)},0\right),&a_{tj}=1,\\
 0,&a_{tj}=0,
 \end{cases}\\
 s_{\mathrm{pose}}&=\frac{1}{|\mathcal T_+|}\sum_{t\in\mathcal T_+}
                 \frac{1}{|\mathcal J_t|}\sum_{j\in\mathcal J_t}u_{tj},
\end{align}
where $\mathcal T_+=\{t:|\mathcal J_t|>0\}$. An unrecovered prediction contributes zero, including an entirely missing hand. The score first averages points within a frame and then reference-valid frames within the case. Pose Control measures capped image-plane distance.

\paragraph{Keypoint Coverage.}
Coverage measures recovery of the same fixed reference population,
\begin{equation}
 s_{\mathrm{coverage}}=
 \frac{\sum_{t}\sum_{j\in\mathcal J_t}a_{tj}}{\sum_t|\mathcal J_t|}.
\end{equation}
Unlike Pose Control, it pools points across the case rather than averaging frame-level ratios. A confident but misplaced point can contribute to coverage while receiving little pose credit. The native score is undefined for a case with no eligible reference points.

\paragraph{Hand Morphology.}
We use VBench-2.0's hand classifier~\cite{zheng2025vbench2} on shared detector crops. The score is the predicted probability of normal morphology, equivalently one minus the deformation probability. At each reference-valid frame, every expected hand contributes a probability; a missing predicted crop contributes zero. We average hands within the frame and then frames within the case. This human-anatomy classifier applies only to the human-hand subset. Robot cases are marked inapplicable for this metric.

\subsection{Appearance Consistency}
\label{app:metrics:ac}
\paragraph{Entity Consistency.}
The reference defines the hand sides and manipulated-object inventory. Hands and objects are localized in each video's own pixels and segmented using SAM3~\cite{carion2025sam3}. Object localization uses Qwen3-VL~\cite{bai2025qwen3vl}; generated regions do not receive reference boxes or masks as prediction guidance. We encode each mask-isolated entity crop with CLIP ViT-B/32~\cite{radford2021clip}. Pixels outside the mask are filled with RGB $(127,127,127)$; the crop receives ten-percent bounding-box padding, is square-padded with the same fill, and uses native CLIP preprocessing. For entity $e$ on a reference-available time set $\mathcal T_e$,
\begin{equation}
 s_e=\frac{1}{|\mathcal T_e|}\sum_{t\in\mathcal T_e}
 \begin{cases}
 \max(0,\cos(f^R_{et},f^P_{et})),&\text{prediction present},\\
 0,&\text{prediction absent}.
 \end{cases}
\end{equation}
Hands are equally weighted within the hand group; objects use fixed declared weights. Applicable hand and object groups receive equal weight. Reference-unavailable regions are reported as coverage limitations; a required entity with no usable reference region makes the native combined score unavailable. The score measures the visual appearance of automatically matched regions.

\paragraph{Background Consistency.}
We extend VBench's first/previous-frame cosine aggregation~\cite{huang2024vbench} with SAM3 foreground removal and DINOv3 ViT-S/16 features~\cite{simeoni2025dinov3}. Foreground concepts are fixed from reference metadata, and their mask union is filled with the rounded ImageNet mean RGB. We resize to $256\times256$, apply ImageNet normalization, and extract the normalized CLS feature $b_t$. The score is
\begin{equation}
 s_{\mathrm{bg}}=\frac{1}{T-1}\sum_{t=2}^{T}
 \frac{[\cos(b_t,b_{t-1})]_++[\cos(b_t,b_1)]_+}{2},
\end{equation}
where $[v]_+=\max(0,v)$. A branch receives zero if either participating frame has an empty or full foreground union. A nonempty union does not guarantee that every foreground object was removed; concept-detection coverage is retained for inspection.

\paragraph{Photometric Consistency.}
At 3~fps, Depth Anything~3~\cite{lin2025da3} predicts depth and camera parameters using a processing resolution of 504. SAM3 identifies background pixels from the recorded foreground concepts. We back-project valid positive-depth background pixels, select at most 5,000 source points per frame with random seed zero, and reproject them into a target background. Projection removes nonfinite and behind-camera points, out-of-image points, and target foreground; a z-buffer keeps the closest projected source point per target pixel. Target depth must be finite and positive; depth discrepancy is diagnostic rather than an occlusion-rejection threshold. For source $r$ and target $t$, let $E_{r,t}$ be the mean squared RGB discrepancy on these correspondences, with RGB divided by 255. Define
\begin{align}
 h(E)&=(1+\max(E,0)/0.01)^{-1},\\
 s_{\mathrm{photo}}&=\frac{1}{T-1}\sum_{t=2}^{T}
            \frac{h(E_{t-1,t})+h(E_{1,t})}{2}.
\end{align}
A branch with fewer than 100 accepted projected pixels contributes zero; branch weights remain fixed. The score compares colors across generated frames using estimated geometry.

\subsection{Instruction Following}
\label{app:metrics:if}
\paragraph{Action Adherence and Event Adherence.}
We adapt WBench's five-question evaluation~\cite{ying2026wbench}. Each annotated action or event is evaluated using its mapped temporal windows sampled at 3~fps. The five questions test inactivity or absence, recognizable execution, visible completion, key semantic details, and a prohibited anomaly. Their expected answers are respectively No, Yes, Yes, Yes, and No. The fifth action question concerns clearly impossible body motion or interaction; the fifth event question concerns an obviously unrelated object or anomaly. For annotated unit $u$,
\begin{equation}
 s_u=\frac{1}{5}\sum_{q=1}^{5}\mathbf1[\hat a_{uq}=a_{uq}],
 \qquad s_{\mathrm{action/event}}=\frac{1}{U}\sum_{u=1}^{U}s_u.
\end{equation}
Semantic unknown answers receive no credit, while multiple units are equally weighted within the case. The inherited rubric accepts recognizable partial execution for its recognition question and reserves the completion question for a visible outcome. The score can credit recognizable partial execution under this rubric. Event Adherence is restricted to cases with explicit event requirements.

\paragraph{Semantic model and answer handling.}
The local semantic evaluator is Qwen3.5-9B~\cite{qwen2026qwen35}. For actions, events, and causal evaluation, decoding is greedy with thinking disabled. Binary parsing requires exactly one \texttt{Answer: Yes/No} line; an explicit unknown is recorded separately and scores zero. The causal parser accepts one \texttt{Score: 0/1/2/3} line, with explicit unknown mapped to zero. A formatting failure receives at most one retry using identical evidence and criteria, then remains an execution error. State and memory branches use constrained numeric or codebook responses instead of these parsers. The full prompt builders, inherited question wording, grading instructions, and parser rules are released with the evaluation code.

\subsection{Physical Consistency}
\label{app:metrics:pc}
\paragraph{Trajectory Accuracy.}
This is a two-dimensional, ordered geometric comparison of visible manipulated-object mask centroids. Reference masks are transformed onto the generated video's image grid before comparison. For object $e$, let $c^R_{et}$ and $c^P_{et}$ be reference and predicted centroids. An isolated interior missing prediction is linearly interpolated using its two \emph{originally observed} neighbors and actual timestamps; consecutive or endpoint gaps are not filled. Define the tolerance
\begin{equation}
 \tau_e=\max\big(0.1\,\operatorname{diag}(B^R_{e,\mathrm{first}}),
                    0.005\sqrt{W^2+H^2}\big).
\end{equation}
Here $B^R_{e,\mathrm{first}}$ is the first available reference mask's bounding box on the aligned evaluation schedule, which need not be the video's first frame. Write $\Delta c_{et}=c_{e,t+1}-c_{et}$. For every predetermined adjacent pair $(t,t+1)$ whose two reference masks exist, the segment succeeds only if both endpoint errors and the displacement error are at most $\tau_e$:
\begin{align}
 \|c^P_{et}-c^R_{et}\|_2&\leq\tau_e,\\
 \|c^P_{e,t+1}-c^R_{e,t+1}\|_2&\leq\tau_e,\\
 \|\Delta c^P_{et}-\Delta c^R_{et}\|_2&\leq\tau_e.
\end{align}
The object score is successful divided by reference-available segments; remaining missing predictions fail their segments. We combine required objects with fixed weights. We never bridge missing reference endpoints, and report unavailable-reference coverage separately. Mask centroids are neither three-dimensional physical centers nor independently verified object identities. Conditional ADE/FDE diagnostics do not replace the primary missing-penalized score.

\paragraph{Causal Plausibility.}
We adapt WBench's rendering-physics and causal evaluation~\cite{ying2026wbench}, supplying the task description and a 3~fps chronological sample of the evaluated interval. A global assessment $r_0\in\{0,1,2,3\}$ checks solid-body integrity, gravity, motion continuity, permanence, deformation, and visible cause--effect relations. When the reference annotates relevant specialist dimensions, their grades $r_1,\ldots,r_D$ are averaged with the global assessment:
\begin{equation}
 s_{\mathrm{causal}}=\begin{cases}
 r_0/3,&D=0,\\
 (r_0+D^{-1}\sum_{d=1}^D r_d)/6,&D>0.
 \end{cases}
\end{equation}
Only annotated relevant dimensions are queried. Available specialist rubrics cover fluids, collision, surface effects, deformation, environmental forces, reflections, and human motion. The judge rates the plausibility of the visible interaction.

\paragraph{Evolution Correctness.}
We separate visible progress from the plausibility of its physical realization. State Progression below supplies a gate $P$. For the same initially identified manipulated entity, up to two annotated global physical properties receive grades $g\in\{1,\ldots,5\}$, converted by $v(g)=(g-1)/4$. Their prompts specify failure, partial, and passing evidence without assuming that the intended transformation occurred. A fixed overview contains up to twelve uniform samples of the operation interval and, where present, up to four preceding samples. Local checks inspect consecutive native-frame triplets around motion-change peaks within predetermined temporal strata. The local component is the \emph{minimum} normalized triplet grade; the global properties and this local component are then equally averaged to obtain $C$. Finally,
\begin{equation}
 s_{\mathrm{evolution}}=P\,C.
\end{equation}
An unchanged plausible object can receive high raw physical consistency $C$ but no evolution credit when $P=0$. Unknown evidence receives zero at its component. This explicit dependence must be considered when interpreting correlations with State Progression.

\paragraph{State Regulation.}
Every annotated persistence unit specifies an initial window $B$, establishment window $E$, later check windows $L_j$, and permitted reversals. An establishment question first tests whether the \emph{same} object acquired the specified new state relative to $B$; mere object presence or rigid rotation does not suffice. In retention-only cases, an explicit annotation may require the full condition. When progression is also evaluated, a linked checkpoint must have resolved nonzero progress before the establishment question is queried. Failure to establish the state yields zero. Otherwise, each later window receives a grade $1$--$5$ for how much newly gained configuration remains, mapped by $v(g)$. Unknown essential evidence contributes zero. We average check windows within each unit and then units within the case.

The legal-termination protocol distinguishes a supported permitted reversal from an unsupported reset. If an initial retention answer selects the dedicated permitted-reversal code, the evaluator retains the original $B/E/L$ samples and adds twelve predetermined uniform samples from establishment through that check. It then grades preservation up to a visibly supported permitted termination, or through the original check if termination is not demonstrated. This follow-up always remains scored: a permitted action alone neither grants full credit nor removes the unit. An earlier unsupported loss remains penalized. No additional minimum holding duration is invented.

\subsection{Temporal Coherence}
\label{app:metrics:tc}
\paragraph{State Progression.}
Each checkpoint defines a minimum visible change and a full completion condition for the same new local state. The evidence includes the first initial-window image and the ordered union of process/after samples. A three-way constrained question asks whether the minimum is demonstrated: Yes, No, or Cannot Determine. Only Yes triggers the full-condition question. The checkpoint score is
\begin{equation}
 p_j=\begin{cases}
 1,&\begin{gathered}\text{minimum and full conditions}\\\text{demonstrated}\end{gathered},\\
 0.5,&\text{minimum Yes and full No},\\
 0,&\begin{gathered}\text{minimum No or a required}\\\text{judgment Unknown}\end{gathered}.
 \end{cases}
\end{equation}
We average checkpoints equally. A later reopening or repositioning does not erase an earlier visibly demonstrated milestone; persistence is assessed by State Regulation. Invalid execution is retained as an error, separately from a valid unknown-evidence judgment.

\paragraph{Memory Consistency.}
Each reference-defined unit identifies a historical local feature, history/intervening/return windows, allowed changes, and a frozen feature checklist with an intrinsic or topological coordinate definition. An opportunity gate first requires the region to be visible in history, unavailable through intervening occlusion or turning away, and visible again on return. A continuously visible frozen view does not pass. The gate uses up to three history images, two intervening images, and three return images selected deterministically.

After a successful gate, each feature is observed \emph{independently} in its history and return windows using the same frozen codebook. The question names the measured local feature, not merely its carrier object. Code zero means unknown; nonzero absence codes are distinct from unreadability. For feature $f$ and planned return $r$, let $v^H_{fr}$ and $v^R_{fr}$ be the decoded historical and return values, with $\bot$ denoting an unknown observation. Then
\begin{equation}
 m_{fr}=\mathbf1[v^H_{fr}=v^R_{fr}\ne\bot].
\end{equation}
The score averages planned returns within a feature, features within a unit, and units within a case, with equal weights at each level. A failed opportunity gate receives zero; an unknown feature comparison also receives zero while known peers retain credit. Identical known absence values can agree under the explicit codebook. The comparison uses the feature observed in the generated history as its baseline. It applies only to the annotated memory subset.

\paragraph{Numeric grading and reproducibility.}
The state and memory evaluators constrain their final answers to the declared labels or feature-codebook values, preserving the exact selected evidence and prompt text. Code meanings differ by branch: the progression gate uses $1/2/3$ for Yes/No/Unknown, whereas a physical grade uses $0$ for unknown and $1$--$5$ for fulfillment. Each scoring branch uses its own codebook. The state backend supports budgeted reasoning followed by a constrained final label as well as a zero-thinking alternative; the numeric backend defaults to a 4,096-token reasoning budget and seed 1729. Repeated scoring of fixed videos reproduces these judgments exactly (Appendix~\ref{app:diagnostics:stability}).

\subsection{HWBScore and Aggregation}
\label{app:metrics:aggregation}
\paragraph{Measurement states.}
We retain (i) genuine measured zeros, (ii) missing-prediction penalties within a fixed reference population, (iii) unavailable reference evidence, (iv) structural inapplicability, (v) pending annotation, and (vi) execution errors as distinct states. In a missing-as-zero score export, remaining applicable missing scores receive zero under the declared export rule, while their native status and reason are retained alongside the exported value. Structurally inapplicable entries remain empty, and pending applicability remains unresolved. JEPA's per-video feature-only records are collection inputs rather than missing numerical scores and are exempt from zero filling.

\paragraph{Comparison population.}
A common-case report fixes the contributing case IDs jointly for the selected models, metric protocol, and embodiment subset. Native common-valid reports intersect eligible records, whereas missing-as-zero exports apply their declared missing-score rule within the specified population. These reporting policies are distinct; each report specifies its policy, effective case membership, and sample count. Table~\ref{tab:main_results} and Tables~\ref{tab:results_human}--\ref{tab:results_real} use every applicable case for each model; an applicable case without a usable observation contributes zero under the missing-observation rule. Neither a run's video count nor the benchmark size establishes the effective metric sample count. JEPA is recomputed on the exact collection selected by the report. A ground-truth diagnostic run is not a ranked generation model, and its inclusion in a validity intersection is disclosed explicitly.

The comparison contains 500 videos for each of 11 models. Table~\ref{tab:main_results} gives reference-defined applicability counts; the composite combines metric means over their respective declared populations.

\paragraph{Category and overall score.}
Let $\bar s_{m,k}^{(b)}$ be model $m$'s metric score on the metric-specific population declared by the selected report for subset $b$, and let $\mathcal K_g^{(b)}$ be the fixed applicable metrics in category $g$. Define
\begin{align}
 c_{m,g}^{(b)}&=\frac{1}{|\mathcal K_g^{(b)}|}
                     \sum_{k\in\mathcal K_g^{(b)}}\bar s_{m,k}^{(b)},\\
 \mathrm{HWBScore}_m^{(b)}&=\frac{100}{6}\sum_{g=1}^{6}c_{m,g}^{(b)}.
\end{align}
Each category contributes one sixth of HWBScore, with its weight divided evenly among its applicable metrics. Structurally inapplicable metrics are excluded using the same subset-level rule for every model. In the evaluation set, Hand Morphology is human-only and real dexterous-hand cases have no applicable Memory Consistency units. The composite is undefined if any category has no applicable metric or a required score remains unresolved under the reporting policy.

The overall score pools applicable cases across embodiments before category aggregation. We apply the fixed mappings described above and clip scores to $[0,1]$, rejecting nonfinite values. These transforms are independent of the evaluated baselines and do not use min--max scaling across a leaderboard.

\begin{figure*}[!p]
\centering
\includegraphics[width=0.56\textwidth]{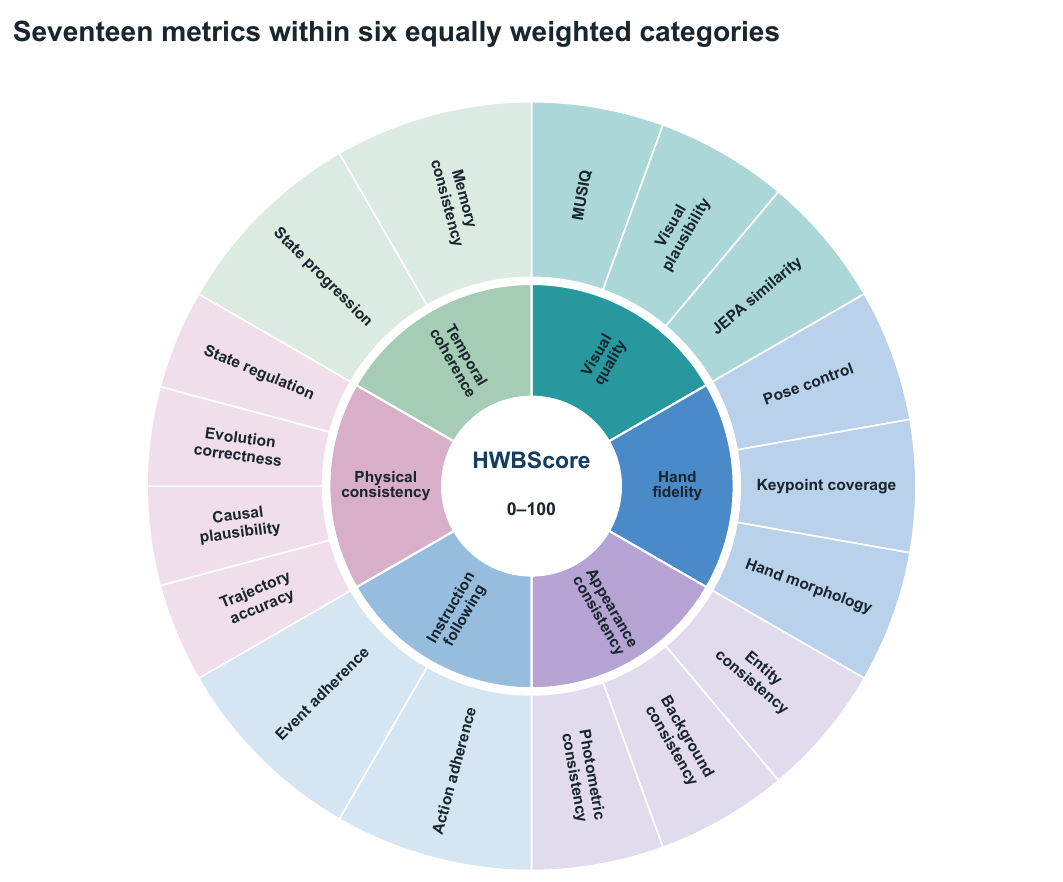}
\caption{\textbf{Seventeen metrics within six categories.} Metric means are combined within category. HWBScore equally weights the six category scores. JEPA is evaluated at the collection level. The outer ring lists the metrics within each category. Sector areas encode the hierarchy and equal category weights.}
\label{fig:supp_metric_map}
\par\smallskip{\raggedright\noindent Case-based metric means use their applicable cases. JEPA is recomputed from the report's selected collection.\par}
\par\vspace{9pt}
\centering
\includegraphics[width=0.70\textwidth]{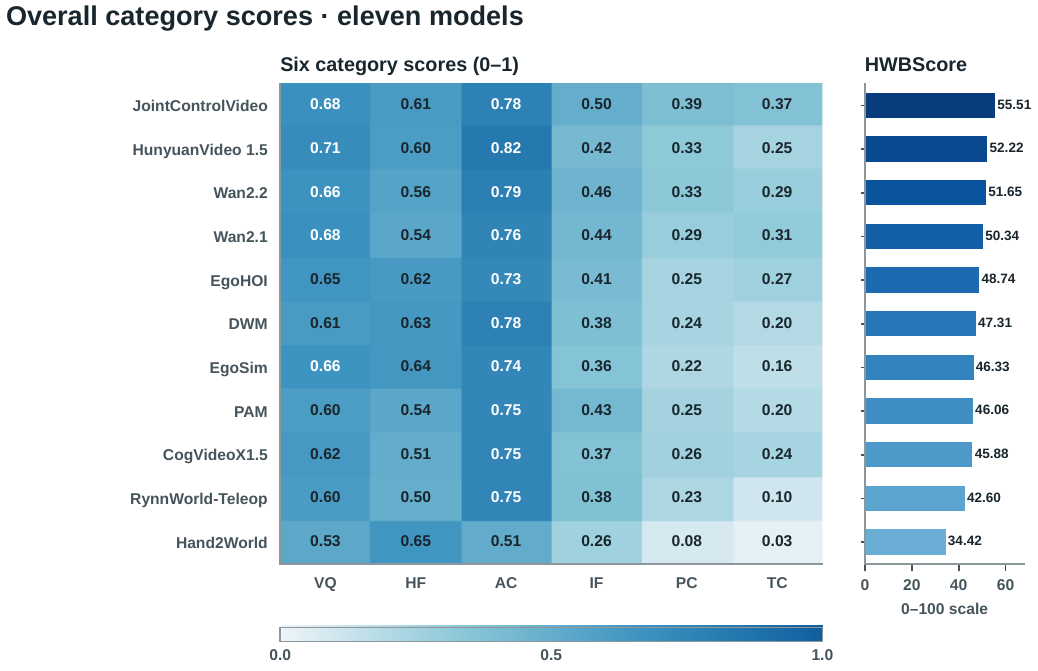}
\caption{\textbf{Category scores of the overall results.} The heatmap shows the six category scores of the eleven models in Table~\ref{tab:main_results}, ordered by HWBScore. The right panel shows the corresponding HWBScore. Appendix~\ref{app:protocol:reporting} defines the reporting protocol.}
\label{fig:supp_results_archive}
\par\smallskip{\raggedright\noindent Tables~\ref{tab:results_human}--\ref{tab:results_real} show the corresponding embodiment results and their scoring populations.\par}
\end{figure*}

\section{Baseline Configurations}
\label{app:baselines}

\subsection{Model Versions and Checkpoints}
\label{app:baselines:versions}
We evaluate the eleven models in Table~\ref{tab:main_results} using released checkpoints and model-specific inference adapters.
Table~\ref{tab:baseline_configs} lists the deployed variants and generation sizes.
Specialized models load their trained conditioning modules together with the corresponding base generator.
PAM uses its Stage-III DexYCB hand-only checkpoint; RynnWorld-Teleop uses the SFT checkpoint with additive control.
Hand2World uses the V0.9 ARCTIC autoregressive release, its two adapters at weight $0.5$ each, and the full VAE.

\subsection{Conditioning Configurations}
\label{app:baselines:conditioning}
Models are evaluated on shared reference tasks using their native conditioning interfaces. Tables~\ref{tab:conditioning_inputs} and~\ref{tab:conditioning_time} specify the information available to each evaluated configuration.
$C$ denotes task text; $A_s$ is the RGB anchor for window $s$; $W_s$ is its target control interval; and $I_0$ is the original first frame. The first anchor is the initial RGB image, and continuations use generated history. Camera parameters used only for rendering are distinguished from explicit generator inputs.
EgoSim's human scene preparation estimates camera/depth information using reference RGB across the complete clip. This provides privileged future-reference access through derived scene conditions; target-window hand or camera controls are specified separately. The generator receives scene renderings and visibility masks. Robot scenes use estimated static-background priors.
\begin{table*}[!tp]\centering
\caption{\textbf{Evaluated checkpoints and generation configurations.} $L$ is the typical window length; $S$ denotes sampling steps. Sizes are width$\times$height.}
\label{tab:baseline_configs}
\begingroup\fontsize{8}{9.5}\selectfont\setlength{\tabcolsep}{3pt}
\begin{tabularx}{\textwidth}{@{}p{2.7cm}>{\raggedright\arraybackslash}Xcrr@{}}\toprule
Model / variant & Checkpoint or weight combination & Size & $L$ & $S$ \\\midrule
CogVideoX1.5 & CogVideoX1.5-5B-I2V & $1360\times768$ & 81 & 50 \\
HunyuanVideo 1.5 & HunyuanVideo-1.5-Diffusers-480p\_i2v & 480p$^{*}$ & 121 & 50 \\
Wan2.1 & Wan2.1-I2V-14B-480P & $832\times480^{*}$ & 81 & 40 \\
Wan2.2 & Wan2.2-TI2V-5B & $1280\times704^{*}$ & 121 & 50 \\
PAM (hand-only) & PAM-w\_hand + CogVideoX-5b-I2V & $720\times480$ & 49 & 50 \\
JointControlVideo & JointControlvideo + Wan2.1-I2V-14B-480P & $832\times480$ & 81 & 50 \\
RynnWorld-Teleop (SFT) & SFT control + Wan2.2-TI2V-5B & $832\times480$ & 81 & 50 \\
DWM & DWM-CogVideoX-Fun-5b-LoRA + CogVideoX-Fun-V1.1-5b-InP & $720\times480$ & 49 & 50 \\
Hand2World (V0.9 AR) & Author base / camera adapter + bidirectional.safetensors + ar\_stage3\_dmd.safetensors & $672\times480$ & 81 & 4 \\
EgoHOI & EgoHOI mp\_rank\_00\_model\_states.pt + Wan2.1-I2V-14B-720P DiT & $480\times480$ & 81 & 50 \\
EgoSim & Released 52-channel DiT: diffusion\_pytorch\_model.safetensors & $832\times480$ & 61 & 50 \\
\bottomrule\end{tabularx}
\par\vspace{3pt}
{\raggedright\footnotesize\noindent $^{*}$HunyuanVideo uses adaptive 480p buckets; Wan specifies an area budget, with aspect ratio determined by the input.\par}
\endgroup\par\vspace{10pt}
\centering
\caption{\textbf{Conditioning inputs for the evaluated checkpoints.} Model identifiers match Table~\ref{tab:baseline_configs}; camera-only projection is performed upstream of the generator.}
\label{tab:conditioning_inputs}
\begingroup\fontsize{8}{9.5}\selectfont\setlength{\tabcolsep}{3pt}\renewcommand{\arraystretch}{1.12}
\begin{tabularx}{\textwidth}{@{}>{\raggedright\arraybackslash}p{2.6cm}cc>{\raggedright\arraybackslash}p{2.05cm}>{\raggedright\arraybackslash}p{2.55cm}>{\raggedright\arraybackslash}p{2.7cm}>{\raggedright\arraybackslash}X@{}}\toprule
Model / variant & Text & Anchor & Extra RGB & Hand signal & Camera signal & Scene / object conditions \\\midrule
CogVideoX1.5 & $C$ & $A_s$ & None & None & None & None \\
HunyuanVideo 1.5 & $C$ & $A_s$ & None & None & None & None \\
Wan2.1 & $C$ & $A_s$ & None & None & None & None \\
Wan2.2 & $C$ & $A_s$ & None & None & None & None \\
PAM (hand-only) & $C$ & $A_s$ & None & Black-background skeleton video & Projection only & None \\
JointControlVideo & $C$ & $A_s$ & None & Camera-space 3D joints & Explicit intrinsics & None \\
RynnWorld-Teleop (SFT) & $C$ & $A_s$ & None & White-background skeleton video & Projection only & None \\
DWM & $C$ & $A_s$ & Repeated anchor & Surface-mesh video & Projection only & None \\
Hand2World (V0.9 AR) & $C$ & $A_s$ & None & X-ray mesh video & Per-frame poses and intrinsics & None \\
EgoHOI & $C$ & $A_s$ & Masked object RGB from $I_0$ & Surface-mesh images & Per-frame poses and intrinsics & Initial object mask \\
EgoSim & $C$ & $A_s$ & Scene rendering & Sphere/cylinder skeleton video & Encoded by scene rendering & Depth/point cloud and visibility; static robot background \\
\bottomrule\end{tabularx}\endgroup\par\vspace{12pt}
\centering
\caption{\textbf{Temporal context and reference access.} Window lengths are requested frames; the final sequence follows the task-specific generation schedule.}
\label{tab:conditioning_time}
\begingroup\fontsize{8}{9.5}\selectfont\setlength{\tabcolsep}{3pt}\renewcommand{\arraystretch}{1.12}
\begin{tabularx}{\textwidth}{@{}>{\raggedright\arraybackslash}p{2.6cm}>{\raggedright\arraybackslash}p{3.75cm}>{\raggedright\arraybackslash}p{1.8cm}c>{\raggedright\arraybackslash}X@{}}\toprule
Model / variant & Visual context & Control interval & Frames & Reference access \\\midrule
CogVideoX1.5 & $A_s$; generated history & None & 81 & Initial RGB and generated history \\
HunyuanVideo 1.5 & $A_s$; generated history & None & 121 & Initial RGB and generated history \\
Wan2.1 & $A_s$; generated history & None & 81 & Initial RGB and generated history \\
Wan2.2 & $A_s$; generated history & None & 121 & Initial RGB and generated history \\
PAM (hand-only) & $A_s$; generated history & $W_s$ & 49 & Target-window hand control \\
JointControlVideo & $A_s$; generated history & $W_s$ & 81 & Target-window 3D hand control and intrinsics \\
RynnWorld-Teleop (SFT) & $A_s$; generated history & $W_s$ & 81 & Target-window hand control \\
DWM & Repeated $A_s$; generated history & $W_s$ & 49 & Target-window surface motion; repeated anchor \\
Hand2World (V0.9 AR) & $A_s$; generated history & $W_s$ & 81 & Target-window hand and camera control \\
EgoHOI & $A_s$, object crop from $I_0$; generated history & $W_s$ & 81 & Target-window hand/camera control and original first-frame object crop \\
EgoSim & $A_s$; generated history & $W_s$ & 61 & Human scene preparation reads reference RGB over $[0,T]$; robot scenes use static priors \\
\bottomrule\end{tabularx}\endgroup\end{table*}
\subsection{Generation and Sequence Assembly}
\label{app:baselines:generation}
Sampling seeds are 42, except for Wan2.1/2.2 (20261015), JointControlVideo (0), EgoHOI (123), and Hand2World (43).
Long sequences use overlapping windows with model-compatible lengths.
Each continuation takes its initial image from the preceding generated output; temporal controls use the matching source interval, and assembly retains each output time step once.
The case-specific sampling plan sets the final clip length.
The batch protocols sample human references every second frame at 15\,fps, retain all simulation frames at 30\,fps, and sample real-robot references every third frame at 30\,fps.
The latter yields temporally compressed playback; source-frame mappings preserve correspondence for evaluation.
Run records retain window boundaries, seeds, sampling settings, and coordinate transforms.
Output checks verify decodability, frame counts, frame rates, and control-to-output correspondence.

\section{Complete Evaluation Results}
\label{app:results}

\subsection{Human Hands}
\label{app:results:human}
JointControlVideo ranks first on human-hand tasks with a \hwbscore{} of 56.34, followed by HunyuanVideo 1.5 at 54.42 and Wan2.2 at 53.85 (Table~\ref{tab:results_human}).
Their category profiles differ: JointControlVideo leads Physical Consistency and Temporal Coherence, whereas HunyuanVideo 1.5 leads Visual Quality and Appearance Consistency.
Hand2World leads Pose Control (0.535) and Keypoint Coverage (0.803), but its Physical Consistency is 0.102.
The strongest geometric alignment therefore selects a different model from the strongest aggregate interaction performance within this task set.
\begin{table*}[t]
\centering
\caption{\textbf{Human Hands results.} 322 reference tasks; all eleven models, seventeen metrics, and six category scores. Metrics and categories use a 0--1 scale; HWBScore uses 0--100. Rows are sorted by this subset\textquotesingle s HWBScore. Bold and underlining mark the best and second-best unrounded values per column. Cases reports reference-defined applicability (JEPA: collection size); \textemdash{} denotes structural inapplicability.}
\label{tab:results_human}
\begingroup
\linespread{1}
\fontsize{8}{9.2}\selectfont
\setlength{\tabcolsep}{1.5pt}
\renewcommand{\arraystretch}{1.12}
\begin{tabular*}{\textwidth}{@{\extracolsep{\fill}}l*{9}{c}@{}}
\toprule
\multirow[c]{3}{*}[-1.5pt]{Model} & \multicolumn{3}{c}{Visual Quality} & \multicolumn{3}{c}{Hand Fidelity} & \multicolumn{3}{c}{Appearance Consistency} \\
\cmidrule(lr){2-4}\cmidrule(lr){5-7}\cmidrule(lr){8-10}
 & \multirow[c]{2}{*}{MUSIQ} & Visual & JEPA & Pose & Keypoint & Hand & Entity & Background & Photometric \\
 &  & Plausibility & Similarity & Control & Coverage & Morphology & Consistency & Consistency & Consistency \\
\midrule
JointControlVideo~\cite{zhang2026jointcontrol} & 0.651 & 0.604 & \underline{0.789} & 0.485 & 0.625 & 0.844 & 0.857 & 0.908 & 0.594 \\
HunyuanVideo 1.5~\cite{wu2025hunyuanvideo15} & \underline{0.681} & \textbf{0.623} & 0.779 & 0.405 & 0.669 & \underline{0.878} & \underline{0.867} & \textbf{0.943} & \textbf{0.780} \\
Wan2.2~\cite{wan2025wan,wan2025wan22} & 0.631 & \underline{0.616} & 0.595 & 0.298 & 0.680 & 0.841 & \textbf{0.877} & 0.915 & 0.725 \\
Wan2.1~\cite{wan2025wan} & 0.676 & 0.601 & 0.711 & 0.298 & 0.602 & 0.829 & 0.846 & 0.902 & 0.636 \\
EgoHOI~\cite{li2026egohoi} & 0.660 & 0.601 & 0.620 & 0.440 & \underline{0.706} & 0.870 & 0.834 & 0.861 & 0.539 \\
DWM~\cite{kim2026dwm} & \textbf{0.699} & 0.600 & 0.570 & \underline{0.521} & 0.650 & \textbf{0.887} & 0.846 & \underline{0.926} & \underline{0.766} \\
EgoSim~\cite{hao2026egosim} & 0.650 & 0.585 & 0.686 & 0.517 & 0.676 & 0.851 & 0.851 & 0.873 & 0.672 \\
CogVideoX1.5~\cite{yang2025cogvideox} & 0.596 & 0.613 & 0.662 & 0.296 & 0.563 & 0.789 & 0.859 & 0.887 & 0.659 \\
RynnWorld-Teleop~\cite{zhao2026rynnworldteleop} & 0.533 & 0.572 & \textbf{0.810} & 0.329 & 0.422 & 0.789 & 0.833 & 0.918 & 0.727 \\
PAM~\cite{gao2026pam} & 0.570 & 0.598 & 0.545 & 0.346 & 0.476 & 0.842 & 0.828 & 0.887 & 0.589 \\
Hand2World~\cite{wang2026hand2world} & 0.643 & 0.578 & 0.308 & \textbf{0.535} & \textbf{0.803} & 0.771 & 0.755 & 0.643 & 0.452 \\
\midrule
Cases & 322 & 322 & 322 & 322 & 322 & 322 & 322 & 322 & 322 \\
\bottomrule
\end{tabular*}
\par\vspace{5pt}
\begin{tabular*}{\textwidth}{@{\extracolsep{\fill}}l*{9}{c}@{}}
\toprule
\multirow[c]{3}{*}[-1.5pt]{Model} & \multicolumn{2}{c}{Instruction Following} & \multicolumn{4}{c}{Physical Consistency} & \multicolumn{2}{c}{Temporal Coherence} & \multirow[c]{3}{*}[-1.5pt]{HWBScore} \\
\cmidrule(lr){2-3}\cmidrule(lr){4-7}\cmidrule(lr){8-9}
 & Action & Event & Trajectory & Causal & Evolution & State & State & Memory &  \\
 & Adherence & Adherence & Accuracy & Plausibility & Correctness & Regulation & Progression & Consistency &  \\
\midrule
JointControlVideo & 0.549 & 0.520 & \underline{0.225} & \textbf{0.693} & \textbf{0.327} & 0.250 & \underline{0.347} & \underline{0.360} & \textbf{56.34} \\
HunyuanVideo 1.5 & 0.525 & 0.440 & 0.184 & \underline{0.683} & 0.274 & 0.175 & 0.306 & 0.184 & \underline{54.42} \\
Wan2.2 & 0.549 & \underline{0.576} & 0.173 & 0.649 & \underline{0.327} & \underline{0.300} & \textbf{0.358} & 0.137 & 53.85 \\
Wan2.1 & \textbf{0.565} & \textbf{0.584} & 0.137 & 0.637 & 0.264 & 0.244 & 0.310 & 0.182 & 52.90 \\
EgoHOI & 0.543 & 0.496 & 0.200 & 0.546 & 0.170 & \textbf{0.350} & 0.198 & \textbf{0.361} & 52.66 \\
DWM & 0.505 & 0.464 & 0.211 & 0.560 & 0.214 & 0.200 & 0.257 & 0.141 & 52.26 \\
EgoSim & 0.473 & 0.472 & \textbf{0.254} & 0.550 & 0.110 & 0.200 & 0.138 & 0.211 & 50.77 \\
CogVideoX1.5 & 0.516 & 0.424 & 0.199 & 0.611 & 0.244 & 0.150 & 0.272 & 0.182 & 49.55 \\
RynnWorld-Teleop & 0.515 & 0.456 & 0.204 & 0.611 & 0.203 & 0.162 & 0.205 & 0.097 & 48.50 \\
PAM & \underline{0.552} & 0.544 & 0.192 & 0.625 & 0.215 & 0.100 & 0.231 & 0.106 & 48.21 \\
Hand2World & 0.379 & 0.256 & 0.095 & 0.288 & 0.025 & 0.000 & 0.052 & 0.044 & 38.27 \\
\midrule
Cases & 322 & 25 & 322 & 322 & 67 & 20 & 67 & 19 &  \\
\bottomrule
\end{tabular*}
\par\vspace{5pt}
\begin{tabular*}{\textwidth}{@{\extracolsep{\fill}}l*{6}{c}@{}}
\toprule
\multicolumn{7}{@{}l}{\textit{Category scores}} \\
\midrule
Model & \shortstack{Visual\\Quality} & \shortstack{Hand\\Fidelity} & \shortstack{Appearance\\Consistency} & \shortstack{Instruction\\Following} & \shortstack{Physical\\Consistency} & \shortstack{Temporal\\Coherence} \\
\midrule
JointControlVideo & \underline{0.681} & 0.651 & 0.786 & 0.535 & \textbf{0.374} & \textbf{0.353} \\
HunyuanVideo 1.5 & \textbf{0.694} & 0.651 & \textbf{0.863} & 0.483 & 0.329 & 0.245 \\
Wan2.2 & 0.614 & 0.606 & 0.839 & \underline{0.563} & \underline{0.362} & 0.248 \\
Wan2.1 & 0.663 & 0.576 & 0.795 & \textbf{0.574} & 0.321 & 0.246 \\
EgoHOI & 0.627 & 0.672 & 0.745 & 0.520 & 0.317 & \underline{0.280} \\
DWM & 0.623 & \underline{0.686} & \underline{0.846} & 0.484 & 0.296 & 0.199 \\
EgoSim & 0.641 & 0.681 & 0.799 & 0.473 & 0.278 & 0.174 \\
CogVideoX1.5 & 0.624 & 0.549 & 0.802 & 0.470 & 0.301 & 0.227 \\
RynnWorld-Teleop & 0.639 & 0.514 & 0.826 & 0.485 & 0.295 & 0.151 \\
PAM & 0.571 & 0.555 & 0.768 & 0.548 & 0.283 & 0.169 \\
Hand2World & 0.509 & \textbf{0.703} & 0.617 & 0.317 & 0.102 & 0.048 \\
\bottomrule
\end{tabular*}
\endgroup
\end{table*}

\subsection{Simulated Dexterous Hands}
\label{app:results:simulated}
Wan2.1 leads the simulated-hand set with a \hwbscore{} of 49.85, followed by JointControlVideo at 48.09 and Wan2.2 at 47.95 (Table~\ref{tab:results_simulated}).
Its Temporal Coherence of 0.660 combines State Progression of 0.321 with Memory Consistency of 1.000 on three applicable cases.
JointControlVideo leads Action Adherence and Event Adherence, whereas HunyuanVideo 1.5 leads Evolution Correctness (0.323) and State Progression (0.399).
Memory, event execution, and object-state outcomes consequently yield different leaders within the same simulated embodiment.
\begin{table*}[t]
\centering
\caption{\textbf{Simulated Dexterous Hands results.} 108 reference tasks; all eleven models, seventeen metrics, and six category scores. Metrics and categories use a 0--1 scale; HWBScore uses 0--100. Rows are sorted by this subset\textquotesingle s HWBScore. Bold and underlining mark the best and second-best unrounded values per column. Cases reports reference-defined applicability (JEPA: collection size); \textemdash{} denotes structural inapplicability.}
\label{tab:results_simulated}
\begingroup
\linespread{1}
\fontsize{8}{9.2}\selectfont
\setlength{\tabcolsep}{1.5pt}
\renewcommand{\arraystretch}{1.12}
\begin{tabular*}{\textwidth}{@{\extracolsep{\fill}}l*{9}{c}@{}}
\toprule
\multirow[c]{3}{*}[-1.5pt]{Model} & \multicolumn{3}{c}{Visual Quality} & \multicolumn{3}{c}{Hand Fidelity} & \multicolumn{3}{c}{Appearance Consistency} \\
\cmidrule(lr){2-4}\cmidrule(lr){5-7}\cmidrule(lr){8-10}
 & \multirow[c]{2}{*}{MUSIQ} & Visual & JEPA & Pose & Keypoint & Hand & Entity & Background & Photometric \\
 &  & Plausibility & Similarity & Control & Coverage & Morphology & Consistency & Consistency & Consistency \\
\midrule
Wan2.1~\cite{wan2025wan} & \textbf{0.714} & \underline{0.569} & 0.538 & 0.135 & 0.505 & \textemdash{} & 0.805 & 0.818 & \textbf{0.719} \\
JointControlVideo~\cite{zhang2026jointcontrol} & 0.654 & 0.553 & \textbf{0.612} & \textbf{0.335} & 0.530 & \textemdash{} & \textbf{0.826} & \underline{0.824} & 0.681 \\
Wan2.2~\cite{wan2025wan,wan2025wan22} & 0.660 & 0.564 & 0.512 & 0.126 & \textbf{0.569} & \textemdash{} & 0.789 & 0.794 & 0.648 \\
HunyuanVideo 1.5~\cite{wu2025hunyuanvideo15} & 0.693 & \textbf{0.584} & \underline{0.548} & 0.164 & \underline{0.555} & \textemdash{} & \underline{0.818} & \textbf{0.838} & \underline{0.714} \\
PAM~\cite{gao2026pam} & 0.596 & 0.527 & 0.472 & 0.251 & 0.508 & \textemdash{} & 0.770 & 0.767 & 0.602 \\
CogVideoX1.5~\cite{yang2025cogvideox} & 0.598 & 0.536 & 0.425 & 0.096 & 0.444 & \textemdash{} & 0.777 & 0.769 & 0.580 \\
DWM~\cite{kim2026dwm} & \underline{0.695} & 0.503 & 0.243 & \underline{0.275} & 0.480 & \textemdash{} & 0.780 & 0.597 & 0.492 \\
EgoHOI~\cite{li2026egohoi} & 0.664 & 0.541 & 0.372 & 0.190 & 0.522 & \textemdash{} & 0.738 & 0.778 & 0.591 \\
EgoSim~\cite{hao2026egosim} & 0.669 & 0.516 & 0.501 & 0.221 & 0.508 & \textemdash{} & 0.765 & 0.637 & 0.574 \\
RynnWorld-Teleop~\cite{zhao2026rynnworldteleop} & 0.490 & 0.508 & 0.382 & 0.222 & 0.463 & \textemdash{} & 0.744 & 0.608 & 0.515 \\
Hand2World~\cite{wang2026hand2world} & 0.657 & 0.530 & 0.131 & 0.275 & 0.468 & \textemdash{} & 0.619 & 0.133 & 0.078 \\
\midrule
Cases & 108 & 108 & 108 & 108 & 108 & \textemdash{} & 108 & 108 & 108 \\
\bottomrule
\end{tabular*}
\par\vspace{5pt}
\begin{tabular*}{\textwidth}{@{\extracolsep{\fill}}l*{9}{c}@{}}
\toprule
\multirow[c]{3}{*}[-1.5pt]{Model} & \multicolumn{2}{c}{Instruction Following} & \multicolumn{4}{c}{Physical Consistency} & \multicolumn{2}{c}{Temporal Coherence} & \multirow[c]{3}{*}[-1.5pt]{HWBScore} \\
\cmidrule(lr){2-3}\cmidrule(lr){4-7}\cmidrule(lr){8-9}
 & Action & Event & Trajectory & Causal & Evolution & State & State & Memory &  \\
 & Adherence & Adherence & Accuracy & Plausibility & Correctness & Regulation & Progression & Consistency &  \\
\midrule
Wan2.1 & 0.272 & 0.439 & 0.198 & 0.361 & 0.234 & 0.278 & 0.321 & \textbf{1.000} & \textbf{49.85} \\
JointControlVideo & \textbf{0.293} & \textbf{0.526} & 0.225 & \underline{0.367} & \underline{0.304} & \textbf{0.349} & 0.364 & 0.333 & \underline{48.09} \\
Wan2.2 & \underline{0.278} & 0.519 & 0.198 & 0.343 & 0.274 & 0.333 & \underline{0.378} & \underline{0.667} & 47.95 \\
HunyuanVideo 1.5 & 0.274 & \underline{0.525} & 0.230 & \textbf{0.383} & \textbf{0.323} & \underline{0.337} & \textbf{0.399} & 0.000 & 44.58 \\
PAM & 0.250 & 0.408 & 0.217 & 0.275 & 0.176 & 0.175 & 0.242 & \underline{0.667} & 43.63 \\
CogVideoX1.5 & 0.263 & 0.386 & 0.230 & 0.321 & 0.191 & 0.222 & 0.272 & \underline{0.667} & 42.22 \\
DWM & 0.241 & 0.395 & 0.236 & 0.216 & 0.154 & 0.151 & 0.245 & 0.333 & 37.96 \\
EgoHOI & 0.243 & 0.326 & 0.148 & 0.299 & 0.071 & 0.095 & 0.111 & 0.333 & 37.41 \\
EgoSim & 0.239 & 0.291 & \textbf{0.308} & 0.179 & 0.030 & 0.032 & 0.087 & 0.000 & 33.85 \\
RynnWorld-Teleop & 0.219 & 0.288 & \underline{0.248} & 0.253 & 0.033 & 0.032 & 0.041 & 0.000 & 30.66 \\
Hand2World & 0.235 & 0.182 & 0.049 & 0.148 & 0.000 & 0.000 & 0.000 & 0.000 & 22.42 \\
\midrule
Cases & 108 & 80 & 108 & 108 & 92 & 63 & 92 & 3 &  \\
\bottomrule
\end{tabular*}
\par\vspace{5pt}
\begin{tabular*}{\textwidth}{@{\extracolsep{\fill}}l*{6}{c}@{}}
\toprule
\multicolumn{7}{@{}l}{\textit{Category scores}} \\
\midrule
Model & \shortstack{Visual\\Quality} & \shortstack{Hand\\Fidelity} & \shortstack{Appearance\\Consistency} & \shortstack{Instruction\\Following} & \shortstack{Physical\\Consistency} & \shortstack{Temporal\\Coherence} \\
\midrule
Wan2.1 & \underline{0.607} & 0.320 & \underline{0.781} & 0.355 & 0.268 & \textbf{0.660} \\
JointControlVideo & 0.606 & \textbf{0.433} & 0.777 & \textbf{0.409} & \underline{0.311} & 0.349 \\
Wan2.2 & 0.579 & 0.347 & 0.743 & 0.398 & 0.287 & \underline{0.522} \\
HunyuanVideo 1.5 & \textbf{0.608} & 0.359 & \textbf{0.790} & \underline{0.400} & \textbf{0.318} & 0.200 \\
PAM & 0.532 & \underline{0.380} & 0.713 & 0.329 & 0.211 & 0.454 \\
CogVideoX1.5 & 0.520 & 0.270 & 0.708 & 0.325 & 0.241 & 0.469 \\
DWM & 0.480 & 0.378 & 0.623 & 0.318 & 0.189 & 0.289 \\
EgoHOI & 0.526 & 0.356 & 0.702 & 0.284 & 0.153 & 0.222 \\
EgoSim & 0.562 & 0.365 & 0.659 & 0.265 & 0.137 & 0.043 \\
RynnWorld-Teleop & 0.460 & 0.342 & 0.622 & 0.253 & 0.142 & 0.020 \\
Hand2World & 0.439 & 0.371 & 0.276 & 0.209 & 0.049 & 0.000 \\
\bottomrule
\end{tabular*}
\endgroup
\end{table*}

\subsection{Real Dexterous Hands}
\label{app:results:real}
JointControlVideo leads the real-hand set at 47.73, well ahead of Wan2.2 at 39.87, EgoHOI at 39.59, EgoSim at 39.21, and HunyuanVideo 1.5 at 38.55 (Table~\ref{tab:results_real}).
It leads Evolution Correctness, State Regulation, and State Progression, while HunyuanVideo 1.5 leads MUSIQ and Visual Plausibility.
Hand2World has the highest Pose Control (0.434) and Keypoint Coverage (0.716), yet scores 0.002 on Evolution Correctness and 0.000 on State Regulation.
The real-hand results distinguish geometric tracking from both visible image quality and the establishment and retention of manipulated states.
\begin{table*}[t]
\centering
\caption{\textbf{Real Dexterous Hands results.} 70 reference tasks; all eleven models, seventeen metrics, and six category scores. Metrics and categories use a 0--1 scale; HWBScore uses 0--100. Rows are sorted by this subset\textquotesingle s HWBScore. Bold and underlining mark the best and second-best unrounded values per column. Cases reports reference-defined applicability (JEPA: collection size); \textemdash{} denotes structural inapplicability.}
\label{tab:results_real}
\begingroup
\linespread{1}
\fontsize{8}{9.2}\selectfont
\setlength{\tabcolsep}{1.5pt}
\renewcommand{\arraystretch}{1.12}
\begin{tabular*}{\textwidth}{@{\extracolsep{\fill}}l*{9}{c}@{}}
\toprule
\multirow[c]{3}{*}[-1.5pt]{Model} & \multicolumn{3}{c}{Visual Quality} & \multicolumn{3}{c}{Hand Fidelity} & \multicolumn{3}{c}{Appearance Consistency} \\
\cmidrule(lr){2-4}\cmidrule(lr){5-7}\cmidrule(lr){8-10}
 & \multirow[c]{2}{*}{MUSIQ} & Visual & JEPA & Pose & Keypoint & Hand & Entity & Background & Photometric \\
 &  & Plausibility & Similarity & Control & Coverage & Morphology & Consistency & Consistency & Consistency \\
\midrule
JointControlVideo~\cite{zhang2026jointcontrol} & 0.571 & 0.547 & 0.238 & 0.219 & 0.409 & \textemdash{} & \textbf{0.832} & \textbf{0.757} & 0.586 \\
Wan2.2~\cite{wan2025wan,wan2025wan22} & 0.582 & \underline{0.588} & 0.217 & 0.060 & 0.447 & \textemdash{} & 0.764 & 0.684 & 0.527 \\
EgoHOI~\cite{li2026egohoi} & 0.617 & 0.539 & 0.188 & 0.183 & 0.495 & \textemdash{} & 0.763 & 0.732 & 0.551 \\
EgoSim~\cite{hao2026egosim} & 0.661 & 0.524 & \textbf{0.275} & \underline{0.340} & \underline{0.659} & \textemdash{} & \underline{0.808} & 0.667 & 0.398 \\
HunyuanVideo 1.5~\cite{wu2025hunyuanvideo15} & \textbf{0.694} & \textbf{0.608} & \underline{0.258} & 0.050 & 0.510 & \textemdash{} & 0.767 & \underline{0.744} & \underline{0.599} \\
Wan2.1~\cite{wan2025wan} & 0.635 & 0.555 & 0.159 & 0.051 & 0.412 & \textemdash{} & 0.735 & 0.633 & 0.364 \\
DWM~\cite{kim2026dwm} & \underline{0.676} & 0.488 & 0.125 & 0.201 & 0.395 & \textemdash{} & 0.806 & 0.713 & 0.541 \\
PAM~\cite{gao2026pam} & 0.537 & 0.515 & 0.151 & 0.156 & 0.411 & \textemdash{} & 0.790 & 0.701 & \textbf{0.600} \\
RynnWorld-Teleop~\cite{zhao2026rynnworldteleop} & 0.428 & 0.497 & 0.188 & 0.137 & 0.333 & \textemdash{} & 0.780 & 0.585 & 0.399 \\
Hand2World~\cite{wang2026hand2world} & 0.641 & 0.514 & 0.094 & \textbf{0.434} & \textbf{0.716} & \textemdash{} & 0.721 & 0.235 & 0.147 \\
CogVideoX1.5~\cite{yang2025cogvideox} & 0.463 & 0.528 & 0.082 & 0.051 & 0.369 & \textemdash{} & 0.753 & 0.587 & 0.436 \\
\midrule
Cases & 70 & 70 & 70 & 70 & 70 & \textemdash{} & 70 & 70 & 70 \\
\bottomrule
\end{tabular*}
\par\vspace{5pt}
\begin{tabular*}{\textwidth}{@{\extracolsep{\fill}}l*{9}{c}@{}}
\toprule
\multirow[c]{3}{*}[-1.5pt]{Model} & \multicolumn{2}{c}{Instruction Following} & \multicolumn{4}{c}{Physical Consistency} & \multicolumn{2}{c}{Temporal Coherence} & \multirow[c]{3}{*}[-1.5pt]{HWBScore} \\
\cmidrule(lr){2-3}\cmidrule(lr){4-7}\cmidrule(lr){8-9}
 & Action & Event & Trajectory & Causal & Evolution & State & State & Memory &  \\
 & Adherence & Adherence & Accuracy & Plausibility & Correctness & Regulation & Progression & Consistency &  \\
\midrule
JointControlVideo & \underline{0.449} & \textbf{0.513} & 0.149 & \underline{0.429} & \textbf{0.443} & \textbf{0.649} & \textbf{0.475} & \textemdash{} & \textbf{47.73} \\
Wan2.2 & \textbf{0.494} & 0.284 & 0.117 & 0.357 & \underline{0.271} & \underline{0.351} & 0.355 & \textemdash{} & \underline{39.87} \\
EgoHOI & 0.363 & \underline{0.380} & 0.092 & 0.348 & 0.238 & 0.324 & 0.285 & \textemdash{} & 39.59 \\
EgoSim & 0.351 & 0.290 & 0.309 & 0.262 & 0.093 & 0.162 & 0.215 & \textemdash{} & 39.21 \\
HunyuanVideo 1.5 & 0.337 & 0.241 & 0.185 & \textbf{0.481} & 0.228 & 0.108 & 0.270 & \textemdash{} & 38.55 \\
Wan2.1 & 0.394 & 0.273 & 0.080 & 0.381 & 0.261 & 0.135 & \underline{0.395} & \textemdash{} & 36.68 \\
DWM & 0.311 & 0.249 & \underline{0.321} & 0.257 & 0.102 & 0.000 & 0.185 & \textemdash{} & 34.15 \\
PAM & 0.323 & 0.344 & \textbf{0.332} & 0.295 & 0.108 & 0.027 & 0.140 & \textemdash{} & 34.10 \\
RynnWorld-Teleop & 0.343 & 0.330 & 0.216 & 0.324 & 0.134 & 0.135 & 0.160 & \textemdash{} & 31.55 \\
Hand2World & 0.280 & 0.189 & 0.056 & 0.148 & 0.002 & 0.000 & 0.010 & \textemdash{} & 27.57 \\
CogVideoX1.5 & 0.271 & 0.187 & 0.213 & 0.252 & 0.092 & 0.027 & 0.110 & \textemdash{} & 27.42 \\
\midrule
Cases & 70 & 70 & 70 & 70 & 50 & 37 & 50 & \textemdash{} &  \\
\bottomrule
\end{tabular*}
\par\vspace{5pt}
\begin{tabular*}{\textwidth}{@{\extracolsep{\fill}}l*{6}{c}@{}}
\toprule
\multicolumn{7}{@{}l}{\textit{Category scores}} \\
\midrule
Model & \shortstack{Visual\\Quality} & \shortstack{Hand\\Fidelity} & \shortstack{Appearance\\Consistency} & \shortstack{Instruction\\Following} & \shortstack{Physical\\Consistency} & \shortstack{Temporal\\Coherence} \\
\midrule
JointControlVideo & 0.452 & 0.314 & \textbf{0.725} & \textbf{0.481} & \textbf{0.417} & \textbf{0.475} \\
Wan2.2 & 0.462 & 0.253 & 0.658 & \underline{0.389} & \underline{0.274} & 0.355 \\
EgoHOI & 0.448 & 0.339 & 0.682 & 0.371 & 0.250 & 0.285 \\
EgoSim & \underline{0.487} & \underline{0.499} & 0.625 & 0.321 & 0.206 & 0.215 \\
HunyuanVideo 1.5 & \textbf{0.520} & 0.280 & \underline{0.703} & 0.289 & 0.251 & 0.270 \\
Wan2.1 & 0.449 & 0.232 & 0.577 & 0.334 & 0.214 & \underline{0.395} \\
DWM & 0.429 & 0.298 & 0.687 & 0.280 & 0.170 & 0.185 \\
PAM & 0.401 & 0.283 & 0.697 & 0.334 & 0.191 & 0.140 \\
RynnWorld-Teleop & 0.371 & 0.235 & 0.588 & 0.336 & 0.202 & 0.160 \\
Hand2World & 0.416 & \textbf{0.575} & 0.367 & 0.234 & 0.051 & 0.010 \\
CogVideoX1.5 & 0.358 & 0.210 & 0.592 & 0.229 & 0.146 & 0.110 \\
\bottomrule
\end{tabular*}
\endgroup
\end{table*}

\subsection{Task-Level Results and Coverage}
\label{app:results:tasks}
Table~\ref{tab:task_level_results_coverage} reports four metrics by operation family, subcategory, and embodiment.

\begin{table*}[!tp]
{\centering
\centering
\caption{\textbf{Task-level results.} Scores are case-weighted means over all applicable references of each group (higher is better); bold and underline mark the best and second-best distinct unrounded means. $A$ counts applicable references in the 500-case evaluation set; every model is scored on the same $A$ cases. Subcategories refine the family columns; the last three columns compare human, simulated, and real-robot hands within Move \& Arrange. Native missing-observation penalties are retained.}
\label{tab:task_level_results_coverage}
\begingroup\linespread{1}\fontsize{7.5}{8.0}\selectfont
\setlength{\tabcolsep}{0.6pt}\renewcommand{\arraystretch}{1.0}
\scalebox{0.89}{\begin{tabular*}{\textwidth}{@{\extracolsep{\fill}}l*{7}{c}@{\hspace{4pt}}*{6}{c}@{\hspace{4pt}}*{3}{c}@{}}
\toprule
 & \multicolumn{7}{c}{Primary operation families} & \multicolumn{6}{c}{Selected subcategories} & \multicolumn{3}{c}{Move $\times$ embodiment} \\
\cmidrule(lr){2-8}\cmidrule(lr){9-14}\cmidrule(lr){15-17}
Model & Move & Operate & Assemble & Shape & Transfer & Clean & Inspect & \shortstack{Turn /\\Press} & \shortstack{Wipe /\\Brush} & \shortstack{Fold /\\Bend} & \shortstack{Pick /\\Place} & \shortstack{Cut /\\Pinch} & \shortstack{Scoop /\\Pour} & Human & Sim. & Robot \\
\midrule
\multicolumn{17}{l}{\textbf{(a) Action Adherence $\uparrow$}} \\
Applicable references $A$ & 88 & 117 & 46 & 104 & 41 & 74 & 30 & 80 & 72 & 56 & 54 & 31 & 24 & 15 & 46 & 27 \\
\cmidrule(lr){1-17}
JointControlVideo & \textbf{0.411} & \underline{0.462} & 0.522 & \underline{0.542} & 0.390 & 0.516 & 0.500 & \underline{0.490} & 0.517 & \underline{0.546} & \textbf{0.467} & 0.490 & 0.400 & 0.560 & \textbf{0.274} & \textbf{0.563} \\
HunyuanVideo 1.5 & 0.336 & 0.421 & 0.509 & 0.498 & 0.341 & \underline{0.527} & 0.513 & 0.465 & \underline{0.525} & 0.482 & 0.352 & 0.458 & 0.358 & 0.507 & \textbf{0.274} & 0.348 \\
Wan2.2 & \underline{0.391} & 0.439 & 0.496 & 0.519 & \textbf{0.468} & \textbf{0.568} & \textbf{0.587} & 0.478 & \textbf{0.569} & 0.521 & \underline{0.426} & 0.490 & \textbf{0.525} & \textbf{0.613} & \underline{0.265} & \underline{0.481} \\
Wan2.1 & 0.330 & \textbf{0.468} & \underline{0.548} & \textbf{0.546} & \underline{0.424} & 0.516 & \underline{0.580} & \textbf{0.512} & 0.519 & \textbf{0.557} & 0.333 & \textbf{0.535} & \underline{0.475} & 0.533 & 0.248 & 0.356 \\
EgoHOI & 0.348 & 0.436 & 0.483 & 0.521 & 0.371 & 0.495 & 0.560 & 0.485 & 0.489 & 0.525 & 0.374 & \underline{0.523} & 0.392 & \underline{0.587} & 0.243 & 0.393 \\
DWM & 0.298 & 0.410 & 0.487 & 0.498 & 0.317 & 0.476 & 0.460 & 0.445 & 0.475 & 0.511 & 0.289 & 0.465 & 0.350 & 0.507 & 0.252 & 0.259 \\
PAM & 0.307 & 0.441 & \textbf{0.552} & 0.529 & 0.371 & 0.522 & 0.480 & 0.488 & \underline{0.525} & 0.536 & 0.311 & 0.490 & 0.383 & 0.547 & 0.239 & 0.289 \\
CogVideoX1.5 & 0.305 & 0.421 & 0.517 & 0.488 & 0.322 & 0.481 & 0.467 & 0.460 & 0.478 & 0.504 & 0.307 & 0.497 & 0.325 & 0.520 & \underline{0.265} & 0.252 \\
EgoSim & 0.284 & 0.374 & 0.452 & 0.469 & 0.390 & 0.468 & 0.460 & 0.400 & 0.467 & 0.446 & 0.281 & 0.458 & 0.433 & 0.480 & 0.239 & 0.252 \\
RynnWorld-Teleop & 0.311 & 0.409 & 0.465 & 0.515 & 0.371 & 0.465 & 0.453 & 0.462 & 0.461 & 0.511 & 0.333 & 0.490 & 0.375 & 0.533 & 0.213 & 0.356 \\
Hand2World & 0.255 & 0.354 & 0.313 & 0.375 & 0.307 & 0.338 & 0.407 & 0.398 & 0.336 & 0.407 & 0.252 & 0.316 & 0.342 & 0.320 & 0.243 & 0.237 \\
\midrule
\multicolumn{17}{l}{\textbf{(b) Evolution Correctness $\uparrow$}} \\
Applicable references $A$ & 72 & 40 & 14 & 52 & 21 & 8 & 2 & 12 & 8 & 30 & 45 & 22 & 12 & 1 & 44 & 27 \\
\cmidrule(lr){1-17}
JointControlVideo & \textbf{0.397} & \textbf{0.483} & 0.214 & \textbf{0.316} & \textbf{0.119} & \underline{0.281} & 0.000 & \textbf{0.403} & \underline{0.281} & \underline{0.175} & \textbf{0.526} & \textbf{0.508} & \textbf{0.167} & 0.000 & \underline{0.231} & \textbf{0.682} \\
HunyuanVideo 1.5 & 0.299 & \underline{0.418} & \underline{0.411} & \underline{0.272} & 0.016 & 0.125 & 0.000 & 0.306 & 0.125 & 0.128 & \underline{0.379} & \underline{0.470} & 0.028 & 0.000 & \textbf{0.259} & 0.373 \\
Wan2.2 & \underline{0.300} & 0.368 & \textbf{0.652} & 0.261 & 0.012 & 0.182 & 0.000 & 0.194 & 0.182 & \textbf{0.197} & 0.378 & 0.348 & 0.021 & \textbf{0.333} & 0.214 & \underline{0.439} \\
Wan2.1 & 0.249 & 0.353 & 0.381 & 0.223 & 0.040 & \textbf{0.307} & 0.000 & 0.264 & \textbf{0.307} & 0.103 & 0.313 & 0.386 & 0.010 & 0.000 & 0.187 & 0.361 \\
EgoHOI & 0.198 & 0.102 & 0.119 & 0.167 & 0.008 & 0.130 & 0.000 & 0.083 & 0.130 & 0.106 & 0.278 & 0.250 & 0.014 & 0.000 & 0.080 & 0.396 \\
DWM & 0.107 & 0.303 & 0.143 & 0.199 & 0.000 & 0.151 & \textbf{0.125} & 0.264 & 0.151 & 0.078 & 0.131 & 0.364 & 0.000 & 0.000 & 0.087 & 0.144 \\
PAM & 0.161 & 0.262 & 0.155 & 0.195 & 0.028 & 0.125 & 0.000 & \underline{0.375} & 0.125 & 0.093 & 0.182 & 0.333 & 0.021 & 0.000 & 0.177 & 0.142 \\
CogVideoX1.5 & 0.115 & 0.305 & 0.229 & 0.228 & \underline{0.048} & 0.247 & 0.000 & 0.222 & 0.247 & 0.164 & 0.125 & 0.314 & 0.000 & 0.000 & 0.152 & 0.060 \\
EgoSim & 0.075 & 0.054 & 0.000 & 0.132 & 0.008 & 0.023 & 0.000 & 0.042 & 0.023 & 0.050 & 0.112 & 0.244 & 0.014 & 0.000 & 0.025 & 0.159 \\
RynnWorld-Teleop & 0.117 & 0.075 & 0.107 & 0.188 & 0.032 & 0.000 & 0.000 & 0.083 & 0.000 & 0.058 & 0.154 & 0.364 & \underline{0.056} & 0.000 & 0.055 & 0.223 \\
Hand2World & 0.001 & 0.000 & 0.000 & 0.032 & 0.000 & 0.000 & 0.000 & 0.000 & 0.000 & 0.011 & 0.002 & 0.061 & 0.000 & 0.000 & 0.000 & 0.004 \\
\midrule
\multicolumn{17}{l}{\textbf{(c) Pose Control $\uparrow$}} \\
Applicable references $A$ & 88 & 117 & 46 & 104 & 41 & 74 & 30 & 80 & 72 & 56 & 54 & 31 & 24 & 15 & 46 & 27 \\
\cmidrule(lr){1-17}
JointControlVideo & \underline{0.367} & \underline{0.457} & 0.402 & 0.462 & 0.287 & 0.362 & \textbf{0.569} & 0.485 & 0.361 & 0.435 & \underline{0.346} & \textbf{0.536} & 0.253 & \underline{0.509} & \textbf{0.363} & 0.294 \\
HunyuanVideo 1.5 & 0.171 & 0.356 & 0.308 & 0.403 & 0.130 & 0.288 & 0.405 & 0.408 & 0.285 & 0.436 & 0.145 & 0.367 & 0.074 & 0.342 & 0.171 & 0.077 \\
Wan2.2 & 0.135 & 0.266 & 0.219 & 0.314 & 0.144 & 0.199 & 0.245 & 0.320 & 0.197 & 0.309 & 0.120 & 0.319 & 0.114 & 0.257 & 0.139 & 0.059 \\
Wan2.1 & 0.154 & 0.257 & 0.233 & 0.312 & 0.113 & 0.198 & 0.260 & 0.306 & 0.197 & 0.325 & 0.131 & 0.294 & 0.082 & 0.323 & 0.152 & 0.065 \\
EgoHOI & 0.256 & 0.379 & 0.363 & 0.425 & 0.210 & 0.338 & 0.455 & 0.421 & 0.335 & 0.433 & 0.265 & 0.407 & 0.191 & 0.382 & 0.209 & 0.264 \\
DWM & 0.298 & \textbf{0.468} & \underline{0.422} & 0.506 & 0.250 & 0.433 & 0.545 & \textbf{0.501} & 0.435 & \underline{0.533} & 0.286 & 0.465 & 0.237 & 0.481 & 0.286 & 0.217 \\
PAM & 0.252 & 0.335 & 0.303 & 0.332 & 0.176 & 0.281 & 0.386 & 0.375 & 0.283 & 0.372 & 0.238 & 0.272 & 0.125 & 0.312 & 0.258 & 0.210 \\
CogVideoX1.5 & 0.114 & 0.248 & 0.203 & 0.324 & 0.109 & 0.186 & 0.295 & 0.287 & 0.188 & 0.348 & 0.097 & 0.272 & 0.061 & 0.241 & 0.101 & 0.066 \\
EgoSim & 0.325 & 0.415 & 0.413 & \underline{0.519} & \underline{0.361} & \underline{0.440} & \underline{0.557} & 0.455 & \underline{0.440} & 0.521 & 0.344 & \underline{0.513} & \underline{0.367} & 0.478 & 0.237 & \underline{0.391} \\
RynnWorld-Teleop & 0.241 & 0.322 & 0.250 & 0.321 & 0.182 & 0.243 & 0.349 & 0.351 & 0.245 & 0.348 & 0.221 & 0.302 & 0.148 & 0.276 & 0.257 & 0.193 \\
Hand2World & \textbf{0.396} & 0.455 & \textbf{0.450} & \textbf{0.528} & \textbf{0.433} & \textbf{0.477} & 0.521 & \underline{0.500} & \textbf{0.481} & \textbf{0.537} & \textbf{0.424} & 0.505 & \textbf{0.503} & \textbf{0.540} & \underline{0.308} & \textbf{0.467} \\
\midrule
\multicolumn{17}{l}{\textbf{(d) Trajectory Accuracy $\uparrow$}} \\
Applicable references $A$ & 88 & 117 & 46 & 104 & 41 & 74 & 30 & 80 & 72 & 56 & 54 & 31 & 24 & 15 & 46 & 27 \\
\cmidrule(lr){1-17}
JointControlVideo & 0.249 & \underline{0.248} & 0.142 & 0.163 & 0.160 & 0.191 & \textbf{0.399} & 0.248 & 0.186 & 0.105 & 0.272 & 0.259 & 0.130 & \textbf{0.292} & 0.264 & 0.199 \\
HunyuanVideo 1.5 & 0.288 & 0.239 & 0.129 & 0.156 & 0.143 & 0.137 & 0.191 & 0.281 & 0.134 & 0.119 & 0.359 & 0.200 & 0.131 & 0.247 & 0.260 & 0.359 \\
Wan2.2 & 0.191 & 0.232 & 0.145 & 0.147 & 0.160 & 0.128 & 0.105 & \underline{0.289} & 0.124 & 0.088 & 0.230 & 0.212 & 0.164 & 0.177 & 0.216 & 0.156 \\
Wan2.1 & 0.193 & 0.192 & 0.144 & 0.121 & 0.077 & 0.088 & 0.103 & 0.204 & 0.086 & 0.065 & 0.224 & 0.166 & 0.074 & 0.194 & 0.234 & 0.123 \\
EgoHOI & 0.156 & 0.217 & 0.080 & 0.149 & 0.099 & 0.189 & 0.354 & 0.221 & 0.187 & 0.119 & 0.188 & 0.213 & 0.075 & 0.222 & 0.147 & 0.135 \\
DWM & \underline{0.331} & 0.238 & \underline{0.165} & 0.152 & 0.247 & 0.190 & 0.380 & 0.247 & 0.189 & 0.107 & \textbf{0.434} & 0.251 & \underline{0.283} & 0.243 & 0.274 & \textbf{0.477} \\
PAM & 0.293 & 0.211 & 0.143 & 0.155 & \textbf{0.264} & \underline{0.219} & 0.271 & 0.241 & \underline{0.218} & 0.109 & 0.384 & 0.225 & \textbf{0.320} & 0.221 & 0.217 & \underline{0.464} \\
CogVideoX1.5 & 0.301 & 0.225 & \textbf{0.168} & \underline{0.198} & 0.146 & 0.156 & 0.174 & 0.253 & 0.153 & \underline{0.146} & 0.356 & \underline{0.274} & 0.150 & 0.256 & \underline{0.277} & 0.366 \\
EgoSim & \textbf{0.344} & \textbf{0.315} & 0.104 & \textbf{0.236} & \underline{0.263} & \textbf{0.238} & \underline{0.393} & \textbf{0.330} & \textbf{0.237} & \textbf{0.204} & \underline{0.434} & \textbf{0.306} & 0.266 & 0.222 & \textbf{0.319} & 0.455 \\
RynnWorld-Teleop & 0.287 & 0.234 & 0.154 & 0.179 & 0.148 & 0.193 & 0.300 & 0.245 & 0.195 & 0.125 & 0.340 & 0.271 & 0.112 & \underline{0.262} & 0.255 & 0.354 \\
Hand2World & 0.074 & 0.106 & 0.046 & 0.051 & 0.059 & 0.062 & 0.218 & 0.110 & 0.063 & 0.063 & 0.093 & 0.032 & 0.068 & 0.123 & 0.058 & 0.074 \\
\bottomrule
\end{tabular*}}
\endgroup
\par}

\end{table*}

\section{Additional Diagnostics and Robustness}
\label{app:diagnostics}

\subsection{Relationships Between Interaction Capabilities}
\label{app:diagnostics:relationships}
Figure~\ref{fig:capability_diagnostics_scatter} compares models on four pair-specific task sets.
Both metrics in a pair use the same tasks with equal weights, producing one coordinate per model.
Table~\ref{tab:diagnostic_relationships} gives the resulting task counts and correlations; Table~\ref{tab:diagnostic_model_means} reports all coordinates.
Spearman correlation is computed across eleven model means.

The motion--outcome comparisons locate where model orderings change.
On the 209 pose--progression tasks, Hand2World has the highest Pose Control at 0.402, but State Progression is 0.019; JointControlVideo scores 0.358 and 0.385, respectively.
The object-motion comparison shows the same reversal: EgoSim reaches Trajectory Accuracy of 0.292, compared with JointControlVideo's 0.194, while their Evolution Correctness scores are 0.063 and 0.334.
These matched-task results place the discrepancy within shared interactions, where spatial agreement and the required transformation remain different prediction targets.

State Progression and State Regulation share an establishment requirement, which contributes to their closer ordering.
Their distinction appears in EgoSim and CogVideoX1.5: progression scores are 0.171 and 0.190, whereas regulation scores are 0.085 and 0.154.
The memory comparison measures how the output preserves the object's visible history.
JointControlVideo and HunyuanVideo 1.5 have similar Entity Consistency of 0.831 and 0.857, yet Memory Consistency is 0.356 and 0.159.
The local historical requirements discriminate outputs whose overall entity features remain close.

\begin{table*}[!tp]\centering
\caption{\textbf{Model-level relationships on matched tasks.} Spearman $\rho$ uses eleven unrounded model means. Intervals are percentile bootstrap intervals over reference cases, stratified by embodiment. The final column gives the range after dropping one model at a time.}
\label{tab:diagnostic_relationships}
\small\setlength{\tabcolsep}{4pt}
\begin{tabularx}{\textwidth}{@{}c>{\raggedright\arraybackslash}X>{\raggedright\arraybackslash}Xrrcc@{}}\toprule
Panel & First metric & Second metric & Tasks & $\rho$ & 95\% CI & Leave-one-model-out \\
\midrule
(a) & Pose Control & State Progression & 209 & $-0.345$ & $[-0.564, -0.291]$ & $[-0.673, -0.127]$ \\
(b) & Trajectory Accuracy & Evolution Correctness & 200 & $-0.173$ & $[-0.300, 0.000]$ & $[-0.564, -0.018]$ \\
(c) & State Progression & State Regulation & 117 & $0.936$ & $[0.782, 0.980]$ & $[0.915, 0.988]$ \\
(d) & Entity Consistency & Memory Consistency & 22 & $0.009$ & $[-0.218, 0.545]$ & $[-0.321, 0.224]$ \\
\bottomrule\end{tabularx}
\end{table*}

\begin{table*}[!t]\centering
\caption{\textbf{Matched-task means underlying Figure~\ref{fig:capability_diagnostics_scatter}.} The two columns in each group use the same reference tasks; task sets differ between groups. Scores are displayed to three decimal places.}
\label{tab:diagnostic_model_means}
\begingroup\fontsize{8}{9.5}\selectfont\setlength{\tabcolsep}{2.5pt}\renewcommand{\arraystretch}{1.14}
\begin{tabular*}{\textwidth}{@{\extracolsep{\fill}}lrrrrrrrr@{}}\toprule
& \multicolumn{2}{c}{(a) Motion / progress} & \multicolumn{2}{c}{(b) Path / outcome} & \multicolumn{2}{c}{(c) Progress / retention} & \multicolumn{2}{c}{(d) Entity / memory} \\
\cmidrule(lr){2-3}\cmidrule(lr){4-5}\cmidrule(lr){6-7}\cmidrule(l){8-9}
Model & \shortstack{Pose\\Control} & \shortstack{State\\Progression} & \shortstack{Trajectory\\Accuracy} & \shortstack{Evolution\\Correctness} & \shortstack{State\\Progression} & \shortstack{State\\Regulation} & \shortstack{Entity\\Consistency} & \shortstack{Memory\\Consistency} \\
\midrule
JointControlVideo~\cite{zhang2026jointcontrol} & 0.358 & 0.385 & 0.194 & 0.334 & 0.440 & 0.427 & 0.831 & 0.356 \\
HunyuanVideo 1.5~\cite{wu2025hunyuanvideo15} & 0.210 & 0.339 & 0.204 & 0.268 & 0.368 & 0.246 & 0.857 & 0.159 \\
Wan2.2~\cite{wan2025wan,wan2025wan22} & 0.152 & 0.366 & 0.150 & 0.277 & 0.389 & 0.342 & 0.846 & 0.209 \\
Wan2.1~\cite{wan2025wan} & 0.156 & 0.335 & 0.134 & 0.237 & 0.365 & 0.234 & 0.803 & 0.293 \\
EgoHOI~\cite{li2026egohoi} & 0.259 & 0.181 & 0.147 & 0.139 & 0.207 & 0.205 & 0.784 & 0.357 \\
EgoSim~\cite{hao2026egosim} & 0.343 & 0.134 & 0.292 & 0.063 & 0.171 & 0.085 & 0.784 & 0.182 \\
DWM~\cite{kim2026dwm} & 0.327 & 0.234 & 0.236 & 0.152 & 0.216 & 0.107 & 0.834 & 0.168 \\
PAM~\cite{gao2026pam} & 0.247 & 0.214 & 0.220 & 0.166 & 0.192 & 0.120 & 0.802 & 0.183 \\
CogVideoX1.5~\cite{yang2025cogvideox} & 0.150 & 0.233 & 0.218 & 0.172 & 0.190 & 0.154 & 0.849 & 0.248 \\
RynnWorld-Teleop~\cite{zhao2026rynnworldteleop} & 0.235 & 0.122 & 0.218 & 0.102 & 0.118 & 0.079 & 0.803 & 0.084 \\
Hand2World~\cite{wang2026hand2world} & 0.402 & 0.019 & 0.058 & 0.006 & 0.013 & 0.000 & 0.698 & 0.038 \\
\bottomrule\end{tabular*}
\endgroup
\par\smallskip{\raggedright\noindent \textbf{Correlation robustness.} We jointly resample reference cases within each embodiment for each of Figure~\ref{fig:capability_diagnostics_scatter}'s four matched task sets (5,000 draws, seed 20261118), then recompute each metric mean and Spearman $\rho$ across the eleven models. Table~\ref{tab:diagnostic_relationships} reports the percentile intervals and the range obtained by dropping each model in turn. The pose--progression interval is below zero. The trajectory--evolution interval reaches zero and the entity--memory interval includes zero. Removing Hand2World changes the four correlations to $-0.127$, $-0.564$, $0.915$, and $-0.321$, respectively. The state progression--regulation relationship remains positive in every leave-one-model-out comparison.\par}
\end{table*}

\paragraph{Conditioning in the reruns.}
The stability study in Appendix~\ref{app:diagnostics:stability} fixes 24 reference cases (eight per embodiment), their first RGB images, task text, generation windows, and source-frame evaluation horizons. JointControlVideo receives camera-space hand control and intrinsics, and HunyuanVideo 1.5 receives image and text. The real-robot JointControlVideo cases use the refined geometry examined in Appendix~\ref{app:unified_control:adapters}. Separately, an audit of four EgoSim human-hand runs confirmed that scene preprocessing read all 81 RGB frames of each case to estimate camera pose and depth. We report this access as part of the evaluated configuration.

\subsection{Generation and Scoring Stability}
\label{app:diagnostics:stability}

\paragraph{Generation seeds.}
We regenerate JointControlVideo and HunyuanVideo 1.5 with seeds 0, 42, and 20261118 on the 24 cases above while keeping all inputs and evaluation windows fixed, which yields 144 generated videos. Table~\ref{tab:seed_stability} reports the paired JointControlVideo minus HunyuanVideo 1.5 difference for every metric. The 95\% intervals jointly resample source cases within embodiment and each model's seed (5,000 draws, seed 20261118), so they include both case and generation variability. Pose Control favors JointControlVideo in 210 of the 216 case--seed combinations, with a difference of $+0.160$ $[0.127, 0.194]$; a simultaneous 95\% band over the eight scalar metrics with full coverage gives $[0.113, 0.207]$. Collection-level JEPA, recomputed on every draw, favors JointControlVideo by $+0.143$ $[0.085, 0.228]$, and the advantage holds for each of the nine seed pairings. State Regulation ($+0.311$) and Event Adherence ($+0.110$) also favor JointControlVideo, whereas HunyuanVideo 1.5 retains higher Keypoint Coverage, MUSIQ, Visual Plausibility, and Photometric Consistency; the latter four intervals stay below zero under the simultaneous band. Memory Consistency has only three applicable cases in this design and is not summarized. These directions agree with the category profiles in Table~\ref{tab:main_results}.

\begin{table}[!t]
\centering
\caption{\textbf{Three-seed paired differences.} JointControlVideo minus HunyuanVideo 1.5 on 24 cases with three seeds per model. $n$ counts cases with valid scores for all six videos. Wins count the 9 seed pairings per case (J/H/tie).}
\label{tab:seed_stability}
\footnotesize\setlength{\tabcolsep}{2pt}
\begin{tabular*}{\columnwidth}{@{\extracolsep{\fill}}lrrcr@{}}
\toprule
Metric & $n$ & Diff. & 95\% CI & J/H/tie \\
\midrule
MUSIQ & 24 & $-0.028$ & $[-0.046, -0.010]$ & 74/142/0 \\
Visual Plausibility & 24 & $-0.028$ & $[-0.038, -0.018]$ & 39/176/1 \\
JEPA Similarity & 24 & $+0.143$ & $[0.085, 0.228]$ & \textemdash{} \\
Pose Control & 24 & $+0.160$ & $[0.127, 0.194]$ & 210/6/0 \\
Keypoint Coverage & 24 & $-0.093$ & $[-0.146, -0.046]$ & 67/147/2 \\
Hand Morphology & 8 & $+0.018$ & $[-0.018, 0.056]$ & 43/29/0 \\
Entity Consistency & 13 & $+0.014$ & $[-0.006, 0.028]$ & 79/38/0 \\
Background Consistency & 24 & $-0.015$ & $[-0.032, 0.004]$ & 81/135/0 \\
Photometric Consistency & 24 & $-0.102$ & $[-0.141, -0.058]$ & 35/181/0 \\
Action Adherence & 24 & $+0.014$ & $[-0.056, 0.083]$ & 61/56/99 \\
Event Adherence & 14 & $+0.110$ & $[0.014, 0.195]$ & 58/33/35 \\
Trajectory Accuracy & 17 & $+0.008$ & $[-0.060, 0.082]$ & 75/55/23 \\
Causal Plausibility & 24 & $+0.022$ & $[-0.037, 0.079]$ & 67/48/101 \\
Evolution Correctness & 18 & $+0.059$ & $[-0.057, 0.162]$ & 79/42/41 \\
State Regulation & 15 & $+0.311$ & $[0.133, 0.467]$ & 63/21/51 \\
State Progression & 18 & $+0.083$ & $[-0.028, 0.185]$ & 66/21/75 \\
\bottomrule
\end{tabular*}
\end{table}

\paragraph{Semantic scoring repeatability.}
For the 12 cases of the first generation round, we score each model's videos three times, regenerating every semantic response from the same prompts and evidence with greedy decoding. Across 130 case--metric comparisons covering Action Adherence, Event Adherence, Causal Plausibility, Evolution Correctness, State Progression, State Regulation, and Memory Consistency, the three scorings agree exactly in every comparison. Changing the generation seed changes the score in 79 of the same 130 comparisons; for Action Adherence, the median across-seed range is 0.20 for both models. Score variation in the semantic metrics therefore arises from the generated videos rather than from the judge.

\paragraph{Case-sampling uncertainty.}
The HWBScore intervals in Table~\ref{tab:case_sampling_uncertainty} and the bootstrap analyses in Appendix~\ref{app:diagnostics:sensitivity} use a stratified subset of 200 cases (129 human, 43 simulated, and 28 real-robot cases from 121, 43, and 28 source groups), drawn in proportion to the evaluation set. Point estimates on this subset therefore differ slightly from the full-set scores in Table~\ref{tab:main_results} (55.32 and 51.51 versus 55.51 and 52.22). We draw 5,000 paired source-cluster bootstrap samples, stratified by embodiment (seed 20261118), and recompute metric and category scores on every draw, including collection-level JEPA (Table~\ref{tab:case_sampling_uncertainty}). Across draws, JointControlVideo ranks first or second overall and first on real hands. Simulation retains 4,792 valid composite draws because some resamples contain no applicable memory case. On the three simulated memory cases, JointControlVideo averages 0.333; deleting one case yields means from 0.000 to 0.500. The point estimates and intervals describe sensitivity to case composition.

\begin{table*}[!tp]
\centering
\caption{\textbf{Case-sampling uncertainty.} HWBScore of JointControlVideo and HunyuanVideo 1.5 and their paired difference on the 200-case analysis subset, in points, with 95\% percentile intervals.}
\label{tab:case_sampling_uncertainty}
\small
\begin{tabular*}{\textwidth}{@{\extracolsep{\fill}}lrrrrrr@{}}
\toprule
& \multicolumn{2}{c}{JointControlVideo} & \multicolumn{2}{c}{HunyuanVideo 1.5} & \multicolumn{2}{c}{Difference} \\
\cmidrule(lr){2-3}\cmidrule(lr){4-5}\cmidrule(l){6-7}
Subset & Score & 95\% CI & Score & 95\% CI & Points & 95\% CI \\
\midrule
Overall & $55.32$ & $[53.07, 57.47]$ & $51.51$ & $[49.54, 53.44]$ & $3.81$ & $[1.75, 6.01]$ \\
Human & $57.84$ & $[54.65, 60.79]$ & $54.19$ & $[50.83, 57.52]$ & $3.65$ & $[0.54, 6.60]$ \\
Simulated & $46.50$ & $[41.21, 52.95]$ & $44.01$ & $[40.70, 46.36]$ & $2.49$ & $[-2.42, 9.28]$ \\
Real robot & $48.16$ & $[42.54, 53.30]$ & $36.47$ & $[33.00, 39.36]$ & $11.69$ & $[7.19, 15.95]$ \\
\bottomrule
\end{tabular*}
\end{table*}

\begin{table*}[!tp]
\centering
\caption{\textbf{Aggregation alternatives.} Scores (ranks) under the published HWBScore, equal weighting of individual metrics, and removal of each category. All scores use the same frozen metric inputs.}
\label{tab:aggregation_alternatives}
\begingroup\fontsize{8}{9.5}\selectfont\setlength{\tabcolsep}{2pt}
\begin{tabular*}{\textwidth}{@{\extracolsep{\fill}}l*{8}{c}@{}}
\toprule
Model & HWBScore & Metric-equal & $-$VQ & $-$HF & $-$AC & $-$IF & $-$PC & $-$TC \\
\midrule
JointControlVideo & 55.51 (1) & 55.96 (1) & 53.00 (1) & 54.39 (1) & 51.10 (1) & 56.62 (1) & 58.76 (1) & 59.20 (1) \\
HunyuanVideo 1.5 & 52.22 (2) & 53.27 (2) & 48.48 (3) & 50.63 (3) & 46.16 (2) & 54.22 (2) & 56.12 (2) & 57.68 (2) \\
Wan2.2 & 51.65 (3) & 52.27 (3) & 48.72 (2) & 50.70 (2) & 46.12 (3) & 52.82 (3) & 55.30 (3) & 56.23 (3) \\
Wan2.1 & 50.34 (4) & 50.60 (4) & 46.76 (4) & 49.66 (4) & 45.18 (4) & 51.70 (4) & 54.59 (4) & 54.12 (4) \\
EgoHOI & 48.74 (5) & 49.06 (5) & 45.50 (5) & 46.11 (5) & 43.95 (5) & 50.24 (5) & 53.54 (5) & 53.11 (5) \\
DWM & 47.31 (6) & 48.05 (6) & 44.53 (6) & 44.20 (8) & 41.27 (6) & 49.11 (6) & 52.03 (6) & 52.76 (6) \\
EgoSim & 46.33 (7) & 47.28 (7) & 42.41 (9) & 42.81 (9) & 40.71 (7) & 48.37 (7) & 51.23 (7) & 52.43 (7) \\
PAM & 46.06 (8) & 46.57 (8) & 43.26 (7) & 44.51 (7) & 40.35 (8) & 46.72 (9) & 50.23 (8) & 51.31 (8) \\
CogVideoX1.5 & 45.88 (9) & 46.52 (9) & 42.56 (8) & 44.94 (6) & 40.01 (9) & 47.67 (8) & 49.86 (9) & 50.25 (9) \\
RynnWorld-Teleop & 42.60 (10) & 43.61 (10) & 39.03 (10) & 41.21 (10) & 36.15 (10) & 43.57 (10) & 46.59 (10) & 49.06 (10) \\
Hand2World & 34.42 (11) & 35.20 (11) & 30.68 (11) & 28.28 (11) & 31.15 (11) & 36.01 (11) & 39.68 (11) & 40.74 (11) \\
\bottomrule
\end{tabular*}
\endgroup
\par\vspace{13pt}
\centering
\caption{\textbf{Sensitivity to Memory Consistency.} HWBScore points and rank in parentheses. Ranks are computed before rounding. Equal uses the original weighting within Temporal Coherence, 3:1 weights State Progression three times as much as Memory Consistency, and Remove drops Memory Consistency. Outer weights are unchanged.}
\label{tab:memory_sensitivity}
\small\setlength{\tabcolsep}{2.6pt}
\begin{tabular*}{\textwidth}{@{\extracolsep{\fill}}l*{9}{r}@{}}
\toprule
& \multicolumn{3}{c}{Overall} & \multicolumn{3}{c}{Human hands} & \multicolumn{3}{c}{Simulated hands} \\
\cmidrule(lr){2-4}\cmidrule(lr){5-7}\cmidrule(l){8-10}
Model & Equal & 3:1 & Remove & Equal & 3:1 & Remove & Equal & 3:1 & Remove \\
\midrule
JointControlVideo & 55.51 (1) & 55.63 (1) & 55.75 (1) & 56.34 (1) & 56.29 (1) & 56.24 (1) & 48.09 (2) & 48.22 (1) & 48.35 (1) \\
HunyuanVideo 1.5 & 52.22 (2) & 52.96 (2) & 53.71 (2) & 54.42 (2) & 54.92 (2) & 55.43 (3) & 44.58 (4) & 46.25 (4) & 47.91 (2) \\
Wan2.2 & 51.65 (3) & 52.30 (3) & 52.96 (3) & 53.85 (3) & 54.77 (3) & 55.70 (2) & 47.95 (3) & 46.74 (3) & 45.54 (3) \\
Wan2.1 & 50.34 (4) & 50.51 (4) & 50.68 (4) & 52.90 (4) & 53.43 (4) & 53.96 (4) & 49.85 (1) & 47.02 (2) & 44.18 (4) \\
EgoHOI & 48.74 (5) & 48.01 (5) & 47.27 (6) & 52.66 (5) & 51.98 (6) & 51.30 (6) & 37.41 (8) & 36.48 (8) & 35.56 (8) \\
DWM & 47.31 (6) & 47.59 (6) & 47.87 (5) & 52.26 (6) & 52.74 (5) & 53.22 (5) & 37.96 (7) & 37.59 (7) & 37.22 (7) \\
EgoSim & 46.33 (7) & 46.13 (8) & 45.93 (8) & 50.77 (7) & 50.47 (7) & 50.17 (8) & 33.85 (9) & 34.21 (9) & 34.58 (9) \\
PAM & 46.06 (8) & 46.20 (7) & 46.33 (7) & 48.21 (10) & 48.74 (10) & 49.26 (10) & 43.63 (5) & 41.86 (5) & 40.09 (5) \\
CogVideoX1.5 & 45.88 (9) & 45.82 (9) & 45.76 (9) & 49.55 (8) & 49.92 (8) & 50.30 (7) & 42.22 (6) & 40.57 (6) & 38.93 (6) \\
RynnWorld-Teleop & 42.60 (10) & 42.76 (10) & 42.92 (10) & 48.50 (9) & 48.95 (9) & 49.40 (9) & 30.66 (10) & 30.83 (10) & 31.00 (10) \\
Hand2World & 34.42 (11) & 34.35 (11) & 34.27 (11) & 38.27 (11) & 38.30 (11) & 38.34 (11) & 22.42 (11) & 22.42 (11) & 22.42 (11) \\
\bottomrule
\end{tabular*}
\end{table*}
\begin{table*}[!tp]
\centering
\caption{\textbf{Sensitivity across evaluation sets.} Spearman correlations compare alternative rankings with published HWBScore rankings. Grid ranges cover 88,913 bounded category-weight settings.}
\label{tab:aggregation_subsets}
\begingroup\small\setlength{\tabcolsep}{4pt}
\begin{tabular*}{\textwidth}{@{\extracolsep{\fill}}lcccc>{\raggedright\arraybackslash}p{3.65cm}@{}}
\toprule
Set & Metric-equal $\rho$ & Removal $\rho$ & Grid $\rho$ & Max. rank shift & Grid winners \\
\midrule
Overall & 1.000 & 0.909--1.000 & 0.909--1.000 & 3 & JointControlVideo \\
Human hands & 0.991 & 0.936--1.000 & 0.882--1.000 & 3 & JointControlVideo, HunyuanVideo 1.5 \\
Simulated hands & 1.000 & 0.927--0.982 & 0.955--1.000 & 2 & Wan2.1, JointControlVideo \\
Real hands & 0.973 & 0.900--0.982 & 0.864--1.000 & 3 & JointControlVideo \\
\bottomrule
\end{tabular*}
\endgroup
\par\vspace{13pt}
{\captionsetup{type=figure}
\centering
\includegraphics[width=0.81\textwidth]{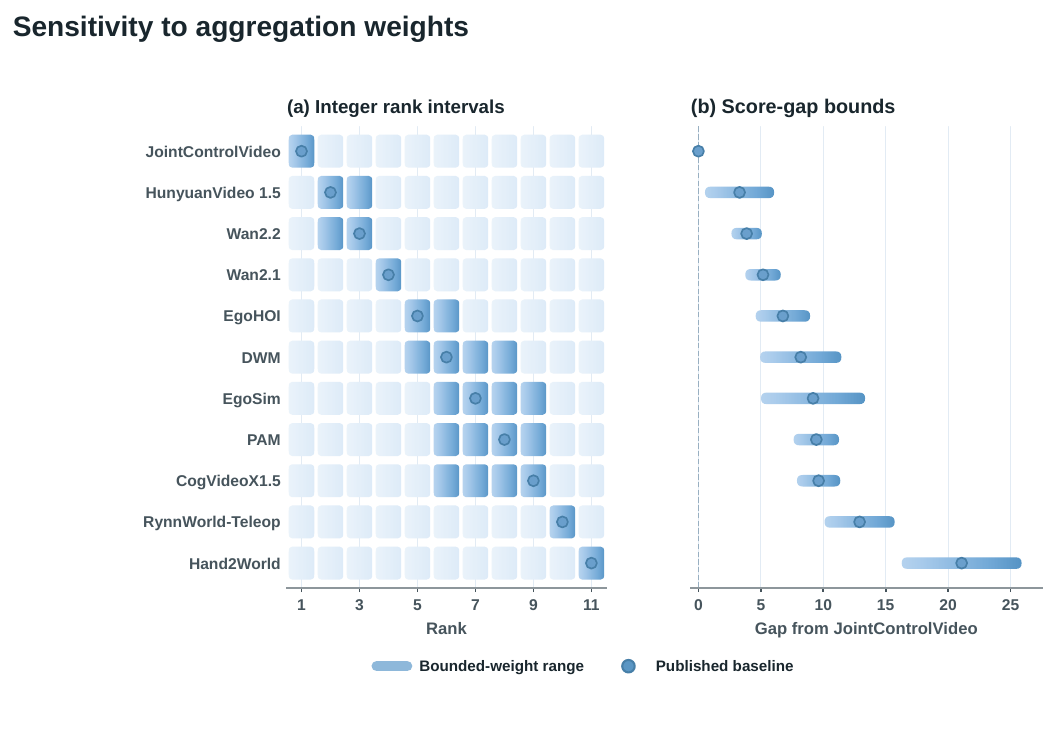}
\caption{\textbf{Sensitivity to aggregation weights.} (a) Integer rank bounds over 88,913 weight vectors with each category weight in $[1/12,1/4]$. Tiles enumerate integer positions within each range. (b) Continuous score-gap bounds relative to JointControlVideo on the HWBScore scale. Dots show the published baseline. The ranges track the effect of varying the category weights.}
\label{fig:aggregation_sensitivity}
\par}

\end{table*}

\subsection{Aggregation Sensitivity}
\label{app:diagnostics:sensitivity}
We test how aggregation affects model comparisons while retaining the published metric inputs. The reference score assigns equal weight to six categories. We compare it with equal weighting of all seventeen metrics and six alternatives that each remove one category and average the remaining five. Embodiment-specific analyses retain their fixed applicable metric sets: 17 metrics for human hands, 16 for simulated hands, and 15 for real hands.

For bounded changes in emphasis, each category weight lies in $[1/12,1/4]$, and the six weights sum to one. A grid with step $1/60$ gives 88,913 weight vectors. We compute scores, ranks, and Spearman correlation with the published ranking for each vector. Continuous pairwise score-gap extrema are evaluated at the twenty vertices of this bounded simplex. Ranks use unrounded inputs, with average ranks for ties.

\paragraph{Overall ranking.}
JointControlVideo remains first under metric-level averaging and all six category removals (Table~\ref{tab:aggregation_alternatives}). Metric-level averaging gives 55.96, 53.27, and 52.27 to JointControlVideo, HunyuanVideo 1.5, and Wan2.2, respectively. Across the bounded weight grid, ranking correlation ranges from 0.909 to 1.000, and no model moves more than three positions (Figure~\ref{fig:aggregation_sensitivity}). JointControlVideo remains ahead of every other model throughout the continuous weight range; its gap to HunyuanVideo 1.5 ranges from 0.52 to 6.06 points. In total, 47 of the 55 pairwise model orderings are preserved throughout that range. The leading overall result is therefore stable, while several middle-ranking comparisons depend on category emphasis.

\paragraph{Embodiment-specific rankings.}
Weight changes have different effects across the three task sets (Table~\ref{tab:aggregation_subsets}). The human-hand winner can switch between JointControlVideo and HunyuanVideo 1.5, and the simulated-hand winner between Wan2.1 and JointControlVideo. JointControlVideo remains first on real hands, although individual models can move by three positions.

On the 200-case analysis subset, we also quantify these comparisons with paired bootstrap intervals (5,000 draws, JEPA recomputed on every draw). Removing one metric at a time while keeping the six categories leaves JointControlVideo first overall in all 17 settings; its smallest margin over any other model across these settings is 2.59 points (95\% CI $[0.03, 4.27]$), reached against HunyuanVideo 1.5 when Memory Consistency is removed. Removing a complete category also leaves JointControlVideo first, with margins over HunyuanVideo 1.5 from 2.10 to 5.59 points. When each category weight varies within $\pm25\%$ of equal weighting, the smallest margin is 2.40 points ($[0.77, 4.10]$); within $\pm50\%$, which corresponds to the $[1/12,1/4]$ range above, it is 0.98 points ($[-0.22, 2.25]$). This margin differs from the 0.52-point minimum reported above for the full set because the two analyses use different case sets; JointControlVideo remains first in both. Computing HWBScore within each embodiment and averaging the three embodiments equally keeps JointControlVideo first, followed by Wan2.2, with a margin of 5.94 points over HunyuanVideo 1.5 ($[3.45, 8.80]$).

\paragraph{Memory weighting.}
Holding the other categories and outer weights fixed, we remove Memory Consistency or change the within-category weights of State Progression and Memory Consistency to 3:1 (Table~\ref{tab:memory_sensitivity}). JointControlVideo retains first place overall and on human hands under both variants. On the simulated subset, the leader switches from Wan2.1 to JointControlVideo under either change. Memory is inapplicable on the real-robot subset, so its scores are identical in all three columns of the underlying analysis.

\paragraph{State-regulation and task-family sensitivity.}
On the 200-case analysis subset, omitting State Regulation leaves a paired JointControlVideo advantage of 3.03 points over HunyuanVideo 1.5 (95\% CI $[1.15, 5.05]$). JointControlVideo remains first when any one of the seven task families is removed, both with and without State Regulation; its advantage ranges from 2.76 to 4.47 points with all metrics and from 2.49 to 3.72 points without State Regulation. The remaining cases are identical across models in each removal, and JEPA is recomputed after each removal. To check that the State Regulation comparison does not depend on scoring batches, we rescore all 120 applicable cases for both models in a single run of the protocol in Appendix~\ref{app:metrics:pc}, with State Progression supplying the progress gate. The JointControlVideo advantage is $+0.188$ ($[0.092, 0.277]$). Real hands account for most of it ($+0.541$, $[0.351, 0.703]$, 37 cases), whereas the human ($+0.025$, 20 cases) and simulated ($+0.032$, 63 cases) differences have intervals that include zero. On the 117 cases with valid scores in both runs, the paired difference is $+0.201$ in each.

\section{Human Evaluation}
\label{app:human_evaluation}

\subsection{Study Design}
\label{app:human_evaluation:design}

The comparison records cover 2,911 video pairs from eleven models on 100 reference cases: 64 human-hand, 22 simulated dexterous-hand, and 14 real dexterous-hand cases. For a given reference case, a comparison shows two generated videos for the same task as A and B, alongside the reference video. A judgment is one recorded choice for one pair and one target, either overall performance or one of the six dimensions. The study records 20,244 judgments; 2,789 pairs have a recorded choice on all seven targets.

\begin{figure*}[!tp]
{\centering
\includegraphics[width=.63\linewidth,trim=0 317bp 0 0,clip]{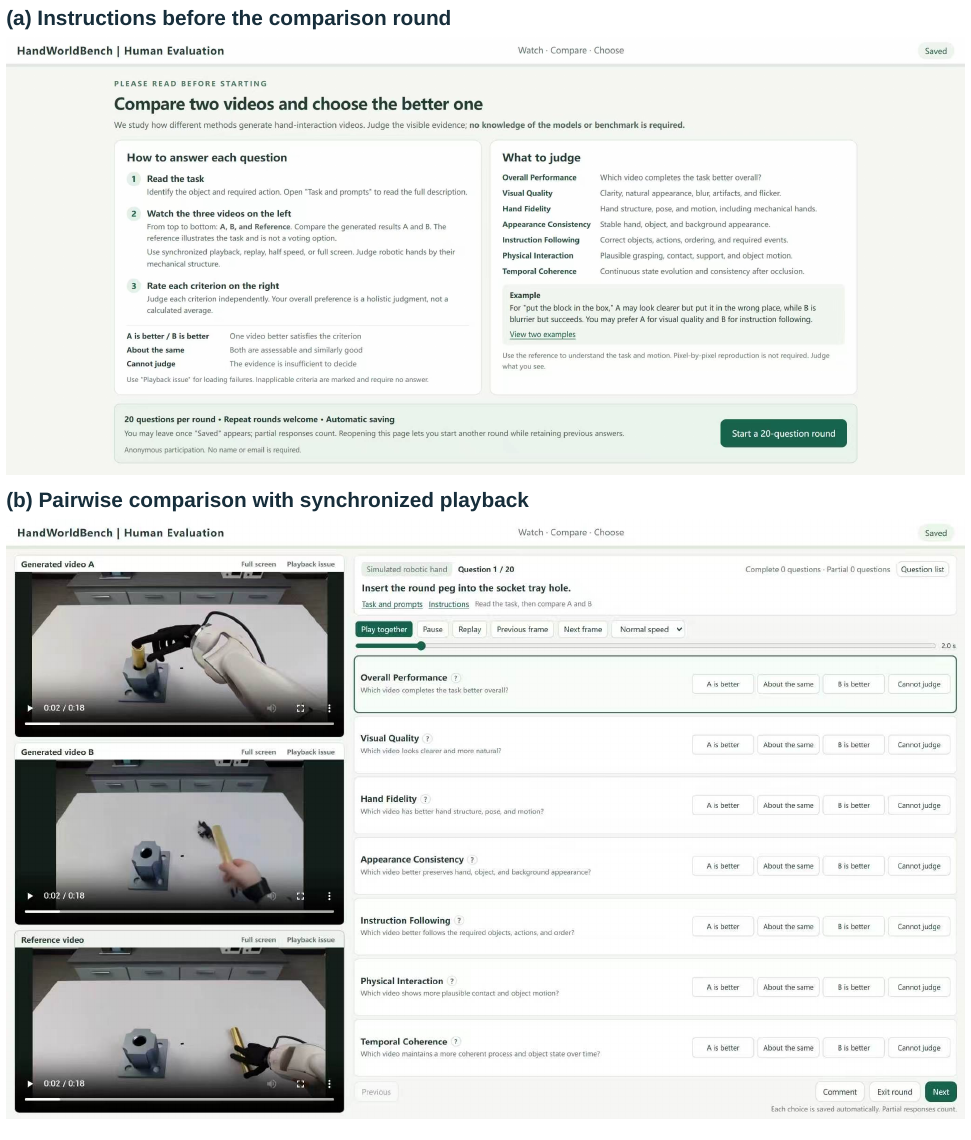}\par
\vspace{8pt}
\includegraphics[width=.63\linewidth,trim=0 0 0 230.037bp,clip]{fig/hwb_human_study_interface.pdf}\par
}
{\captionsetup{justification=raggedright,singlelinecheck=false}
\caption{\textbf{Human preference study interface.} Participants first read the task and rating instructions (a). For each comparison (b), the two generated videos and the reference appear on the left, with the task, synchronized playback controls, and seven response rows on the right. The reference provides context for the choice between A and B. Both panels show the interface used for data collection.}
\label{fig:human_study_interface}
\par}
\end{figure*}

\paragraph{Participant workflow.}
Participants first read the task and identify the object and required action. They then compare A and B with the reference as a task and motion guide. The interface provides synchronized playback, replay, frame stepping, speed controls, and full-screen viewing. Participants answer each applicable criterion independently, including a separate holistic choice for overall performance. The interface presents 20 comparisons per round, permits repeat rounds, and saves partial responses automatically (Fig.~\ref{fig:human_study_interface}).

The study involved 153 participants, all of whom gave informed consent before taking part. Responses are stored without participant identifiers.

\subsection{Evaluation Instructions and Quality Control}
\label{app:human_evaluation:instructions}

The instructions ask participants to judge visible evidence and assess robotic hands against their mechanical structure. They assess overall performance, visual quality, hand fidelity, appearance consistency, instruction following, physical consistency, and temporal coherence. For each applicable target, the options are \emph{A is better}, \emph{B is better}, \emph{About the same}, and \emph{Cannot judge}. A tie means that both videos are assessable and similar on that target. Participants can flag playback failures separately. Targets marked inapplicable require no answer. The interface label \emph{Physical interaction} corresponds to the Physical Consistency dimension.

We verify that each recorded comparison is unique by reference case, judgment target, and unordered model pair. We also check that automatic scores are consistent across repeated appearances of each case--model--target combination. Of 20,244 recorded judgments, 18,901 (93.4\%) are assessable, including 2,966 ties; 1,343 (6.6\%) are marked \emph{Cannot judge} (Table~\ref{tab:human_response_coverage}). Human win rates credit one point to a win, half a point to a tie, and zero to a loss. \emph{Cannot judge} responses are omitted. This coverage measures assessability within the preference study.

\begin{table*}[!tp]
{\captionsetup{justification=raggedright,singlelinecheck=false}
\centering
\caption{\textbf{Judgment coverage.} A/B preferences and assessable ties contribute to human preference rates. The unit is a recorded pair--target judgment. Playback reports, unanswered targets, and targets marked inapplicable are outside these counts.}
\label{tab:human_response_coverage}
\small
\begin{tabular*}{\textwidth}{@{\extracolsep{\fill}}lrrrrr@{}}
\toprule
Target & A/B preference & Tie & Cannot judge & Assessable & Recorded \\
\midrule
Overall performance & 2,293 & 162 & 455 & 2,455 & 2,910 \\
Visual quality & 2,599 & 226 & 83 & 2,825 & 2,908 \\
Hand fidelity & 2,627 & 175 & 107 & 2,802 & 2,909 \\
Appearance consistency & 2,518 & 191 & 199 & 2,709 & 2,908 \\
Instruction following & 2,129 & 699 & 80 & 2,828 & 2,908 \\
Physical consistency & 2,325 & 254 & 330 & 2,579 & 2,909 \\
Temporal coherence & 1,444 & 1,259 & 89 & 2,703 & 2,792 \\
\midrule
Total & 15,935 & 2,966 & 1,343 & 18,901 & 20,244 \\
\bottomrule
\end{tabular*}
\par}
\end{table*}

\subsection{Agreement Analysis}
\label{app:human_evaluation:agreement}

We assess agreement at the video-pair and model-ranking levels (Table~\ref{tab:human_agreement_full}). Pairwise agreement uses decisive A/B preferences with automatic scores for both models. A matching direction receives one point, an opposite direction zero, and an automatic tie half a point.

For model rankings, each dimension uses the reference tasks with automatic scores available for all eleven models and human comparisons on that same set. A model's human preference rate averages its win rates against the ten opponents equally. Its automatic score averages each reference task once. Spearman correlation compares the eleven-model rankings. The 95\% confidence intervals use 3,000 reference-task bootstrap samples stratified by embodiment (seed 20261115), retaining all comparisons associated with each sampled task.

Overall pairwise agreement is 81.7\%, with a model-ranking correlation of 0.964. Across the six dimensions, pairwise agreement ranges from 76.9\% to 84.1\%, and ranking correlation ranges from 0.934 to 0.991.

\begin{table*}[!tp]
{\captionsetup{justification=raggedright,singlelinecheck=false}
\centering
\caption{Complete agreement with human judgments. Decisive pairs count A/B preferences with both automatic scores available; common tasks are used for model-ranking correlation. Brackets give embodiment-stratified reference-task bootstrap 95\% confidence intervals.}
\label{tab:human_agreement_full}
\begingroup\small\setlength{\tabcolsep}{4pt}
\begin{tabular*}{\textwidth}{@{\extracolsep{\fill}}lrrrr@{}}
\toprule
Dimension & Decisive pairs & Common tasks & Preference agreement (\%) & Spearman $\rho$ \\
\midrule
Overall & 2,254 & 75 & 81.7 [80.1, 83.2] & 0.964 [0.791, 0.991] \\
Visual quality & 2,599 & 100 & 83.6 [82.2, 85.0] & 0.973 [0.882, 0.991] \\
Hand fidelity & 2,627 & 99 & 83.3 [81.8, 84.7] & 0.982 [0.855, 0.991] \\
Appearance consistency & 2,510 & 95 & 83.1 [81.7, 84.5] & 0.982 [0.809, 0.991] \\
Instruction following & 2,129 & 100 & 81.6 [80.2, 82.9] & 0.934 [0.709, 0.973] \\
Physical consistency & 2,297 & 79 & 84.1 [82.7, 85.6] & 0.955 [0.627, 0.982] \\
Temporal coherence & 1,444 & 96 & 76.9 [74.6, 79.2] & 0.991 [0.773, 0.991] \\
\bottomrule
\end{tabular*}
\endgroup
\par}
\end{table*}

\paragraph{Scope of the manual evidence.}
Each of the six evaluation dimensions is checked individually against the human choice for the same dimension (Tables~\ref{tab:human_agreement_full} and~\ref{tab:human_agreement_strata}). The comparison uses decisive human preferences only; ties and \emph{Cannot judge} responses are reported in Table~\ref{tab:human_response_coverage} and do not enter the agreement rates.\par

\begin{table*}[!tp]
{\captionsetup{justification=raggedright,singlelinecheck=false}
\centering
\caption{\textbf{Pairwise agreement by embodiment and automatic score gap.} Values are percentages, with eligible judgment counts in parentheses. Gap bins pool the three embodiments and split the absolute A--B score difference within each target at its median. Values at the median enter the lower bin; discrete scores can give unequal bin sizes.}
\label{tab:human_agreement_strata}
\small
\begin{tabular*}{\textwidth}{@{\extracolsep{\fill}}lrrrrr@{}}
\toprule
& \multicolumn{3}{c}{Embodiment} & \multicolumn{2}{c}{Automatic score gap} \\
\cmidrule(lr){2-4}\cmidrule(l){5-6}
Target & Human & Simulated & Real robot & At/below median & Above median \\
\midrule
Overall performance & 81.8 (1,420) & 83.1 (532) & 79.1 (302) & 79.6 (1,127) & 83.9 (1,127) \\
Visual quality & 83.7 (1,665) & 83.8 (561) & 82.6 (373) & 82.3 (1,300) & 84.8 (1,299) \\
Hand fidelity & 82.6 (1,708) & 85.3 (571) & 83.3 (348) & 82.6 (1,314) & 83.9 (1,313) \\
Appearance consistency & 81.4 (1,608) & 84.9 (550) & 87.8 (352) & 80.5 (1,255) & 85.7 (1,255) \\
Instruction following & 82.0 (1,330) & 81.2 (486) & 80.7 (313) & 77.3 (1,078) & 86.0 (1,051) \\
Physical consistency & 84.0 (1,412) & 84.7 (516) & 83.7 (369) & 81.6 (1,151) & 86.6 (1,146) \\
Temporal coherence & 76.7 (923) & 77.0 (315) & 77.7 (206) & 73.7 (1,146) & 88.9 (298) \\
\bottomrule
\end{tabular*}
\par}
\end{table*}

\paragraph{Where pairwise agreement varies.}
Temporal coherence has the lowest pairwise agreement for each embodiment (76.7\%, 77.0\%, and 77.7\%; Table~\ref{tab:human_agreement_strata}). It also has the most human ties (1,259 of 2,792 recorded responses). Across the 1,444 eligible decisive temporal judgments, 394 automatic comparisons tie. Agreement rises from 73.7\% at or below the median automatic gap to 88.9\% above it. Score-gap comparisons are descriptive because the larger-gap temporal bin covers fewer reference tasks (43 versus 96).
Different targets can favor different videos: among pairs with decisive human choices on both visual quality and overall performance, 801 of 2,069 (38.7\%) select different winners. These comparisons describe differences between judgment targets.\par

\section{Additional Qualitative Results}
\label{app:qualitative}
\begin{figure*}[!tp]
\centering\includegraphics[width=0.80\textwidth]{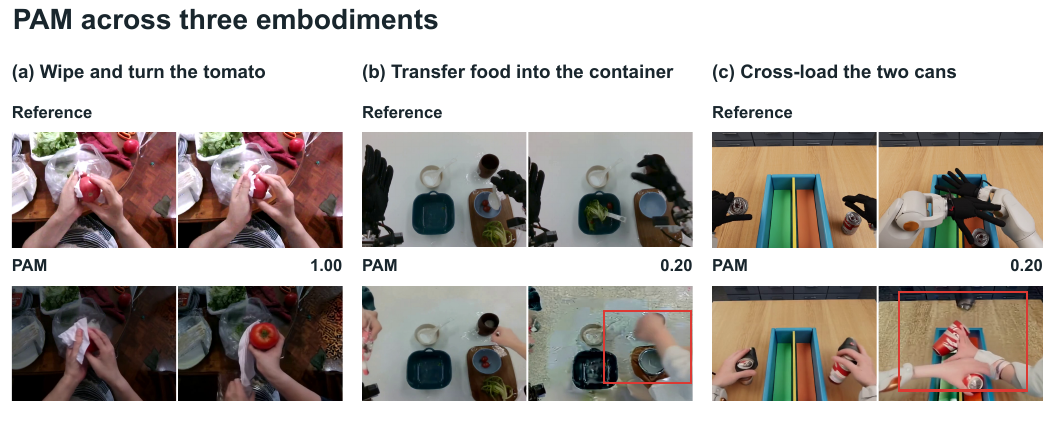}
\caption{\textbf{PAM across three embodiments.} (a) Human-hand tomato wiping and turning, (b) real-robot food transfer, and (c) simulated-hand can cross-loading, with Action Adherence scores of 1.00, 0.20, and 0.20. The upper rows show reference frames and the lower rows show PAM predictions.}
\label{fig:supp_pam}
\end{figure*}

The qualitative examples connect generated behavior to the reference task and its metric. Frames are decoded from the evaluated reference and prediction videos using the evaluated temporal mapping. Time labels follow the reference clock. Scores are the original clip-level results, and the red callouts identify visual evidence.

Each case retains its baseline configuration, reference identity, generated-video path, metric protocol, and temporal mapping. The baseline's configured interface determines the image, instruction, and motion inputs. Human keypoints, simulated kinematic geometry, and real-robot states or estimated projections retain their native conventions. The paired reference row establishes the expected task, manipulator, and object context for interpreting the generated frames.

\subsection{Cross-Embodiment Comparisons}
\label{app:qualitative:embodiments}
\begin{figure*}[!tp]
\centering\includegraphics[width=0.80\textwidth]{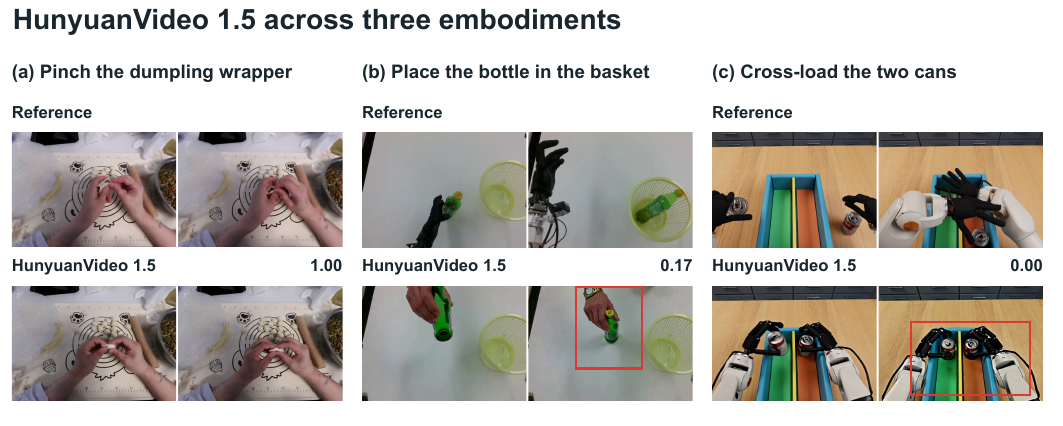}
\caption{\textbf{HunyuanVideo 1.5 across three embodiments.} (a) Human-hand wrapper pinching, (b) real-robot bottle placement, and (c) simulated-hand can cross-loading, with Evolution Correctness scores of 1.00, 0.17, and 0.00. The upper rows show reference frames and the lower rows show predictions.}
\label{fig:supp_hunyuan}
\end{figure*}
PAM preserves the human tomato-wiping interaction while changing the manipulator in the paired real and simulated robot scenes (Figure~\ref{fig:supp_pam}). The human output retains the wiping/rotation sequence and scores 1.00 on Action Adherence. Food transfer and can loading each score 0.20, with bare human hands appearing in the robot environments. The examples reveal how the scene can remain recognizable while the requested embodied interaction changes. Across the two figures, each triplet keeps one model and one task family fixed across embodiments, exposing hand substitution, object drift, and incomplete outcomes.

HunyuanVideo 1.5 produces the local wrapper-edge join in the human reference, while the robot examples show weaker object outcomes (Figure~\ref{fig:supp_hunyuan}). The bottle remains on the table as a human hand appears, and the can-loading output leaves the required placement incomplete. These predictions separate visible hand activity from the object change specified by the task.

\paragraph{Window-grounded state evidence.}
The wrapper case uses a process window of 0.7--4.3\,s and a result window of 4.3--8.03\,s. Its criterion is the local meeting and continued joining of the visible rim sections. The bottle case checks the result at 19.5--20.1\,s; the can case checks it at 9.8--10.7\,s. These bindings locate the result that the Evolution Correctness score addresses and connect the frame pair to its annotated process.

\begin{figure*}[!tp]
\centering\includegraphics[width=0.78\textwidth]{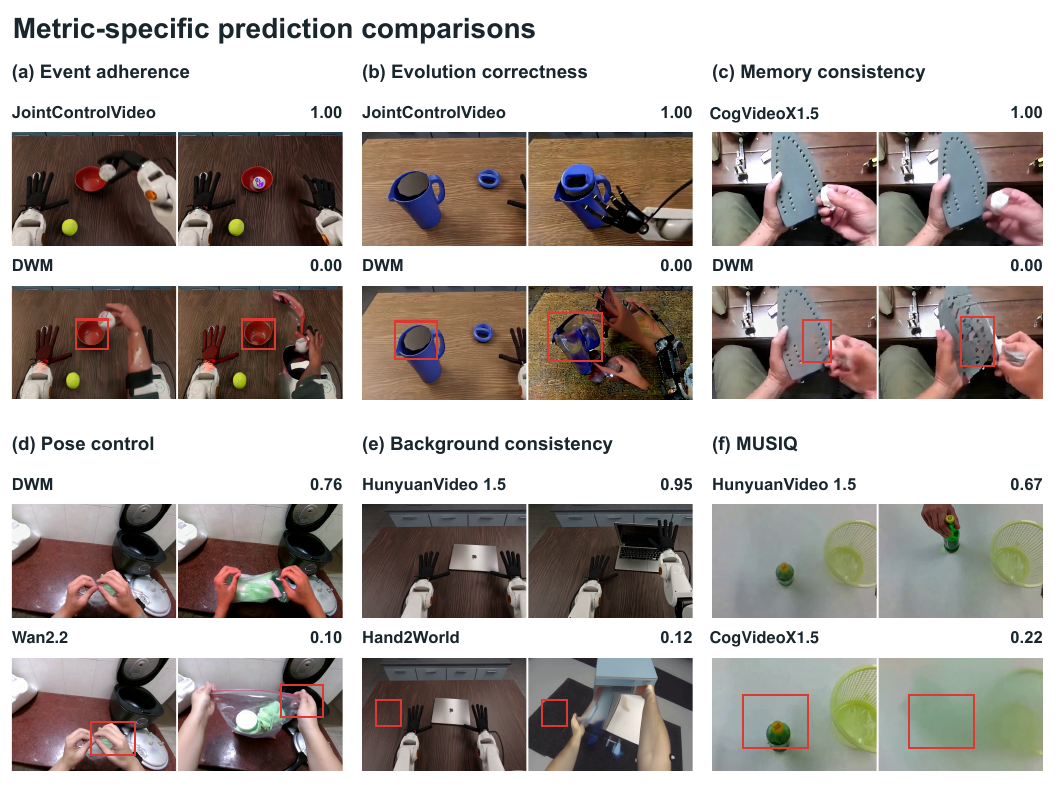}
\caption{\textbf{Metric-specific prediction comparisons.} Each panel shows earlier and later evidence for two models and the scores for the named metric. Panels (a)--(f) show Event Adherence, Evolution Correctness, Memory Consistency, Pose Control, Background Consistency, and MUSIQ. Red outlines mark regions of relevant visible differences. Panels (c)--(f) are discussed in Appendix~\ref{app:qualitative:failures}; panels (a) and (b) contrast an event with a transformation.}
\label{fig:supp_metrics}
\end{figure*}
 
\begin{figure*}[!tp]
\centering\includegraphics[width=0.78\textwidth]{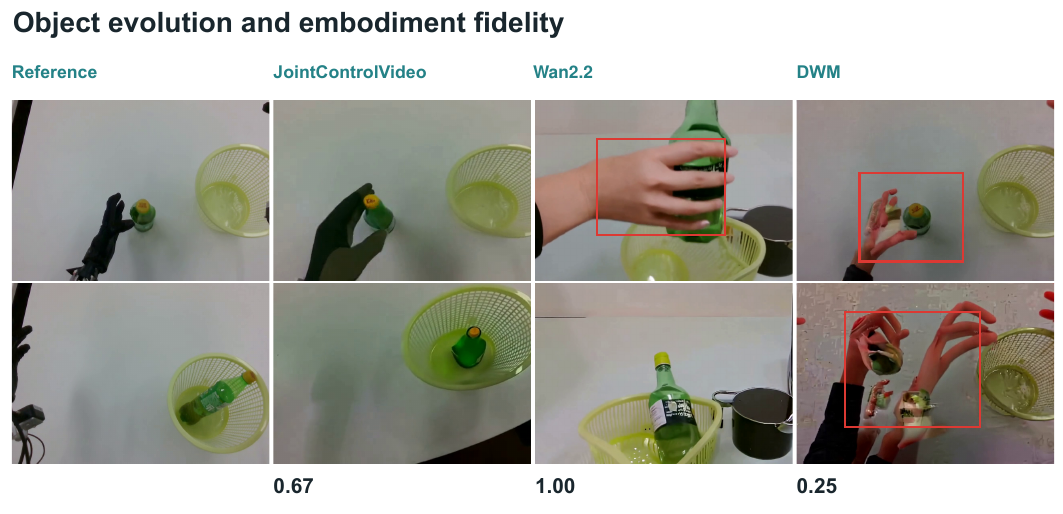}
\caption{\textbf{Object outcome and operator identity.} JointControlVideo, Wan2.2, and DWM score 0.67, 1.00, and 0.25 on Evolution Correctness for the same real-robot bottle reference. Wan2.2 reaches the object target with a human hand.}
\label{fig:supp_scope}
\end{figure*}

\paragraph{Events and transformations.}

Figure~\ref{fig:supp_metrics}(a) and (b) separate a discrete event from a transformation. In (a), the event is the ball entering the red bowl: JointControlVideo scores 1.00 and DWM 0.00 on Event Adherence. In (b), the transformation is fitting the blue lid onto the pitcher, and Evolution Correctness checks the final lid--pitcher connection (JointControlVideo 1.00, DWM 0.00). In every panel, the displayed frames come from one case and remain in temporal order.

\subsection{Interaction Failure Modes}
\label{app:qualitative:failures}
Figure~\ref{fig:supp_metrics} exposes complementary failure modes. The upper row addresses task events, object transformation, and historical detail. The lower row addresses hand-pose agreement, scene stability, and perceptual quality. Each panel identifies the baseline and metric whose score is shown.

\paragraph{Historical object detail.}
In the soleplate case, the memory criterion concerns the lower-right outer-hole arrangement relative to the inner row. The reference sequence and question are shown in Figure~\ref{fig:supp_memory_reference}. CogVideoX1.5 scores 1.00 and DWM 0.00 on Memory Consistency. The comparison follows the same object region through the history, hidden, and reappearance windows, with viewpoint and cloth motion treated as allowed changes. It tests the retention of a specific layout through the interaction.

\paragraph{Complementary visual evidence.}
Pose Control evaluates reference-keypoint agreement with missing-unit penalties. In the bag example, DWM scores 0.76 and Wan2.2 scores 0.10. For background stability during laptop opening, HunyuanVideo 1.5 scores 0.95 and Hand2World 0.12. For perceptual quality in the bottle example, MUSIQ scores are 0.67 for HunyuanVideo 1.5 and 0.22 for CogVideoX1.5.

\subsection{Evaluator Disagreements}
\label{app:qualitative:disagreements}

Metrics can disagree on the same video. In the real-robot bottle case of Figure~\ref{fig:supp_scope}, Wan2.2 moves the bottle to its target and receives Evolution Correctness 1.00, although a human hand replaces the robot manipulator. JointControlVideo retains the robotic manipulator with an evolution score of 0.67. Inspecting the object change and the manipulator identity on the same frames shows how a convincing local transformation can score highly alongside an evident embodiment error; the object-evolution score addresses the evaluated transformation, and the hand and entity metrics address manipulator fidelity.

\paragraph{Locating a disagreement.}
The evidence chain localizes inspection to detection, segmentation, geometry, temporal binding, or semantic judgment. For a missed hand or object, the detected region and mask are inspected first. For a trajectory discrepancy, tracked pixels are checked against the reference target and its source mapping. For an asserted connection or preserved detail, the target region is inspected within the annotated window using the specified criterion and allowed changes.

Expected, observed, and penalized unit counts distinguish unavailable keypoints from positional disagreement. In a memory test, target reappearance and attribute preservation remain separate decisions, so each case score can be traced to the observation that produced it.

\end{document}